\documentclass[10pt,twocolumn,letterpaper]{article}
\usepackage{cvpr}              
\usepackage{graphicx}
\usepackage{amssymb}
\usepackage{booktabs}
\usepackage{multirow}
\usepackage{amsmath}
\usepackage{subcaption}
\usepackage[dvipsnames]{xcolor}

\usepackage{amsmath}
\usepackage{tabularx} 
\usepackage{booktabs}
\usepackage{tikz}
\usepackage{xcolor}         
\usepackage{colortbl}
\usepackage{rotating}

\usepackage{bbding}
\usepackage{multirow}
\usepackage{xcolor}         
\usepackage{colortbl}
\usepackage{overpic}
\usepackage{indentfirst} 
\usepackage{graphicx}

\usetikzlibrary{tikzmark, positioning, arrows.meta}
\usepackage[dvipsnames]{xcolor}

\definecolor{cvprblue}{rgb}{0.21,0.49,0.74}
\usepackage[pagebackref,breaklinks,colorlinks,citecolor=cvprblue]{hyperref}
\usepackage[capitalize]{cleveref}
\crefname{section}{Sec.}{Secs.}
\Crefname{section}{Section}{Sections}
\Crefname{table}{Table}{Tables}
\crefname{table}{Tab.}{Tabs.}

\def\paperID{2167} 
\def\confName{CVPR}
\def\confYear{2024}

\title{SYNC-CLIP: Synthetic Data Make CLIP Generalize Better in\\
Data-Limited Scenarios}

\author{Mushui Liu ~~ Weijie He ~~ Ziqian Lu ~~ Yunlong Yu\thanks{Corresponding author.}\\
{\tt\small lms@zju.edu.cn yuyunlong@zju.edu.cn}
}

\begin{document}
\definecolor{lightblue}{rgb}{0.93,0.95,1.0}
\definecolor{mediumblue}{rgb}{0.0,0.45,0.73}

\maketitle

\begin{abstract}
Prompt learning is a powerful technique for transferring Vision-Language Models (VLMs) such as CLIP to downstream tasks. However, the prompt-based methods that are fine-tuned solely with base classes may struggle to generalize to novel classes in open-vocabulary scenarios, especially when data are limited. To address this issue, we propose an innovative approach called \textbf{SYNC-CLIP} that leverages \textbf{SYN}theti\textbf{C} data for enhancing the generalization capability of CLIP. Based on the observation of the distribution shift between the real and synthetic samples, we treat real and synthetic samples as distinct domains and propose to optimize separate domain prompts to capture domain-specific information, along with the shared visual prompts to preserve the semantic consistency between two domains. By aligning the cross-domain features, the synthetic data from novel classes can provide implicit guidance to rebalance the decision boundaries. Experimental results on three model generalization tasks demonstrate that our method performs very competitively across various benchmarks. Notably, SYNC-CLIP outperforms the state-of-the-art competitor PromptSRC by an average improvement of 3.0\% on novel classes across 11 datasets in open-vocabulary scenarios. 








\end{abstract}

\vspace{-2ex}
\section{Introduction} \label{sec: intro}

Recently, the pre-trained Vision-Language Models (VLMs) such as CLIP \cite{clip} and ALIGN \cite{ALIGN} have demonstrated impressive generalization capabilities across various downstream tasks, including image recognition \cite{coop, cocoop}, object detection \cite{gu2021open}, image segmentation \cite{denseclip}, and action recognition \cite{wasim2023vita}. 


To fine-tune the VLMs for further improvement, some prompt-based \cite{coop, cocoop, maple, PromptSRC} and adapter-based \cite{clip-adapter, tip-adapter, udandarao2022sus, cafo} methods have emerged to quickly adapt the pre-trained model to downstream tasks, by introducing a few learnable parameters. The core idea of these methods is to reorganize the feature representations to fit downstream tasks. Though efficient, these methods \cite{coop, kgcoop, PromptSRC, tip-adapter} are prone to severe imbalance issues when dealing with open-vocabulary scenarios, where some novel classes are encountered during inference. In essence, models trained solely on the base data might overfit the base classes, resulting in poor generalization capability to novel classes. 

In this paper, we propose a prompt-based method named \textbf{SYNC-CLIP} to alleviate the imbalance issues by synthesizing visual samples during training to rebalance the decision boundaries. Thanks to advances in text-to-image generation, generative models \cite{glide, dall-e, saharia2022photorealistic} have demonstrated the capability to produce high-fidelity, photo-realistic images at high resolutions via text descriptions. Recent approaches \cite{cafo,udandarao2022sus,he2022synthetic} also attempt to utilize synthetic data for training classification models for data-limited tasks. However, since the objectives of image generation models and classification tasks are not aligned, the distributions of synthetic and real samples may differ. Thus the attempts to treat both synthetic and real data equally when training the model would result in suboptimal results.

\begin{figure*}
    \centering
    \includegraphics[width=0.99\linewidth]{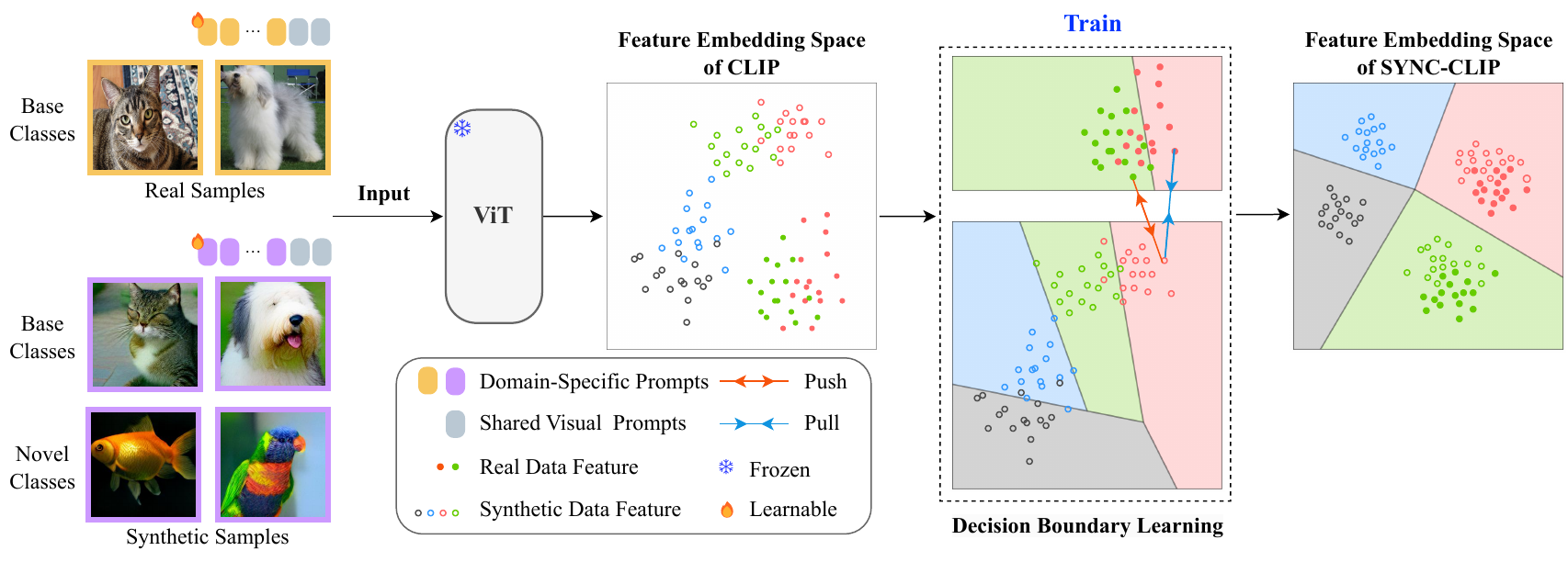}    
    \caption{\textbf{Pipline of our approach}. We treat real and synthetic samples as distinct domains, allocating separate visual prompts for each domain. The domain prompts learn domain-specific information, while the shared visual prompts capture domain-invariant details. Cross-domain feature alignment operations (\textcolor[HTML]{F7530C}{\textbf{Push}} and \textcolor[HTML]{1094E0}{\textbf{Pull}}) aid in rebalancing decision boundaries.}
    \label{fig:pipeline}
\end{figure*}

The problem described above inspires our approach to more effectively utilize synthetic data by learning distinct prompts for various derivatives of the data distributions. Drawing inspiration from the divide and conquer algorithm, we explicitly partition the feature embedding space and the data distribution into two segments based on the data sources. Subsequently, we learn the separate domain-specific prompts for each distribution and its corresponding part of the distribution. The separate domain-specific prompts combined with the shared visual prompts are trained on their corresponding feature embedding space, as illustrated in \cref{fig:pipeline}. The domain-specific prompt guides the pre-trained model in learning information specific to particular domains, while the shared visual prompts capture domain-invariant information, thereby enhancing the generalization capability.

To mitigate the distribution shift between the real samples and synthetic samples, we align the synthetic feature embedding space and the real feature embedding space based on semantic consistency, using a cross-domain feature alignment loss. Once aligned, the synthetic data from the novel classes without access to real visual samples could provide valuable information for rebalancing the decision boundaries, consequently alleviating the imbalance issues encountered in the open-vocabulary tasks. 



In conclusion, our main contributions include:

\begin{itemize}
    \item We provide an empirical study under the open-vocabulary few-shot setting to demonstrate the sub-optimality of the existing prompt-based approaches with synthetic data.

    \item With domain-specific prompts and shared visual prompts, our innovative approach SYNC-CLIP enhances the model's generalization capability. Additionally, by aligning the cross-domain features, our model allows synthetic data from novel classes to provide implicit guidance for rebalancing decision boundaries.
    
    \item In various open-vocabulary and cross-domain experiments, our approach exhibits competitive performance and achieves a more equitable balance in performance between base and novel classes.  
\end{itemize}

\section{Related Work}
\label{sec:related}

\textbf{Prompt Learning Based on VLMs.}  Recent years have witnessed remarkable achievements on large-scale pre-trained vision-language models \cite{clip, ALIGN, wang2021simvlm, alayrac2022flamingo, wang2022beit, huang2023clover}. Representatively, CLIP, ALIGN \cite{clip, ALIGN} jointly associate the images and their corresponding text descriptions by optimizing a contrastive objective. Prompt learning is widely used in large language models \cite{houlsby2019parameter,liu2023pre}, drawing notable interest in the fields of vision and multi-modality \cite{coop, vpt, sam}. Based on the CLIP model, Context Optimization (CoOp) \cite{coop} enhances the downstream few-shot image recognition tasks by refining the learnable soft textual prompts. Similarly, Visual Prompt Tuning (VPT) \cite{vpt} introduces vision prompts to large vision models. CoCoOp \cite{cocoop} and MaPLe \cite{maple} further augment generalization capabilities by incorporating image-conditioned information and multi-modal prompts, respectively. Despite the efficiency, these methods may overfit the task-specific distribution. PromptSRC \cite{PromptSRC} and KgCoOp \cite{kgcoop} attempt to exploit the task-agnostic information with prompt regularization. However, these methods may be struggled under the open-vocabulary few-shot scenarios. In this paper, we design an innovative prompt-based method to effectively exploit the synthetic samples generated with the off-the-shelf text-to-image generation models, to handle the absence of novel classes during training. 



\textbf{Adapting Synthetic Data to Downstream Tasks.} Generative models \cite{vae, gan, ddpm, ddpm2015} have made significant strides in the domain of image synthesis. Thus recent works \cite{jahanian2021generative, zhang2021datasetgan, 2020conditionGAN, cupl, cafo, udandarao2022sus, he2022synthetic} attempt to levarage the synthetic data for enhancing the performance of downstream tasks. \cite{jahanian2021generative, zhang2021datasetgan, 2020conditionGAN} utilizes the GANs \cite{gan, stylegan} to generate images for classification, object part segmentation, and unsupervised contrastive representation learning, respectively. Additionally, advances in text-to-image generation models \cite{dall-e, glide, saharia2022photorealistic} have spurred research \cite{cafo, udandarao2022sus, he2022synthetic} utilizing the generative models like DALL-E \cite{dall-e} and StableDiffusion \cite{saharia2022photorealistic} to synthesize data by text descriptions for downstream classification tasks. CaFo \cite{cafo} employs a cascade of multi-foundation models for few-shot classification. SuS-X \cite{udandarao2022sus} constructs a support set using synthetic data to assist classification. \cite{he2022synthetic} enhances the performance by improving the quality of synthetic data. In this paper, we investigate the distribution difference between the synthetic data and the real data and design the domain-specific prompts to separately optimize and align these two distributions.

\section{Preliminary Analysis}

In this section, we initially present the problem formulation of prompt-based learning and subsequently conduct a comprehensive empirical study to assess the results of the prompt-based methods with synthetic data.

\subsection{Formulation} 



\textbf{Adaptation of pre-trained VLMs} aims to adapt VLMs for downstream tasks, with or without the incorporation of additional training data \cite{coop, shu2022test}. The VLMs are well-trained by aligning the semantic consistency between the visual images and the text descriptions, yielding a visual encoder $\Theta_{I}$ and a text encoder $\Theta_{T}$ that respectively project the visual images and text inputs into a common space, where both zero-shot learning (ZSL) and few-shot learning (FSL) are achieved. In this work, we illustrate the process using the pre-trained CLIP models as an example.

\subsection{Prompt-based Approaches} 
\textbf{Prompt-based approaches} exploit the pre-trained knowledge adaptively with a few parameters while freezing visual and textual backbone parameters, aiming at {efficiently} adapting the VLMs to the downstream tasks. Both the visual and textual backbones consist of multiple consecutive multi-head self-attention (MSA) layers that transform an input sample into a sequence-like output representation. Here we denote the input of the $l$-th MSA layer as $x^l$, which consists of multiple tokens. The prompt-based approaches usually append the visual or textual prompts $p^l$ to the input $x^l$, then the {module's} input becomes $\hat{x}^l = \{x^l, p^l\}$. 




\textbf{CoCoOp} \cite{cocoop} employs an image-conditional textual prompt for text input. Specifically, CoCoOp initiates textual prompts $p_t^0$ into the initial text input $t^0$ and introduces a parameterized MetaNet denoted as $\mathcal{M}$ to infuse image information into the textual prompts, formulated as:
\begin{equation}
    \label{eq:cocoop}
    \hat{p}_t^0 = {\mathcal{M}}\left(\Theta_{I}\left(x_v^0\right)\right) + p_t^0,
\end{equation}
where $\mathcal{M}$ is a light-weight neural network, $\Theta_{I}\left(x_v^0\right)$ is the image feature. CoCoOp enables the dynamic adaptation of textual prompts based on visual information. 


 \begin{figure}
    \centering
    \begin{overpic}[width=0.9\linewidth]{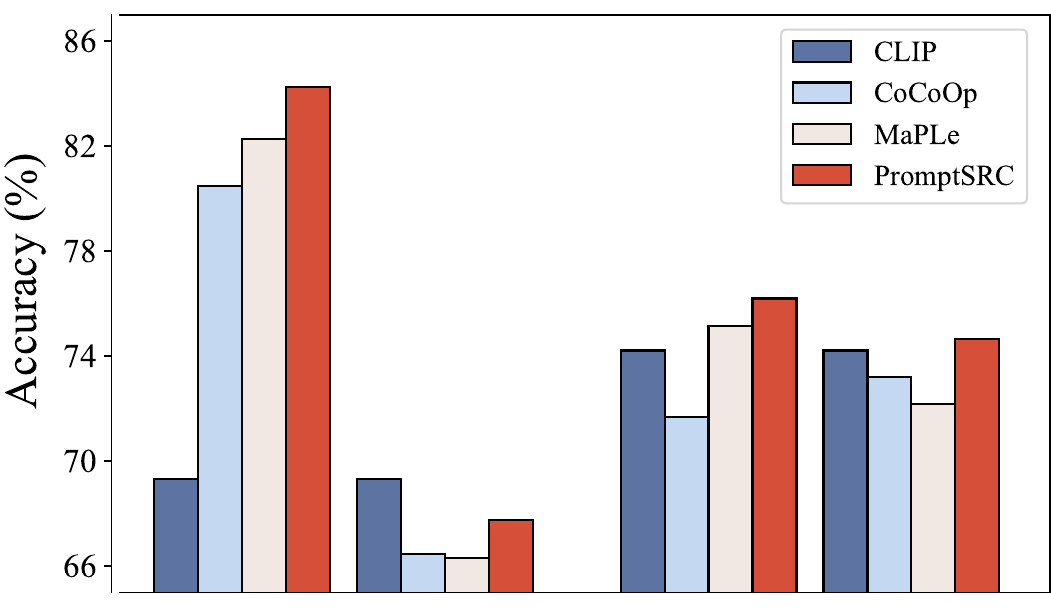} 
    \put(22,-2){\footnotesize R}    
    \put(42,-2){\footnotesize S}
    \put(25,-6){\footnotesize {Base Classes}} 
    \put(66,-2){\footnotesize R}    
    \put(86,-2){\footnotesize S}
    \put(70,-6){\footnotesize {Novel Classes}} 
    \end{overpic}
    \vspace{3mm}
    \caption{Empirical study of prompt-based methods on the average of 11 benchmarks respectively trained with real base data (R) and synthetic data (S), in terms of accuracy metric on base classes and novel classes. }
    \label{fig:empirical}
\end{figure}
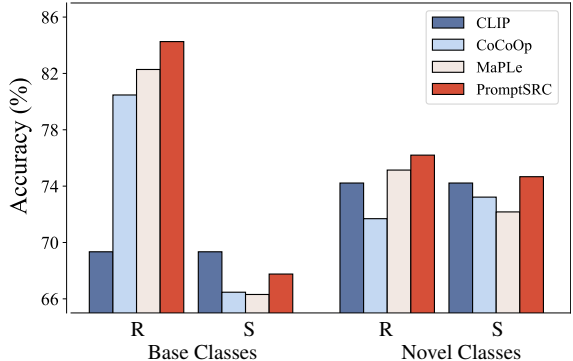

\textbf{MaPLe} \cite{maple} extends the prompts from language branch to multi-modal branches. It appends learnable textual prompts $p_t=\{p_t^i\}_{i=0}^l$ and conditions the visual prompts $p_v=\{p_v^i\}_{i=0}^l$ through coupling functions in multiple layers, establishing robust interdependence between vision and language prompts. Visual prompts $p_v^i$ used in $i$-th layer are obtained by projecting textual prompts $p_t^i$ via: 
\begin{equation}
    \label{eq:maple}
    p_v^i = \mathcal{F}(p_t^i)
\end{equation}
where $\mathcal{F}(\cdot)$ is a single linear projection layer. MaPLe aligns the visual and textual modality for better adaption.


\textbf{PromptSRC} \cite{PromptSRC} {also} utilizes the textual prompts $p_t$ and visual prompts $p_v$ to transfer CLIP model. Moreover, it introduces several regularization approaches to adjust the feature representation, \eg hand-craft prompts constraint, prompts ensemble, and diversity textual prompts. Thus, these self-regularizations help the model alleviate the overfitting issue of prompt learning.



\subsection{Empirical Study of Synthetic Data} \label{subsec: study}

Whether implicitly or explicitly, the prompt-based approaches mentioned above involve incorporating knowledge from the base data into prompt parameters. These parameters are then used to predict the identities of novel classes from uninstructed representations. While these methods can significantly enhance the performance of base classes, the improvements for novel classes are generally modest. In some cases, these methods may even have a detrimental impact on the performance of novel classes, as demonstrated in \cref{tab:open-vocalbu}. To address this limitation, a straightforward strategy to boost the classification performance of novel classes is to synthesize visual samples for these classes using generative models. To evaluate the impact of synthetic data for novel classes, we conduct an empirical study on the widely-used benchmarks, employing prompt-based models trained with data synthesized by the generative model.


In \cref{fig:empirical}, we present the average results from 11 benchmarks with the participation of synthetic data from the DALL-E \cite{dall-e} model. Remarkably, the performance of existing prompt-based models trained solely on synthetic data is even worse than that of pre-trained CLIP. This observation implies that \textit{the synthetic samples are ill-suited for enhancing the performance of prompt-based methods in novel class classification, despite their high fidelity at high resolutions}\footnote{Some synthetic samples are provided in the Appendix.}.

\begin{figure}
  \centering
  \begin{overpic}[width=1\linewidth]{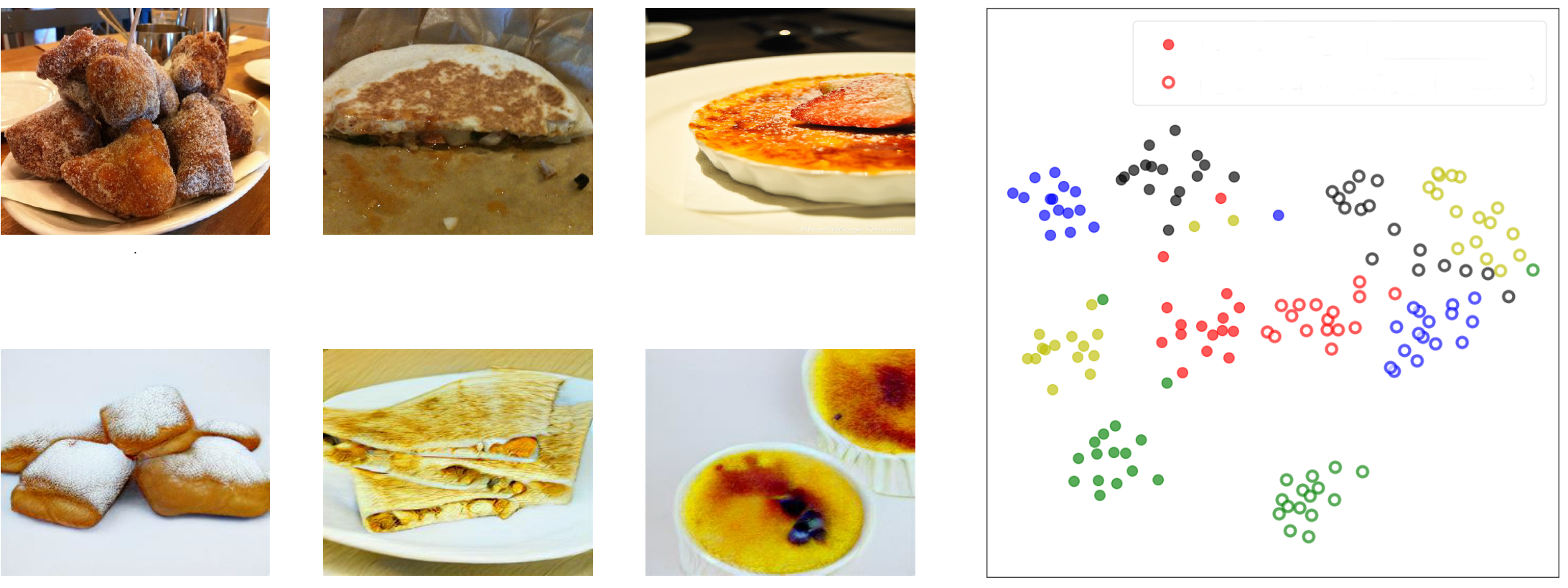}
  \put(17,19){\footnotesize (a) Real Samples}
  \put(17,-4){\footnotesize (b) Synthetic Samples}
  \put(65,-4){\footnotesize (c) t-SNE Visualization}
  \put(76,34){\tiny Real data feature}
  \put(76,31.5){\tiny Synthetic data feature}
  \end{overpic}
  \vspace{-2.0mm}  
  \caption{(a) Real samples and (b) synthetic samples from the Food101 dataset. (c) The t-SNE visualizations of both real and synthetic samples from the Food101 dataset in the feature embedding space spanned by the visual backbone of pre-trained CLIP. The same color represents samples from the same category.}
\label{fig:fake-tsne}
\end{figure}
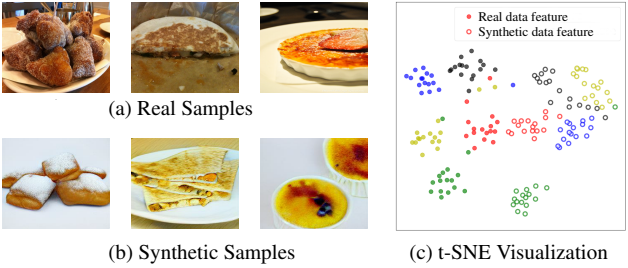

To explore the underlying causes, we analyzed the distributions of both real and synthetic samples in the feature embedding space, which was extracted using the visual backbone of the pre-trained CLIP. As illustrated in \cref{fig:fake-tsne}, the synthetic visual features exhibit significant divergence from the real visual features, despite belonging to the same classes. This substantial difference in distributions is accountable for the observed decline in performance when fine-tuning the model with synthetic data. It's noteworthy that, while the synthetic feature distribution structure is reminiscent of the real feature distribution, the substantial distance between the two indicates the need for further research on effective strategies to leverage synthetic visual data for learning novel classes.

\section{SYNC-CLIP}
In this section, we first present the baseline and then provide a detailed introduction to the proposed framework SYNC-CLIP. 




\subsection{Baseline}
Due to the efficiency, we adopt a multi-modality prompt learning method named the Independent Vision-Language Prompting (\textbf{IVLP}) introduced in \cite{maple, PromptSRC}, as our baseline. IVLP consists of multiple hierarchy visual and textual prompts injected into the transformer blocks. For the visual modality, its inputs consist of $H$ visual tokens $x_v = \{x_{v_1}, x_{v_2}, \cdots, x_{v_H}\}$ of image $x$ and $J$ learnable tokens ${p}_{v} = \{p_{v_1}, p_{v_2}, \cdots, p_{v_J}\}$, denoted as $\hat{x}=\{p_{v}, x_{v}\}$, thus the feature embedding of $x$ could be obtained with: 
\begin{equation}
   {f}_I(x)=\Theta_{I}\left(\hat{x}\right),
    \label{eq:visual prompt}
\end{equation}

Similarly, the feature embedding of class $y$ could be obtained with:
\begin{equation}
    {f}_T(y)=\Theta_{T}\left(\hat{y} \right),
    \label{eq:textual prompt}
\end{equation}
where $\hat{y}=\{p_t, y_t\}$ consists of $K$ learnable tokens ${p}_{t} = \{p_{t_1}, p_{t_2}, \cdots, p_{t_K}\}$ and the embedding token $y_t$ of the text description or name of class $y$.

Then, the probability of the visual sample $x$ belonging to the class $y_i$ could be obtained with:
\begin{align}
p(y_i|x)=\frac{\exp \left(s ({f}_I(x), f_T(y_i))\right)}{\sum_{c=1}^C \exp \left(s ({f}_I(x), f_T(y_c)) \right)},
\label{eq:probability}
\end{align}
where $s(\cdot)$ is the cosine similarity. By optimizing the multi-modal prompts via constraining samples to be correctly classified, both visual and textual representations are refined for adapting to downstream tasks.








\subsection{Division of the Visual Prompts}

As discussed in \cref{subsec: study}, the synthetic samples and the real samples may be located in different distributions. To effectively leverage the structural information from synthetic samples, we process synthetic data and real data separately, drawing inspiration from the well-known divide and conquer algorithm. Specifically, we consider synthetic samples and real samples as data from different domains and propose to learn domain-specific prompts for each domain and shared visual prompts for both domains, aiming to capture domain-specific information and domain-agnostic information, respectively.

\textbf{Domain-Specific Prompts.} For the real data, we define the real domain prompts ${p}_{v}^{r} = \{p_{v_1}, p_{v_2}, \cdots, p_{v_{M_1}}\}$. Similarly, for the synthetic samples, we define the synthetic domain prompts ${p}_{v}^{s} = \{p_{v_1}, p_{v_2}, \cdots, p_{v_{M_2}}\}$, where $M_1$ and $M_2$ represent the number of prompts, respectively.

\textbf{Shared Visual Prompts.} In addition to the domain-specific prompt, we also introduce the domain-agnostic prompts ${p}_{v}^{da} = \{p_{v_1},p_{v_2}, \cdots, p_{v_{N}}\}$ to capture the general information, where $N$ denotes the number of prompts.

With the combination of domain-specific prompts and shared prompts, the model can explicitly capture the uniqueness and relevance of the data. Consequently, the visual features for real samples $x_r$ and synthetic samples $x_s$ can be obtained with 
\begin{align}
    f_{I}(x_r) &= \Theta_I(\hat{x}_{r}), \label{eq:real} \\
    f_{I}(x_s) &= \Theta_I(\hat{x}_{s}), \label{eq:synthetic}
\end{align}
where $\hat{x}_{r}=\{{p}_{v}^{r}, {p}_{v}^{da}, x_r\}$ and $\hat{x}_{s}=\{{p}_{v}^{s}, {p}_{v}^{da}, x_s\}$ are the inputs for real samples and synthetic samples, respectively.



Next, we split the real samples and synthetic samples into two clusters and perform classification in their individual embedding space. For the real samples $x_r\in {I}^{R}$, the optimization objective is:
\begin{equation}
    \label{eq:rce}
    \mathcal{L}_{RCE} = -\sum_{x_r\in {I}^{R}} \log p(y|x_r), 
\end{equation}
where $p(y|x_r)$ denotes the probability of the sample $x_r$ belonging to the class $y \in \mathcal{Y}_b$ obtained with \cref{eq:probability}, in which the visual feature is obtained with \cref{eq:real} and $\mathcal{Y}_b$ represents the base class label spaces. For the synthetic samples $x_s\in {I}^{S}$, the optimization objective is:
\begin{equation}
\label{eq:sce}
    \mathcal{L}_{SCE} = -\sum_{x_s \in {I}^{S} } \log p(y|x_s), 
\end{equation}
where $p(y|x_s)$ denotes the probability of the visual sample $x_s$ belonging to the class $y \in \mathcal{Y}_b \cup \mathcal{Y}_n$ obtained with \cref{eq:probability}, $\mathcal{Y}_b$ and $\mathcal{Y}_n$ represent the base and novel class label spaces, respectively. The visual feature is obtained with \cref{eq:synthetic}. Note that $I^S$ consists of synthetic samples from both base and novel classes without access to real samples during training. 

By optimizing the model with both real and synthetic samples, the model captures the domain-specific and generalized domain-invariant information, which helps eliminate the influence of domains and enhances the model's generalization capability.

\subsection{Alignment of the Feature Spaces}


The above process optimizes synthetic samples and real samples separately, which cannot fully explore synthetic samples and fails to provide information for the decision boundaries of novel classes. To this end, we propose to align the real visual space and synthetic visual space to alleviate the shift between the two domains. Specifically, we select a synthetic base sample $x_{s}^a$ as the anchor and a real base sample $x_{r}^{a}$ derived from the same class, and a real sample $x_{r}^b$ from the distinct class to compose a triplet and align the two spaces with:
\begin{equation}
\label{eq:align}
\mathcal{L}_{FS} = \max \{ d(f_s^a, f_r^a)-d(f_s^a, f_r^b), 0 \} + d(f_s^a, f_r^a),
\end{equation}
where $f_s^a$, $f_r^a$, and $f_r^b$ denote the feature embeddings of sample $x_{s}^a$, $x_{r}^{a}$, and $x_{r}^b$, respectively. $d$ denotes the distance and we choose $L_1$ distance in this work. 

By minimizing \cref{eq:align}, the feature embeddings of synthetic samples and real samples belonging to the same class would be clustered together, while the samples from different classes would repel each other. This alignment of visual feature spaces maintains discrimination while ensuring alignment.  Once the real and synthetic feature spaces are aligned, the distribution shift between synthetic and real samples from novel classes is mitigated. This allows synthetic samples from novel classes to effectively substitute real samples, providing valuable information for learning their decision boundaries.

\subsection{Final Objective Function}
Though synthetic data and real data are separately modeled in their respective spaces, they can be jointly optimized, thus the final objective function is formulated as:
\begin{equation}
    \mathcal{L} = \mathcal{L}_{RCE} + \alpha \cdot \mathcal{L}_{SCE} + \beta \cdot \mathcal{L}_{FS},
\end{equation}
where $\alpha$ and $\beta$ are the two hypermeters to balance the items.

After training, we can obtain visual prompts and textual prompts adapted to the target classes, thus enhancing the model's generalization capability.

\section{Experiments}

\definecolor{tabcolor}{RGB}{200, 200, 200}
\def\tabwidth{.30}



\subsection{Experiment Settings}

\begin{table*}[!ht]
    \centering
    \resizebox{\linewidth}{!}{
        \begin{tabular}{l@{\hspace{0.1cm}}c|c@{\hspace{0.2cm}}c@{\hspace{0.2cm}}c@{\hspace{0.2cm}}c@{\hspace{0.2cm}}c@{\hspace{0.2cm}}c@{\hspace{0.2cm}}c@{\hspace{0.2cm}}c@{\hspace{0.2cm}}c@{\hspace{0.2cm}}c@{\hspace{0.2cm}}c@{\hspace{0.2cm}}c}
            \toprule
            Dataset & ~ & ImageNet & Caltech101 & OxfordPets & Cars & Flowers & Food101 & Aircraft & SUN397 & DTD & EuroSAT & UCF101 & Average \\ \midrule
            & B & 68.4 (72.4) & 93.6 (96.8) & 86.5 (91.2) & 59.5 (63.4) & 62.7 (72.1) & 85.5 (90.1) & 19.4 (27.2) & 60.3 (69.4) & 39.6 (53.2) & 45.7 (56.5) & 63.2 (70.5) & 62.2 (69.3) \\
            CLIP \cite{clip} & N & 65.0 (68.1) & 91.7 (94.0) & 89.7 (97.3) & 70.9 (74.9) & 71.0 (\textbf{77.8}) & 85.3 (91.2) & 28.0 (36.3) & 64.9 (75.4) & 49.6 (59.9) & 36.7 (64.1) & 67.0 (77.5) & 65.4 (74.2) \\
            ~ & HM & 66.7 (70.2) & 92.6 (95.4) & 88.1 (94.1) & 64.7 (68.7) & 66.6 (74.8) & 85.4 (90.7) & 22.9 (31.1) & 62.5 (72.2) & 44.0 (56.4) & 40.7 (60.0) & 65.0 (73.9) & 63.8 (71.7) \\
            \midrule
             & B & 72.2 (76.0) & 95.2 (98.0) & 90.7 (95.2) & 67.9 (70.5) & 86.0 (94.9) & 86.4 (90.7) & 26.3 (33.4) & 71.2 (79.7) & 60.1 (77.0) & 70.9 (87.5) & 74.2 (82.3) & 72.8 (80.5) \\
            CoCoOp \cite{cocoop} & N & 67.7 (70.4) & 90.9 (93.8) & 93.4 (97.7) & 69.7 (73.6) & 65.0 (71.8) & 86.6 (91.3) & 26.4 (23.7) & 67.4 (76.9) & 41.2 (56.0) & 42.0 (60.0) & 68.7(73.5) & 65.4 (71.7) \\
            ~ & HM & 69.9 (73.1) & 93.0 (95.8) & 92.0 (96.4) & 68.8 (72.0) & 74.1 (81.7) & 86.5 (91.0) & 26.3 (27.7) & 69.3 (78.3) & 48.9 (64.9) & 52.8 (71.2) & 71.4 (77.6) & 68.9 (75.8) \\
            \midrule
            & B & 72.8 (76.7) & 95.8 (97.7) & 91.0 (95.4) & 69.4 (72.9) & 91.0 (95.9) & 86.8 (90.7) & 24.9 (37.4) & 72.9 (80.8) & 63.5 (80.4) & 80.2 (94.1) & 76.3 (83.0) & 75.0 (82.3) \\
            MaPLe \cite{maple} & N & \textbf{68.1} (70.5) & 92.7 (94.4) & \textbf{93.8} (97.8) & 69.4 (74.0) & 66.9 (72.5) & 86.9 (\textbf{92.1}) & 31.1 (35.6) & 68.9 (78.7) & 46.6 (59.2) & 53.8 (73.2) & 72.5 (78.7) & 68.2 (75.1) \\
            ~ & HM & \textbf{70.4} (73.5) & 94.2 (96.0) & 92.4 (96.6) & 69.4 (73.5) & 77.1 (82.6) & 86.5 (\textbf{91.4}) & 27.7 (36.1) & 70.9 (79.8) & 53.8 (68.2) & 64.4 (82.4) & 74.4 (80.8) & 71.5 (78.6) \\
            \midrule
             & B & \textbf{73.9} (\textbf{77.6}) & 96.0 (98.1) & \textbf{93.3} (95.3) & 75.2 (78.3) & \textbf{93.8} (\textbf{98.1}) & \textbf{87.1} (\textbf{90.7}) & \textbf{35.5} (\textbf{42.7}) & \textbf{75.8} (\textbf{82.7}) & \textbf{67.4} (\textbf{83.4}) & \textbf{88.6} (92.9) & \textbf{81.0} (\textbf{87.1}) & \textbf{78.9} (\textbf{84.3}) \\
            PromptSRC \cite{PromptSRC} & N & 67.0 (\textbf{70.7}) & 91.6 (94.0) & 91.0 (97.3) & 71.1 (75.0) & 69.7 (76.5) & 86.0 (91.5) & 29.3 (37.9) & 69.3 (78.5) & 49.3 (63.0) & 52.6 (73.9) & 71.7 (78.8) & 68.0 (76.2) \\
            ~ & HM & 70.3 (\textbf{74.0}) & 93.8 (96.0) & 92.2 (96.3) & 73.1 (76.6) & 80.0 (\textbf{86.0}) & 86.5 (91.1) & 32.1 (40.2) & \textbf{72.1} (\textbf{80.5}) & 56.9 (\textbf{71.8}) & 66.0 (82.3) & 76.1 (\textbf{82.7}) & 73.0 (80.0) \\
            \midrule
            \rowcolor{gray!20}
            ~ & B & 73.3 (76.9) & \textbf{96.5} (\textbf{98.4}) & 92.6 (\textbf{95.4}) & \textbf{77.0} (\textbf{79.8}) & 93.3 (97.5) & 86.3 (90.6) & 31.2 (41.5) & 73.6 (81.4) & 65.6 (81.6) & 87.4 (\textbf{94.5}) & 79.5 (85.4) & 77.8 (83.9) \\
            \rowcolor{gray!20}
            ~ & N & 66.0 (70.0) & \textbf{93.6} (\textbf{95.2}) & 93.4 (\textbf{98.1}) & \textbf{72.5} (\textbf{76.1}) & \textbf{71.4} (76.0) & \textbf{87.4} (91.8) & \textbf{36.6} (\textbf{42.4}) & \textbf{70.0} (\textbf{79.3}) & \textbf{52.1} (\textbf{63.4}) & \textbf{65.8} (\textbf{78.6}) & \textbf{73.1} (\textbf{79.9}) & \textbf{71.0} (\textbf{77.4}) \\
            \rowcolor{gray!20}
            \multirow{-3}{*}{\textbf{SYNC-CLIP}} & HM & 69.4 (73.3) & \textbf{95.0} (\textbf{96.8}) & \textbf{93.0} (\textbf{96.8}) & \textbf{74.7} (\textbf{77.9}) & \textbf{80.9} (85.5) & \textbf{86.8} (91.2) & \textbf{33.7} (\textbf{41.9}) & 71.6 (80.3) & \textbf{58.1} (71.3) & \textbf{75.1} (\textbf{85.8}) & \textbf{76.2} (82.6) & \textbf{74.3} (\textbf{80.5}) \\
            \bottomrule
        \end{tabular}
    }
    \vspace{-0.2cm}
   \caption{Comparison performances (\%) under the GZSL (ZSL) setting. The ZSL results of competitors are directly obtained from the original literature, and the GZSL performances represent the best results achieved using codes released by ourselves. The best results are highlighted in bold. }
   \label{tab:open-vocalbu}
\end{table*}

\textbf{Traditional \& Generalized ZSL.} In this experiment, we train our model using the limited base dataset in an FSL scenario, where each base class is represented by only a few samples. Then, we evaluate the model's performance on base and novel test data under traditional ZSL and generalized ZSL settings. We report base accuracy (\textbf{B}), novel accuracy (\textbf{N}), and their harmonic mean (\textbf{HM}). Note that in traditional ZSL, the test data from base (novel) classes are exclusively classified into the base (novel) set. However, in the context of generalized ZSL (GZSL), the test data is classified into a unified class space that encompasses both the base and novel sets, which assesses the model's open-vocabulary generalization capability.

\textbf{Domain Generalization.} Based on whether the test samples belong to the domain of the training data, it can be divided into in-domain setting and out-domain setting. Note that under both settings, the training classes and the testing classes are the same. In our experiments, we follow the training protocol presented in \cite{cocoop}, training our model in the 16-shot training set and evaluating its performance on the full test set.


\textbf{Cross-Dataset Generalization.} Following the protocol in \cite{cocoop}, the model undergoes training on the ImageNet dataset in a few-shot scenario and is subsequently evaluated on other datasets.

\textbf{Dataset Settings.} For traditional and generalized ZSL, and cross-dataset settings, we use 11 image classification datasets, i.e., ImageNet \cite{ImageNet} and Caltech-101 \cite{caltech-101} for generic object classification, OxfordPets \cite{OxfordPets}, StanfordCars \cite{cars}, Flowers \cite{Flowers}, Food101 \cite{food101}, and FGVCAircraft \cite{FGVCAircraft} for fine-grained visual categorization, EuroSAT \cite{eurosat} for satellite image classification, UCF101 \cite{ucf101} for action recognition, DTD \cite{dtd} for texture classification, and SUN397 \cite{sun397} for scene recognition. We randomly sample 16 images (shots) from each base class in all the datasets mentioned above under both traditional and generalized ZSL settings. For Domain Generalization experiments, we designate ImageNet as the source domain and assess model performance across several target domains, including ImageNetV2 \cite{ImageNetV2}, ImageNet-Sketch \cite{imagenet-sketch}, ImageNet-A \cite{imagenet-a}, and ImageNet-R \cite{imagenet-r}.




\textbf{Implement Details.} In our implementation, we employ the pre-trained ViTB/16 of CLIP \cite{clip, vit} as the backbone. The optimizer employed is SGD with a cosine annealing strategy. The initial learning rate is set to 2.5e-3 and the batch size is set to 8 for all datasets. The hyperparameters $\alpha$, and $\beta$ are set to 0.1, 0.5 for most datasets, respectively. The length of visual prompts $M_1$, $M_2$, and $N$ are set to 2, 2, and 2, respectively. Note that mix-training is applied in all experiments, with a ratio of synthetic samples to real samples in each iteration set to 2:1.

\begin{table}[t]
    \centering
    \resizebox{\linewidth}{!}{
        \begin{tabular}{lcccccc}
        \toprule
        \multirow{2}{*}{Method} & In-Domain & \multicolumn{5}{c}{Out-of-Domain} \\
        \cmidrule(lr){2-2}  \cmidrule(lr){3-7}
        & ImageNet & -V2 & -S & -A & -R  & Aver. \\
        \midrule
        CLIP \cite{clip} & 66.73 & 60.83 & 46.15 & 47.77 & 73.96 & 57.18 \\
        CoOp \cite{coop} & \textbf{71.51} & 64.20 & 47.99 & 49.71 & 75.21 & 59.28 \\
        Co-CoOp \cite{cocoop} & 71.02 & 64.07 & 48.75 & 50.63 & 76.18 & 59.91 \\
        MaPLe \cite{maple} & 70.72 & 64.07 & 49.15 & \textbf{50.90} & 76.98 & 60.27 \\
        PromptSRC \cite{PromptSRC} & 71.27 & 64.35 & \textbf{49.55} & 50.90 & \textbf{77.80} & \textbf{60.65} \\        
        \midrule
        SYNC-CLIP & {71.50} & \textbf{64.78} & 49.38 & 50.28 & 76.92 & 60.34 \\
        \bottomrule    
        \end{tabular}   
    }
    \vspace{-0.2cm}
    \caption{Domain generalization performances (\%). The results of the competitors are directly from the original literature.}     
  \vspace{-2.5mm}
    
    \label{tab: domain shifts}
\end{table}

\begin{table*}[t]
\footnotesize
    \centering

    \resizebox{\linewidth}{!}{
    
    \begin{tabular}{lcccccccccccc}
    \toprule
    & \textbf{Source} & \multicolumn{11}{c}{\textbf{Target}} \\
    \cmidrule(lr){2-2} \cmidrule(lr){3-13}

    & ImageNet & Caltech101 & OxfordPets & Cars & Flowers & Food101 & Aircraft & SUN397 & DTD & EuroSAT & UCF101 & Average \\


    \midrule
    CoOp \cite{coop} & \textbf{71.51} & 93.70 & 89.14 & 64.51 & 68.71 & 85.30 & 18.47 & 64.15 & 41.92 & 46.39 & 66.55 & 63.88 \\
    Co-CoOp \cite{cocoop} & 71.02 & \textbf{94.43} & 90.14 & 65.32 & 71.88 & 86.06 & 22.94 & \textbf{67.36} & 45.73 & 45.37 & 68.21 & 65.74 \\
    MaPLe \cite{maple} & 70.72 & 93.53 & 90.49 & 65.57 & \textbf{72.23} & 86.20 & \textbf{24.74} & 67.01 & 46.49 & 48.06 & 68.69 & 66.30 \\    
    PromptSRC \cite{PromptSRC}& 71.27 & 93.60 & 90.25 & \textbf{65.70} & 70.25 & 86.15 & 23.90 & 67.10 & 46.87 & 45.50 & 68.75 & 65.81 \\
    \midrule
    SYNC-CLIP & 71.50 & 94.02 & \textbf{90.53} & 65.61 & 71.46 & \textbf{86.20} & 23.40 & 67.05 & \textbf{46.89} & \textbf{51.37} & \textbf{68.83} & \textbf{66.54} \\
    \bottomrule    
    \end{tabular}
    
    }

    \vspace{-0.2cm}
    \caption{Comparison results (\%) under cross-dataset setting. All methods are trained on ImageNet and evaluated on cross-datasets.}
    \label{tab:cross-dataset}
\end{table*}

\subsection{Performance Comparison}

\textbf{Traditional \& Generalized ZSL.} \cref{tab:open-vocalbu} shows both the GZSL and ZSL results of SYNC-CLIP and four competitors on 11 datasets. Based on the results, it is evident that SYNC-CLIP excels in both settings. In particular, SYNC-CLIP outperforms the second-best competitor by 1.3\% and 0.5\% on average across 11 datasets in terms of the \textbf{HM} metric under the GZSL and ZSL settings, respectively. In terms of the \textbf{B} metric, though SYNC-CLIP significantly improves the pre-trained CLIP, it holds only a marginal advantage compared to CoCoOP \cite{cocoop} and MaPLe \cite{maple}, and even performs slightly worse than PromptSRC. However, in terms of the \textbf{N} metric, SYNC-CLIP has a clear advantage on most of the datasets, especially under the GZSL setting. For example, SYNC-CLIP achieves 36.6\% and 65.8\% on the Aircraft and EuroSAT datasets under the GZSL setting, surpassing the second-best competitors by 5.5\% and 12.0\%, respectively. Additionally, we observe that the superiority of the existing competitors over the pre-trained CLIP primarily stems from the improvement of \textbf{B} while SYNC-CLIP achieves significant improvements in both \textbf{B} and \textbf{N}. The performance superiority of the \textbf{N} indicates that the synthetic data could offer valuable information for the novel classes when the model is well-designed. However, in contrast, the inclusion of additional synthetic data hardly improves the base classes, even under this data-limited scenario.  




\textbf{Domain Generalization.}
The domain generalization performances of our method, alongside five competitors, are illustrated in \cref{tab: domain shifts}. In this evaluation, the model is initially trained on the ImageNet dataset under the few-shot setting and subsequently tested on the distinct datasets, namely ImageNetv2, ImageNet-Sketch, ImageNet-A, and ImageNet-R, all sharing class labels with ImageNet but residing in different domains. While our proposed method performs competitively across all datasets, achieving the second-best average performance, it does not lead to additional improvements compared to prompt-based competitors. This suggests that the inclusion of additional synthetic training data has a limited impact on enhancing the model's generalization capability across different domains.


\begin{table}[!htb]
    \centering
    \footnotesize
    \begin{tabular}{l@{\hspace{0.7cm}}ccc}
    \toprule
    Method & B & N & HM \\
    \midrule        
    IVLP & 79.06 (84.21) & 65.04 (71.79) & 71.36 (77.51) \\
    + $\mathcal{L}_{SCE}$ & 78.35 (83.95) & 69.34 (75.34) & 73.57 (79.41) \\
    + $\mathcal{L}_{FS}$ & 77.84 (83.91) & 71.04 (77.35) & 74.28 (80.50) \\
    \bottomrule
    \end{tabular}    
    \vspace{-0.2cm}
    \caption{Impacts (\%) of loss functions. Results are averaged over 11 datasets under the GZSL(ZSL) setting.}
    \label{tab: ab-modules}
\end{table}


\textbf{Cross-Dataset Generalization.} This experiment assesses the impacts of the inclusion of additional synthetic data for the base classes on the novel classes from the other datasets. \cref{tab:cross-dataset} shows the results of our method and four competitors. Our model is trained with both the real and synthetic base data. From the results, we observe that our method demonstrates competitive performance, achieving the highest average accuracy at 66.54\%. Furthermore, the analysis reveals that a substantial portion of the performance improvements is primarily attributed to the EuroSAT dataset. We speculate that the reason for this observation lies in the similarity between the domains of the synthetic data from the base classes and the EuroSAT dataset.



\begin{table}[!t]
    \centering
\footnotesize
    \begin{tabular}{lccc}
    \toprule
    Method & B & N & HM \\
    \midrule
    IVLP & 79.06 (84.21) & 65.04 (71.79) & 71.36 (77.51) \\
    ~+ Shared  & 82.20 (82.54)  & 51.39 (72.47)  & 63.24 (77.18)  \\
    ~+ Independent  & 77.82 (83.91) &  68.79 (75.09) &  73.03 (79.26) \\ 
    \midrule
    SYNC-CLIP & 77.84 (83.91) & 71.04 (77.35) & 74.28 (80.50) \\
    \bottomrule
    \end{tabular}
      \vspace{-2mm}
    \caption{Impact (\%) of visual prompt types. Results are averaged over 11 datasets under the GZSL(ZSL) setting.}
    
    \label{tab:domain prompts}
\end{table}

\subsection{Ablation Study \& Analysis}


\textbf{Impacts of Loss Functions.} \cref{tab: ab-modules} shows the ablation of different loss functions. When we incorporate $\mathcal{L}_{SCE}$ into baseline IVLP, the result demonstrates that \textbf{N} metric under both GZSL and ZSL settings gets a remarkable improvement, 4.30\% and 3.55\%, respectively. It indicates that \textbf{proposed domain-specific prompts} leverages synthetic data to compensate for the missing knowledge of novel classes during the matching process between image features and text features. However, there is a slight decrease in \textbf{B} metric, which can be attributed to the difference between the distributions of synthetic and real data. Additionally, \textbf{N} metric has been further improved with the help of $\mathcal{L}_{FS}$. This confirms that SYNC-CLIP reduces the distribution discrepancy between real and synthetic data by aligning their features, allowing the synthetic data feature belonging to the novel classes to better conform to the real data feature. As a result, the classifier receives more reliable information, leading to further improvements in performance.

\textbf{Impacts of Visual Prompt Types.} We compare three types of visual prompts that are illustrated in \cref{tab:domain prompts}. \textbf{Shared} means the visual prompts of synthetic data are the same as the real data. We observe that it is worse than the baseline and indicates the consistency of synthetic data and real data. \textbf{Independent} refers to that we optimize the visual prompts of synthetic data independently. The results show a noticeable improvement in \textbf{N} metric and signify the inclusion of semantic information from the synthetic data within novel classes. Further, when comparing our proposed SYNC-CLIP with \textbf{Independent} prompts, there is a significant lift of 2.25\% and 2.26\% in \textbf{N} metric under the GZSL and ZSL settings, respectively. This indicates the effectiveness of domain prompts in capturing domain-specific information while concurrently conveying the domain-invariant guidance of novel classes to real data, thereby fostering superior model generalization. 



\begin{figure}[!tb]
  \centering
  \captionsetup[subfigure]{justification=centering}
  \begin{subfigure}{0.49\linewidth}
    \includegraphics[width=\linewidth]{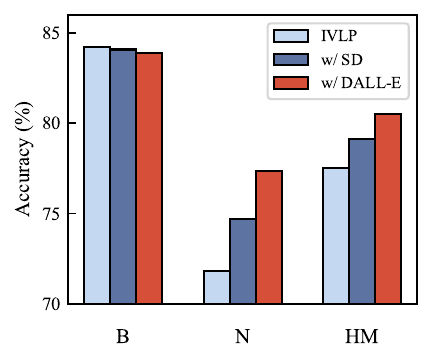}
    \vspace{-0.6cm}
    \caption{\footnotesize Synthetic Models.}
    \label{fig:syn-models}
  \end{subfigure}
  \begin{subfigure}{0.49\linewidth}
  \includegraphics[width=\linewidth]{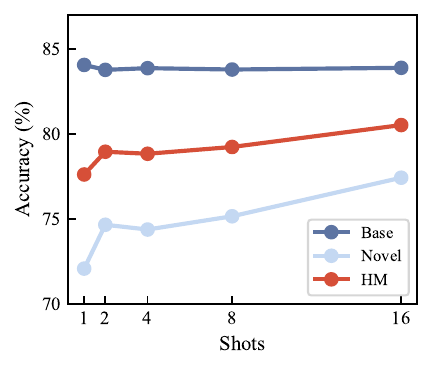}
    \vspace{-0.6cm}
    \caption{\footnotesize Synthetic Amounts.}
    \label{fig:syn-amounts}
  \end{subfigure}
  \vspace{-0.2cm}
  \caption{Impact of synthetic data. Results are averaged over 11 datasets under ZSL setting.}
  \vspace{-2.5mm}
  \label{fig:prompt-syn}
\end{figure}



\begin{table*}[!htb]
    \centering
    \footnotesize
        \begin{tabular}{lcccc|cc}
        \toprule    
        Method  & CLIP \cite{clip} & IVLP \cite{PromptSRC} & IVLP (S) \cite{PromptSRC} & IVLP (R + S) \cite{PromptSRC} & SYNC-CLIP (R + S) & $\Delta$ (Margin)\\
        \midrule
        B & 62.22 (69.34) & 79.06 (84.21) & 56.37 (64.84) & 82.20 (82.54) & 77.84 (83.91) & {\color{blue}-1.22} ({\color{blue}-0.30}) \\
        N & 65.44 (74.22) & 65.04 (71.79) & 62.93 (69.85) & 51.39 (72.47) & 71.04 (77.35) & {\color{red}+6.00} ({\color{red}+5.56}) \\
        HM & 63.79 (71.70) & 71.36 (77.51) & 59.47 (67.25) & 63.24 (77.18) & 74.28 (80.50) & {\color{red}+2.92} ({\color{red}+2.99}) \\
        \bottomrule
        \end{tabular}   
        \vspace{-0.2cm}
    \caption{Comparison results (\%) of our method and the baselines under the GZSL (ZSL) setting, averaging the results across 11 datasets. R means the model trained with real data, and S means the model trained with the synthetic data. $\Delta$ denotes the margin between our method with the best-performing baseline.  }
    \label{tab:synthetic}
\end{table*}

\textbf{Impacts of Synthetic Data.} This experiment assesses the influences of synthetic data including synthetic amounts and synthetic models on the average performance of 11 datasets under the ZSL setting. \cref{fig:syn-models} demonstrates the results of two popular text-to-image models, Stable Diffusion \cite{saharia2022photorealistic} (SD) and DALL-E \cite{dall-e}. We observe that the models with the synthetic data from both SD and DALL-E exhibit improvements over the baseline IVLP in terms of \textbf{N} and \textbf{HM}, indicating the valuable information provided by the synthetic data for novel classification. Additionally, the model enriched with synthetic data from DALL-E outperforms the one with the synthetic data from SD, particularly in terms of the \textbf{N} metric, demonstrating that the synthetic data from DALL-E provides more valuable information for the novel classes. Then we select the DALL-E as the generative model for evaluating the effects of synthetic amounts, as illustrated in \cref{fig:syn-amounts}. From the results, we observe that as the number of generated samples increases, the performance of base classes slightly decreases, but the performance of novel classes gradually improves, leading to an enhancement in the HM metric performance. This indicates that synthesizing more data could focus more on novel classes and rebalance the decision boundaries.

\textbf{Impacts of Fine-tuning Models.} To comprehensively assess the effectiveness of the models with the synthetic data, this experiment compares the baseline and our proposed methods, as illustrated in \cref{tab:synthetic}. The results yield the following observations. Firstly, fine-tuning the model with the real base data indeed significantly improves \textbf{B} but has a slight negative impact on \textbf{N}, when comparing the baseline IVLP and the pre-trained CLIP. Secondly, training the IVLP solely with synthetic data results in a significant degradation in both \textbf{B} and \textbf{N} compared to the pre-trained CLIP. Even when incorporating real data, the performance remains inferior to the IVLP trained exclusively with real data. Note that training with the real data and synthetic data here implies that both share the same visual prompts. This suggests that directly combining synthetic data negatively impacts the learning of generalization patterns and the effect of our decoupled visual prompts. In comparison, although our method sacrifices a small portion of the accuracy on base classes, it significantly boosts the accuracy on novel classes, leading to a better \textbf{HM} metric. This further verifies that only a well-designed model could exploit the valuable information for the novel classes.



\begin{figure}[!t]
  \centering
  \begin{overpic}[width=1\linewidth]{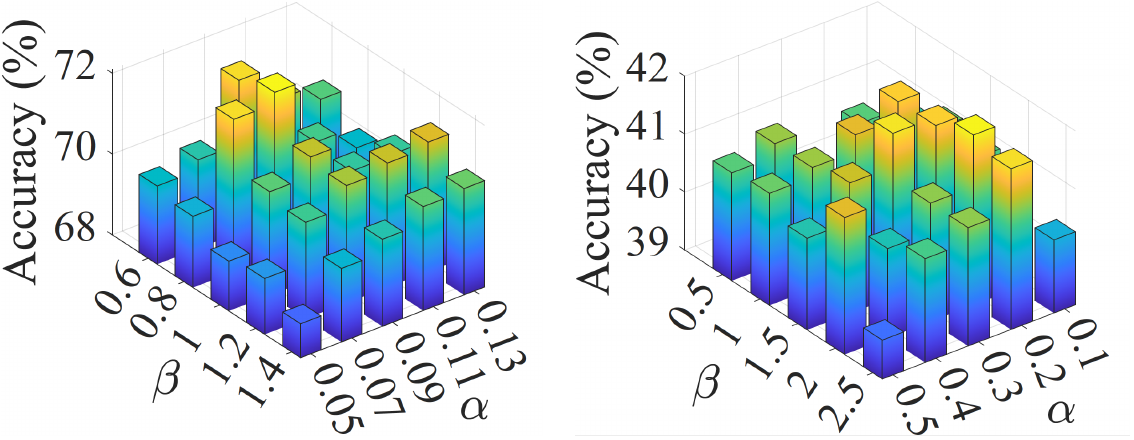}
  \put(20,-4){\footnotesize (a) DTD}
  \put(70,-4){\footnotesize (b) Aircraft}
  \end{overpic}
  \vspace{-3.0mm}  
  \caption{Hyperparameter sensitivity for $\alpha$ and $\beta$ on both DTD and Aircraft datasets. }
  \vspace{-2.5mm}
\label{fig:hyperparameter}
\end{figure}

\textbf{Impacts of Hyper-parameters.} We conduct an ablation study on our hyperparameters $\alpha$ and $\beta$ using the DTD and Aircraft datasets, as presented in \cref{fig:hyperparameter}. Notably, we observe variations in the optimal values for $\alpha$ and $\beta$ across different datasets. This discrepancy suggests that the effectiveness of synthetic data may vary depending on the dataset characteristics. Specifically, the Aircraft dataset exhibits a larger variance in accuracy values, attributed to its fine-grained nature. For more hyper-parameters ablations, please refer to the Appendix.

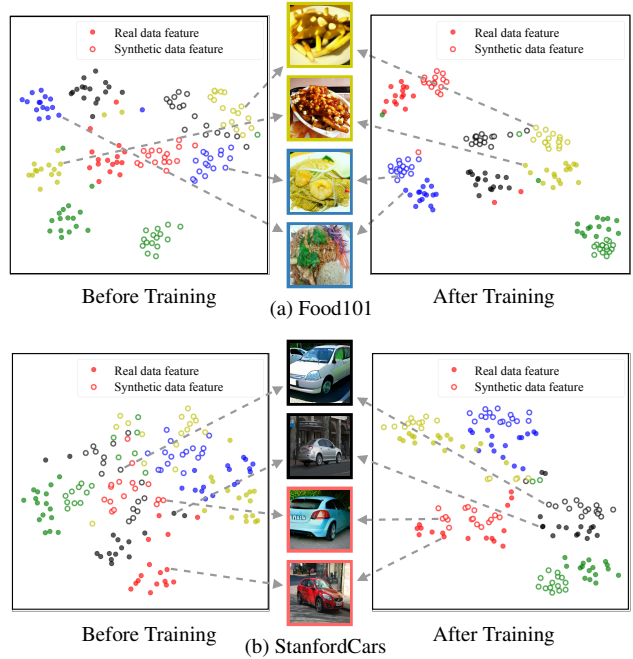
\begin{figure}[!t]
  \centering
  \begin{overpic}[width=1\linewidth]{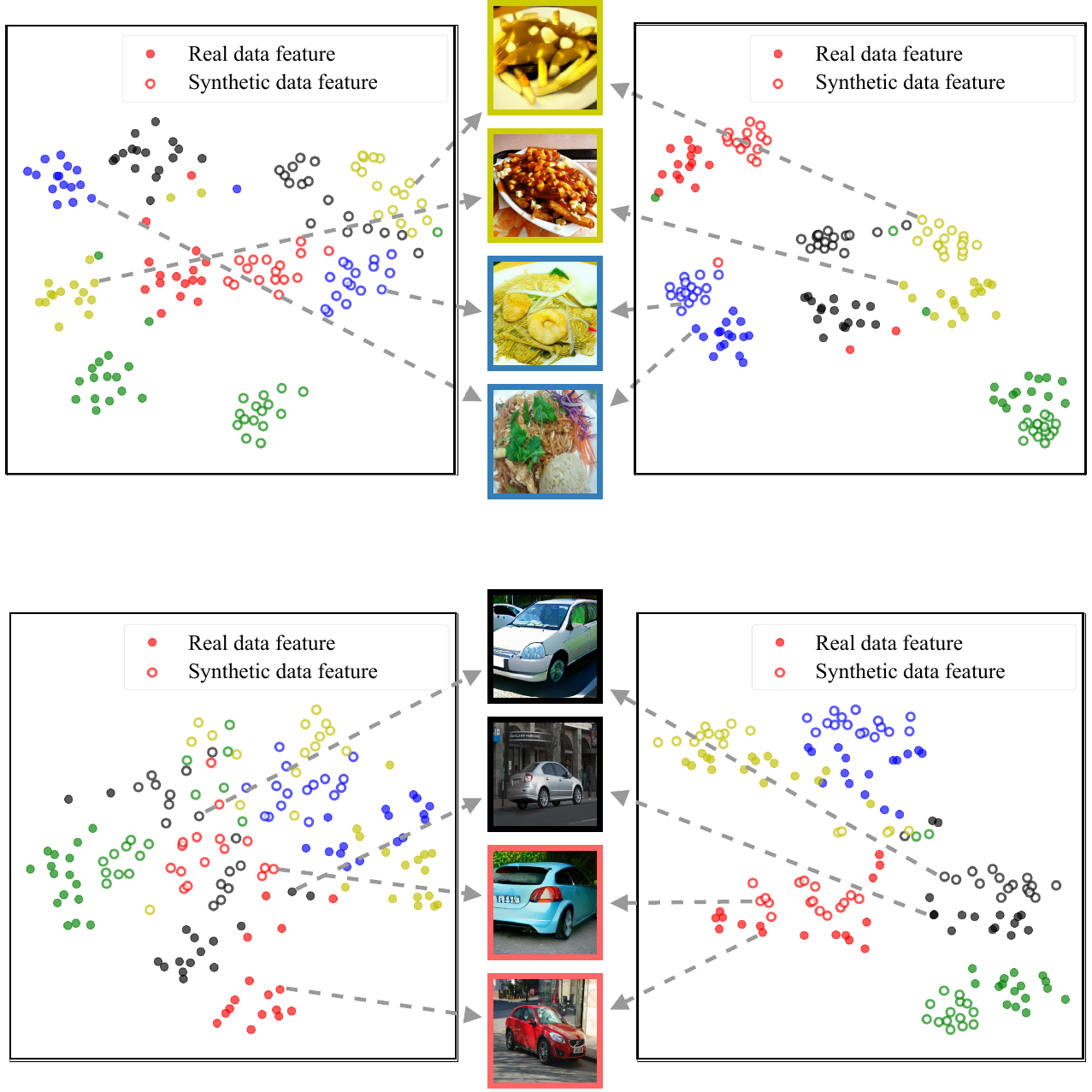}
  \put(12,52){\footnotesize Before Training}
  \put(12,-2){\footnotesize Before Training}
  \put(68,52){\footnotesize After Training}
  \put(42,50){\footnotesize (a) Food101}
  \put(68,-2){\footnotesize After Training}
  \put(38,-4){\footnotesize (b) StanfordCars}
  
  \end{overpic}
  \vspace{-2.0mm}  
  \caption{t-SNE results on Food101 and StanfordCars datasets.}
  \vspace{-2.5mm}
\label{fig:compared_tsne}
\end{figure}

\textbf{Visualization Results.} \cref{fig:compared_tsne} illustrates the changes in the t-SNE distribution before and after training. Before training, a notable disparity exists between the synthetic distribution and the real distribution. After training, alignment between the synthetic and real distributions is achieved, facilitating classification in downstream tasks and expanding the utility of synthetic data.


\section{Conclusion}

In this paper, we have introduced SYNC-CLIP, an innovative approach designed to facilitate the adaption of CLIP to downstream tasks, particularly in data-limited scenarios. By designing divided domain prompts, SYNC-CLIP leverages the synthetic data to alleviate the imbalance issues that current prompt learning methods commonly encounter. Additionally, through cross-domain feature alignment, SYNC-CLIP imparts implicit guidance for open-vocabulary decision boundaries. Experimental results across diverse benchmarks consistently showcase that SYNC-CLIP enhances generalization capabilities and achieves significant improvements, especially in handling novel classes.






{
    \small
    \bibliographystyle{ieeenat_fullname}
    \bibliography{main}

\begin{thebibliography}{52}
\providecommand{\natexlab}[1]{#1}
\providecommand{\url}[1]{\texttt{#1}}
\expandafter\ifx\csname urlstyle\endcsname\relax
  \providecommand{\doi}[1]{doi: #1}\else
  \providecommand{\doi}{doi: \begingroup \urlstyle{rm}\Url}\fi

\bibitem[Alayrac et~al.(2022)Alayrac, Donahue, Luc, Miech, Barr, Hasson, Lenc, Mensch, Millican, Reynolds, et~al.]{alayrac2022flamingo}
Jean-Baptiste Alayrac, Jeff Donahue, Pauline Luc, Antoine Miech, Iain Barr, Yana Hasson, Karel Lenc, Arthur Mensch, Katherine Millican, Malcolm Reynolds, et~al.
\newblock Flamingo: a visual language model for few-shot learning.
\newblock In \emph{NeurIPS}, pages 23716--23736, 2022.

\bibitem[Besnier et~al.(2020)Besnier, Jain, Bursuc, Cord, and Perez]{2020conditionGAN}
Victor Besnier, Himalaya Jain, Andrei Bursuc, Matthieu Cord, and Patrick Perez.
\newblock This dataset does not exist: training models from generated images.
\newblock In \emph{ICASSP}, pages 1--5, 2020.

\bibitem[Bossard et~al.(2014)Bossard, Guillaumin, and Van~Gool]{food101}
Lukas Bossard, Matthieu Guillaumin, and Luc Van~Gool.
\newblock Food-101--mining discriminative components with random forests.
\newblock In \emph{ECCV}, pages 446--461, 2014.

\bibitem[Cimpoi et~al.(2014)Cimpoi, Maji, Kokkinos, Mohamed, and Vedaldi]{dtd}
Mircea Cimpoi, Subhransu Maji, Iasonas Kokkinos, Sammy Mohamed, and Andrea Vedaldi.
\newblock Describing textures in the wild.
\newblock In \emph{CVPR}, pages 3606--3613, 2014.

\bibitem[Deng et~al.(2009)Deng, Dong, Socher, Li, Li, and Fei-Fei]{ImageNet}
Jia Deng, Wei Dong, Richard Socher, Li-Jia Li, Kai Li, and Li Fei-Fei.
\newblock Imagenet: A large-scale hierarchical image database.
\newblock In \emph{CVPR}, pages 248--255, 2009.

\bibitem[Dosovitskiy et~al.(2020)Dosovitskiy, Beyer, Kolesnikov, Weissenborn, Zhai, Unterthiner, Dehghani, Minderer, Heigold, Gelly, et~al.]{vit}
Alexey Dosovitskiy, Lucas Beyer, Alexander Kolesnikov, Dirk Weissenborn, Xiaohua Zhai, Thomas Unterthiner, Mostafa Dehghani, Matthias Minderer, Georg Heigold, Sylvain Gelly, et~al.
\newblock An image is worth 16x16 words: Transformers for image recognition at scale.
\newblock \emph{arXiv:2010.11929}, 2020.

\bibitem[Fei-Fei et~al.(2004)Fei-Fei, Fergus, and Perona]{caltech-101}
Li Fei-Fei, Rob Fergus, and Pietro Perona.
\newblock Learning generative visual models from few training examples: An incremental bayesian approach tested on 101 object categories.
\newblock In \emph{CVPR}, pages 178--178, 2004.

\bibitem[Gao et~al.(2021)Gao, Geng, Zhang, Ma, Fang, Zhang, Li, and Qiao]{clip-adapter}
Peng Gao, Shijie Geng, Renrui Zhang, Teli Ma, Rongyao Fang, Yongfeng Zhang, Hongsheng Li, and Yu Qiao.
\newblock Clip-adapter: Better vision-language models with feature adapters.
\newblock \emph{arXiv:2110.04544}, 2021.

\bibitem[Goodfellow et~al.(2014)Goodfellow, Pouget-Abadie, Mirza, Xu, Warde-Farley, Ozair, Courville, and Bengio]{gan}
Ian Goodfellow, Jean Pouget-Abadie, Mehdi Mirza, Bing Xu, David Warde-Farley, Sherjil Ozair, Aaron Courville, and Yoshua Bengio.
\newblock Generative adversarial nets.
\newblock In \emph{NeurIPS}, 2014.

\bibitem[Gu et~al.(2021)Gu, Lin, Kuo, and Cui]{gu2021open}
Xiuye Gu, Tsung-Yi Lin, Weicheng Kuo, and Yin Cui.
\newblock Open-vocabulary object detection via vision and language knowledge distillation.
\newblock \emph{arXiv:2104.13921}, 2021.

\bibitem[He et~al.(2023)He, Sun, Yu, Xue, Zhang, Torr, Bai, and Qi]{he2022synthetic}
Ruifei He, Shuyang Sun, Xin Yu, Chuhui Xue, Wenqing Zhang, Philip Torr, Song Bai, and Xiaojuan Qi.
\newblock Is synthetic data from generative models ready for image recognition?
\newblock In \emph{ICLR}, 2023.

\bibitem[Helber et~al.(2019)Helber, Bischke, Dengel, and Borth]{eurosat}
Patrick Helber, Benjamin Bischke, Andreas Dengel, and Damian Borth.
\newblock Eurosat: A novel dataset and deep learning benchmark for land use and land cover classification.
\newblock \emph{IEEE Journal of Selected Topics in Applied Earth Observations and Remote Sensing}, 12\penalty0 (7):\penalty0 2217--2226, 2019.

\bibitem[Hendrycks et~al.(2021{\natexlab{a}})Hendrycks, Basart, Mu, Kadavath, Wang, Dorundo, Desai, Zhu, Parajuli, Guo, et~al.]{imagenet-r}
Dan Hendrycks, Steven Basart, Norman Mu, Saurav Kadavath, Frank Wang, Evan Dorundo, Rahul Desai, Tyler Zhu, Samyak Parajuli, Mike Guo, et~al.
\newblock The many faces of robustness: A critical analysis of out-of-distribution generalization.
\newblock In \emph{ICCV}, pages 8340--8349, 2021{\natexlab{a}}.

\bibitem[Hendrycks et~al.(2021{\natexlab{b}})Hendrycks, Zhao, Basart, Steinhardt, and Song]{imagenet-a}
Dan Hendrycks, Kevin Zhao, Steven Basart, Jacob Steinhardt, and Dawn Song.
\newblock Natural adversarial examples.
\newblock In \emph{CVPR}, pages 15262--15271, 2021{\natexlab{b}}.

\bibitem[Ho et~al.(2020)Ho, Jain, and Abbeel]{ddpm}
Jonathan Ho, Ajay Jain, and Pieter Abbeel.
\newblock Denoising diffusion probabilistic models.
\newblock In \emph{NeurIPS}, pages 6840--6851, 2020.

\bibitem[Houlsby et~al.(2019)Houlsby, Giurgiu, Jastrzebski, Morrone, De~Laroussilhe, Gesmundo, Attariyan, and Gelly]{houlsby2019parameter}
Neil Houlsby, Andrei Giurgiu, Stanislaw Jastrzebski, Bruna Morrone, Quentin De~Laroussilhe, Andrea Gesmundo, Mona Attariyan, and Sylvain Gelly.
\newblock Parameter-efficient transfer learning for nlp.
\newblock In \emph{ICML}, pages 2790--2799, 2019.

\bibitem[Huang et~al.(2023)Huang, Li, Feng, Wu, Sun, and Ji]{huang2023clover}
Jingjia Huang, Yinan Li, Jiashi Feng, Xinglong Wu, Xiaoshuai Sun, and Rongrong Ji.
\newblock Clover: Towards a unified video-language alignment and fusion model.
\newblock In \emph{CVPR}, pages 14856--14866, 2023.

\bibitem[Jahanian et~al.(2021)Jahanian, Puig, Tian, and Isola]{jahanian2021generative}
Ali Jahanian, Xavier Puig, Yonglong Tian, and Phillip Isola.
\newblock Generative models as a data source for multiview representation learning.
\newblock \emph{arXiv:2106.05258}, 2021.

\bibitem[Jia et~al.(2021)Jia, Yang, Xia, Chen, Parekh, Pham, Le, Sung, Li, and Duerig]{ALIGN}
Chao Jia, Yinfei Yang, Ye Xia, Yi-Ting Chen, Zarana Parekh, Hieu Pham, Quoc Le, Yun-Hsuan Sung, Zhen Li, and Tom Duerig.
\newblock Scaling up visual and vision-language representation learning with noisy text supervision.
\newblock In \emph{ICML}, pages 4904--4916, 2021.

\bibitem[Jia et~al.(2022)Jia, Tang, Chen, Cardie, Belongie, Hariharan, and Lim]{vpt}
Menglin Jia, Luming Tang, Bor-Chun Chen, Claire Cardie, Serge Belongie, Bharath Hariharan, and Ser-Nam Lim.
\newblock Visual prompt tuning.
\newblock In \emph{ECCV}, pages 709--727, 2022.

\bibitem[Karras et~al.(2019)Karras, Laine, and Aila]{stylegan}
Tero Karras, Samuli Laine, and Timo Aila.
\newblock A style-based generator architecture for generative adversarial networks.
\newblock In \emph{CVPR}, pages 4401--4410, 2019.

\bibitem[Khattak et~al.(2023{\natexlab{a}})Khattak, Rasheed, Maaz, Khan, and Khan]{maple}
Muhammad~Uzair Khattak, Hanoona Rasheed, Muhammad Maaz, Salman Khan, and Fahad~Shahbaz Khan.
\newblock Maple: Multi-modal prompt learning.
\newblock In \emph{CVPR}, pages 19113--19122, 2023{\natexlab{a}}.

\bibitem[Khattak et~al.(2023{\natexlab{b}})Khattak, Wasim, Naseer, Khan, Yang, and Khan]{PromptSRC}
Muhammad~Uzair Khattak, Syed~Talal Wasim, Muzammal Naseer, Salman Khan, Ming-Hsuan Yang, and Fahad~Shahbaz Khan.
\newblock Self-regulating prompts: Foundational model adaptation without forgetting.
\newblock In \emph{ICCV}, pages 15190--15200, 2023{\natexlab{b}}.

\bibitem[Kingma and Welling(2013)]{vae}
Diederik~P Kingma and Max Welling.
\newblock Auto-encoding variational bayes.
\newblock \emph{arXiv:1312.6114}, 2013.

\bibitem[Kirillov et~al.(2023)Kirillov, Mintun, Ravi, Mao, Rolland, Gustafson, Xiao, Whitehead, Berg, Lo, et~al.]{sam}
Alexander Kirillov, Eric Mintun, Nikhila Ravi, Hanzi Mao, Chloe Rolland, Laura Gustafson, Tete Xiao, Spencer Whitehead, Alexander~C Berg, Wan-Yen Lo, et~al.
\newblock Segment anything.
\newblock pages 4015--4026, 2023.

\bibitem[Krause et~al.(2013)Krause, Stark, Deng, and Fei-Fei]{cars}
Jonathan Krause, Michael Stark, Jia Deng, and Li Fei-Fei.
\newblock 3d object representations for fine-grained categorization.
\newblock In \emph{ICCV}, pages 554--561, 2013.

\bibitem[Liu et~al.(2023)Liu, Yuan, Fu, Jiang, Hayashi, and Neubig]{liu2023pre}
Pengfei Liu, Weizhe Yuan, Jinlan Fu, Zhengbao Jiang, Hiroaki Hayashi, and Graham Neubig.
\newblock Pre-train, prompt, and predict: A systematic survey of prompting methods in natural language processing.
\newblock \emph{ACM Computing Surveys}, 55\penalty0 (9):\penalty0 1--35, 2023.

\bibitem[Maji et~al.(2013)Maji, Rahtu, Kannala, Blaschko, and Vedaldi]{FGVCAircraft}
Subhransu Maji, Esa Rahtu, Juho Kannala, Matthew Blaschko, and Andrea Vedaldi.
\newblock Fine-grained visual classification of aircraft.
\newblock \emph{arXiv:1306.5151}, 2013.

\bibitem[Nichol et~al.(2021)Nichol, Dhariwal, Ramesh, Shyam, Mishkin, McGrew, Sutskever, and Chen]{glide}
Alex Nichol, Prafulla Dhariwal, Aditya Ramesh, Pranav Shyam, Pamela Mishkin, Bob McGrew, Ilya Sutskever, and Mark Chen.
\newblock Glide: Towards photorealistic image generation and editing with text-guided diffusion models.
\newblock \emph{arXiv:2112.10741}, 2021.

\bibitem[Nilsback and Zisserman(2008)]{Flowers}
Maria-Elena Nilsback and Andrew Zisserman.
\newblock Automated flower classification over a large number of classes.
\newblock In \emph{Indian Conference on Computer Vision, Graphics \& Image Processing}, pages 722--729, 2008.

\bibitem[Parkhi et~al.(2012)Parkhi, Vedaldi, Zisserman, and Jawahar]{OxfordPets}
Omkar~M Parkhi, Andrea Vedaldi, Andrew Zisserman, and CV Jawahar.
\newblock Cats and dogs.
\newblock In \emph{CVPR}, pages 3498--3505, 2012.

\bibitem[Pratt et~al.(2023)Pratt, Covert, Liu, and Farhadi]{cupl}
Sarah Pratt, Ian Covert, Rosanne Liu, and Ali Farhadi.
\newblock What does a platypus look like? generating customized prompts for zero-shot image classification.
\newblock In \emph{ICCV}, pages 15691--15701, 2023.

\bibitem[Radford et~al.(2021)Radford, Kim, Hallacy, Ramesh, Goh, Agarwal, Sastry, Askell, Mishkin, Clark, et~al.]{clip}
Alec Radford, Jong~Wook Kim, Chris Hallacy, Aditya Ramesh, Gabriel Goh, Sandhini Agarwal, Girish Sastry, Amanda Askell, Pamela Mishkin, Jack Clark, et~al.
\newblock Learning transferable visual models from natural language supervision.
\newblock In \emph{ICML}, pages 8748--8763, 2021.

\bibitem[Ramesh et~al.(2021)Ramesh, Pavlov, Goh, Gray, Voss, Radford, Chen, and Sutskever]{dall-e}
Aditya Ramesh, Mikhail Pavlov, Gabriel Goh, Scott Gray, Chelsea Voss, Alec Radford, Mark Chen, and Ilya Sutskever.
\newblock Zero-shot text-to-image generation.
\newblock In \emph{ICML}, pages 8821--8831, 2021.

\bibitem[Rao et~al.(2022)Rao, Zhao, Chen, Tang, Zhu, Huang, Zhou, and Lu]{denseclip}
Yongming Rao, Wenliang Zhao, Guangyi Chen, Yansong Tang, Zheng Zhu, Guan Huang, Jie Zhou, and Jiwen Lu.
\newblock Denseclip: Language-guided dense prediction with context-aware prompting.
\newblock In \emph{CVPR}, pages 18082--18091, 2022.

\bibitem[Recht et~al.(2019)Recht, Roelofs, Schmidt, and Shankar]{ImageNetV2}
Benjamin Recht, Rebecca Roelofs, Ludwig Schmidt, and Vaishaal Shankar.
\newblock Do imagenet classifiers generalize to imagenet?
\newblock In \emph{ICML}, pages 5389--5400, 2019.

\bibitem[Saharia et~al.(2022)Saharia, Chan, Saxena, Li, Whang, Denton, Ghasemipour, Gontijo~Lopes, Karagol~Ayan, Salimans, et~al.]{saharia2022photorealistic}
Chitwan Saharia, William Chan, Saurabh Saxena, Lala Li, Jay Whang, Emily~L Denton, Kamyar Ghasemipour, Raphael Gontijo~Lopes, Burcu Karagol~Ayan, Tim Salimans, et~al.
\newblock Photorealistic text-to-image diffusion models with deep language understanding.
\newblock pages 36479--36494, 2022.

\bibitem[Shu et~al.(2022)Shu, Nie, Huang, Yu, Goldstein, Anandkumar, and Xiao]{shu2022test}
Manli Shu, Weili Nie, De-An Huang, Zhiding Yu, Tom Goldstein, Anima Anandkumar, and Chaowei Xiao.
\newblock Test-time prompt tuning for zero-shot generalization in vision-language models.
\newblock pages 14274--14289, 2022.

\bibitem[Sohl-Dickstein et~al.(2015)Sohl-Dickstein, Weiss, Maheswaranathan, and Ganguli]{ddpm2015}
Jascha Sohl-Dickstein, Eric Weiss, Niru Maheswaranathan, and Surya Ganguli.
\newblock Deep unsupervised learning using nonequilibrium thermodynamics.
\newblock In \emph{ICML}, pages 2256--2265, 2015.

\bibitem[Soomro et~al.(2012)Soomro, Zamir, and Shah]{ucf101}
Khurram Soomro, Amir~Roshan Zamir, and Mubarak Shah.
\newblock Ucf101: A dataset of 101 human actions classes from videos in the wild.
\newblock \emph{arXiv:1212.0402}, 2012.

\bibitem[Udandarao et~al.(2022)Udandarao, Gupta, and Albanie]{udandarao2022sus}
Vishaal Udandarao, Ankush Gupta, and Samuel Albanie.
\newblock Sus-x: Training-free name-only transfer of vision-language models.
\newblock \emph{arXiv:2211.16198}, 2022.

\bibitem[Wang et~al.(2019)Wang, Ge, Lipton, and Xing]{imagenet-sketch}
Haohan Wang, Songwei Ge, Zachary Lipton, and Eric~P Xing.
\newblock Learning robust global representations by penalizing local predictive power.
\newblock In \emph{NeurIPS}, 2019.

\bibitem[Wang et~al.(2022)Wang, Bao, Dong, Bjorck, Peng, Liu, Aggarwal, Mohammed, Singhal, Som, et~al.]{wang2022beit}
Wenhui Wang, Hangbo Bao, Li Dong, Johan Bjorck, Zhiliang Peng, Qiang Liu, Kriti Aggarwal, Owais~Khan Mohammed, Saksham Singhal, Subhojit Som, et~al.
\newblock Image as a foreign language: Beit pretraining for all vision and vision-language tasks.
\newblock \emph{arXiv:2208.10442}, 2022.

\bibitem[Wang et~al.(2021)Wang, Yu, Yu, Dai, Tsvetkov, and Cao]{wang2021simvlm}
Zirui Wang, Jiahui Yu, Adams~Wei Yu, Zihang Dai, Yulia Tsvetkov, and Yuan Cao.
\newblock Simvlm: Simple visual language model pretraining with weak supervision.
\newblock \emph{arXiv:2108.10904}, 2021.

\bibitem[Wasim et~al.(2023)Wasim, Naseer, Khan, Khan, and Shah]{wasim2023vita}
Syed~Talal Wasim, Muzammal Naseer, Salman Khan, Fahad~Shahbaz Khan, and Mubarak Shah.
\newblock Vita-clip: Video and text adaptive clip via multimodal prompting.
\newblock In \emph{CVPR}, pages 23034--23044, 2023.

\bibitem[Xiao et~al.(2010)Xiao, Hays, Ehinger, Oliva, and Torralba]{sun397}
Jianxiong Xiao, James Hays, Krista~A Ehinger, Aude Oliva, and Antonio Torralba.
\newblock Sun database: Large-scale scene recognition from abbey to zoo.
\newblock In \emph{CVPR}, pages 3485--3492, 2010.

\bibitem[Yao et~al.(2023)Yao, Zhang, and Xu]{kgcoop}
Hantao Yao, Rui Zhang, and Changsheng Xu.
\newblock Visual-language prompt tuning with knowledge-guided context optimization.
\newblock In \emph{CVPR}, pages 6757--6767, 2023.

\bibitem[Zhang et~al.(2022)Zhang, Fang, Zhang, Gao, Li, Dai, Qiao, and Li]{tip-adapter}
Renrui Zhang, Rongyao Fang, Wei Zhang, Peng Gao, Kunchang Li, Jifeng Dai, Yu Qiao, and Hongsheng Li.
\newblock Tip-adapter: Training-free clip-adapter for better vision-language modeling.
\newblock In \emph{ECCV}, pages 493--510, 2022.

\bibitem[Zhang et~al.(2023)Zhang, Hu, Li, Huang, Deng, Qiao, Gao, and Li]{cafo}
Renrui Zhang, Xiangfei Hu, Bohao Li, Siyuan Huang, Hanqiu Deng, Yu Qiao, Peng Gao, and Hongsheng Li.
\newblock Prompt, generate, then cache: Cascade of foundation models makes strong few-shot learners.
\newblock In \emph{CVPR}, pages 15211--15222, 2023.

\bibitem[Zhang et~al.(2021)Zhang, Ling, Gao, Yin, Lafleche, Barriuso, Torralba, and Fidler]{zhang2021datasetgan}
Yuxuan Zhang, Huan Ling, Jun Gao, Kangxue Yin, Jean-Francois Lafleche, Adela Barriuso, Antonio Torralba, and Sanja Fidler.
\newblock Datasetgan: Efficient labeled data factory with minimal human effort.
\newblock In \emph{CVPR}, pages 10145--10155, 2021.

\bibitem[Zhou et~al.(2022{\natexlab{a}})Zhou, Yang, Loy, and Liu]{cocoop}
Kaiyang Zhou, Jingkang Yang, Chen~Change Loy, and Ziwei Liu.
\newblock Conditional prompt learning for vision-language models.
\newblock In \emph{CVPR}, pages 16816--16825, 2022{\natexlab{a}}.

\bibitem[Zhou et~al.(2022{\natexlab{b}})Zhou, Yang, Loy, and Liu]{coop}
Kaiyang Zhou, Jingkang Yang, Chen~Change Loy, and Ziwei Liu.
\newblock Learning to prompt for vision-language models.
\newblock \emph{IJCV}, 130\penalty0 (9):\penalty0 2337--2348, 2022{\natexlab{b}}.

\end{thebibliography}
}
\newpage
\newpage
\appendix
\section{Appendix}
This section contains supplementary material that provides
additional details and further experimental
analysis. The content of this section is as follows:

\begin{itemize}
    \item Additional Experimental
    \item Additional Synthetic Data Analysis
\end{itemize}

\subsection{Additional Experimental Details}
\textbf{Competitors}
We compare the proposed approach with the related competitors, i.e., CLIP, CoOp, CoCoOp, MaPLe, and PromptSRC. The details of competitors are as follows: 
\begin{itemize}
    \item \textbf{CLIP} \cite{clip} is a vision model trained on a web-scale dataset of 400 million examples, showcasing exceptional zero-shot reasoning capability and robust generalization. Comprising both an image encoder and a text encoder, CLIP undergoes joint training through a contrastive pre-training process.
    \item \textbf{CoOp} \cite{coop} employs prompt engineering to tailor a vision-language model, such as CLIP, for downstream tasks. This is achieved by seamlessly incorporating learnable context to construct the prompt.
    \item \textbf{CoCoOp} introduces a lightweight network structure based on CoOp to generate an input-specific token which helps the model overcome the overfitting issue.

    \item \textbf{MaPLe} \cite{maple} innovatively incorporates stage-wise text prompts and vision prompts into both the text and image encoders of CLIP. This enhancement is designed to achieve improved alignment in the vision-language representations of the model. Additionally, the approach introduces a coupling function to ensure effective synergy between the two modalities.
    
    \item \textbf{PromptSRC} \cite{PromptSRC} employs self-regularization techniques on both images and text, as well as prompt ensemble and diverse textural prompts. These strategies are integrated to regulate the learnable prompts, effectively addressing overfitting concerns.
\end{itemize}

\textbf{Dataset Details.} In \cref{tab:detail-datasets}, we list the details of the datasets and the hand-crafted prompt we used in the experiments. The prompts are from the \cite{clip} and we have not adopted more prompt templates to generate the optical text representations.  In this work, we only focus on the effect of synthetic data and the text representations would
be automatically learned during the training.

\begin{table*}
    \centering
    \begin{tabular}{cccc}
         \includegraphics[width=0.20\linewidth]{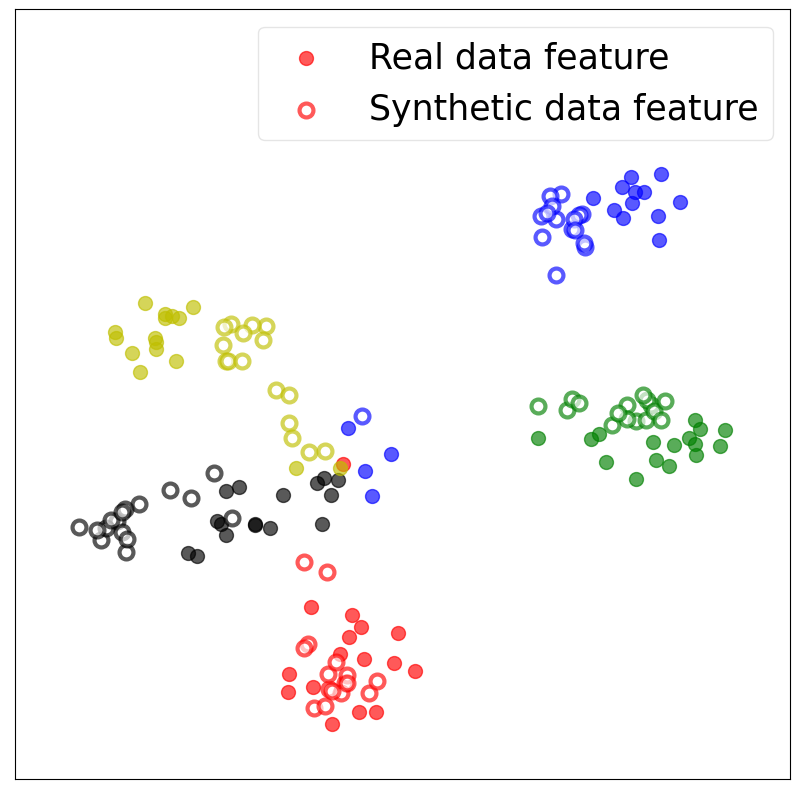} & \includegraphics[width=0.20\linewidth]{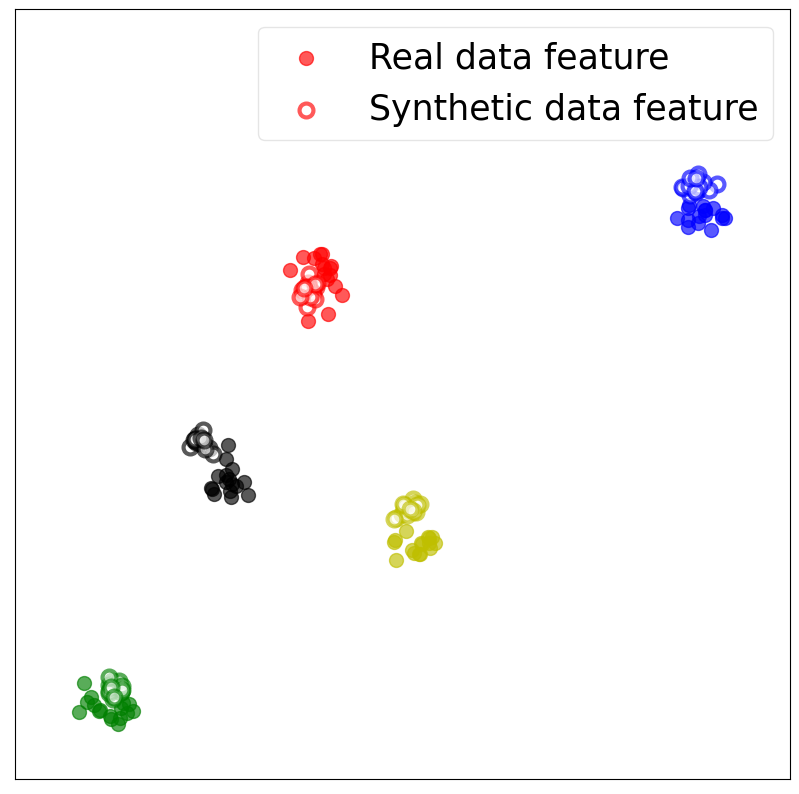} & \includegraphics[width=0.20\linewidth]{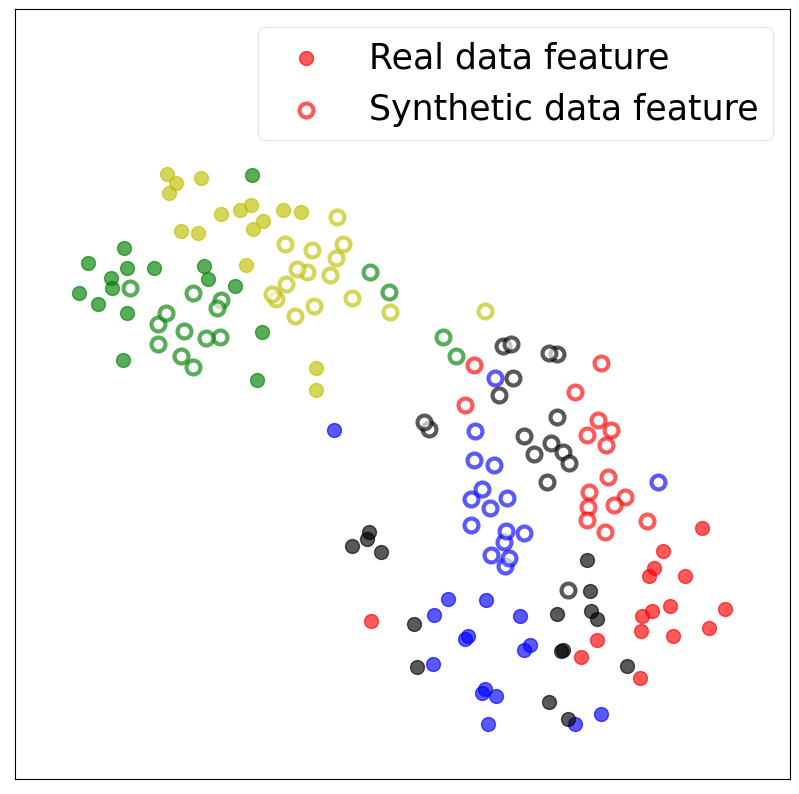} & \includegraphics[width=0.20\linewidth]{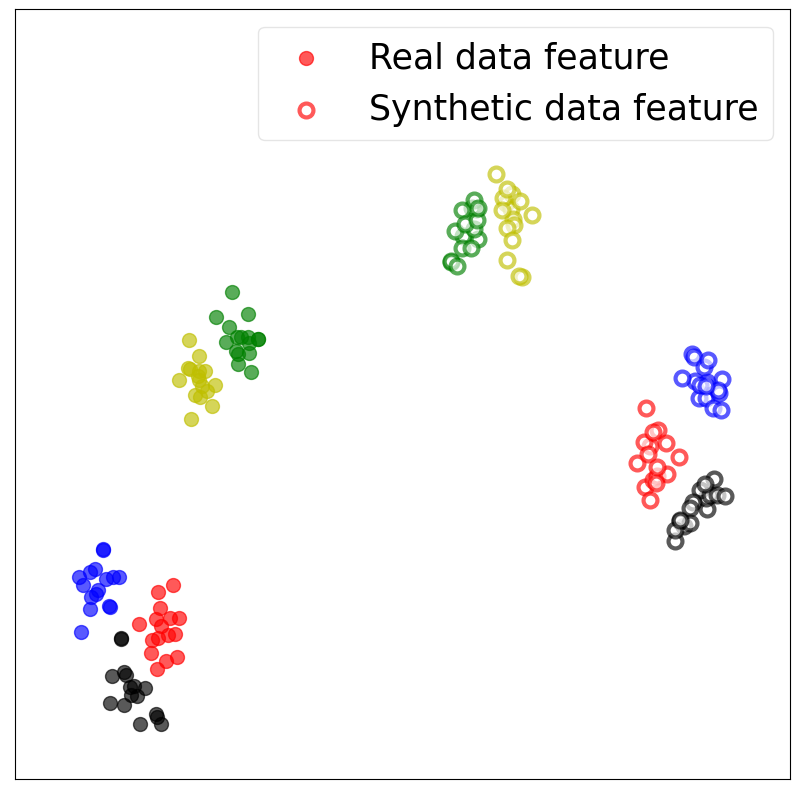}  \\
         \footnotesize Before Training & \footnotesize After Training & \footnotesize Before Training & \footnotesize After Training \\
         \multicolumn{2}{c}{(a) Caltech101} & \multicolumn{2}{c}{(b) OxfordPets} \\

         \includegraphics[width=0.20\linewidth]{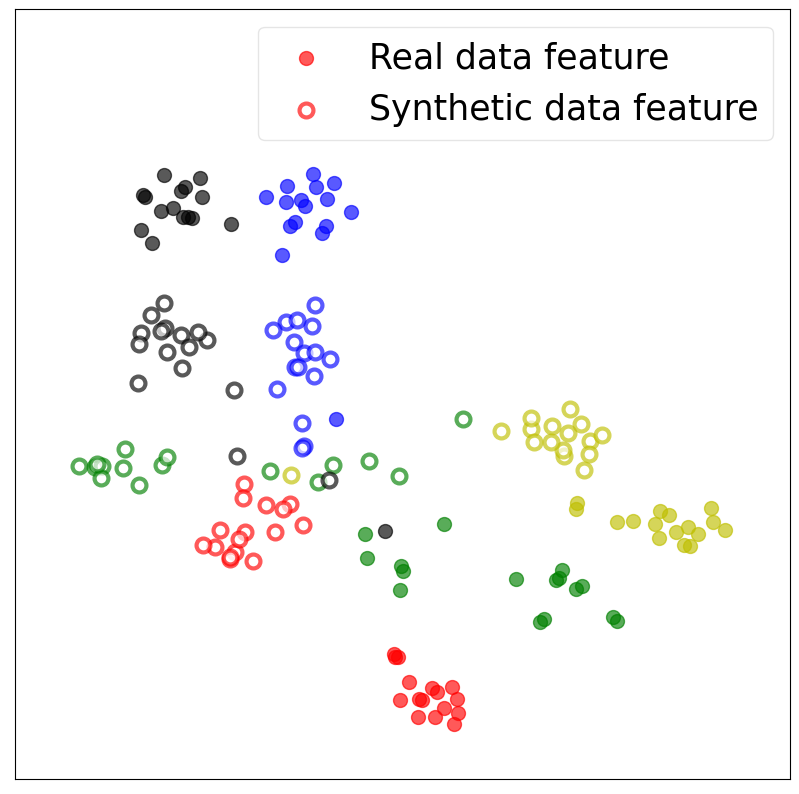} & \includegraphics[width=0.20\linewidth]{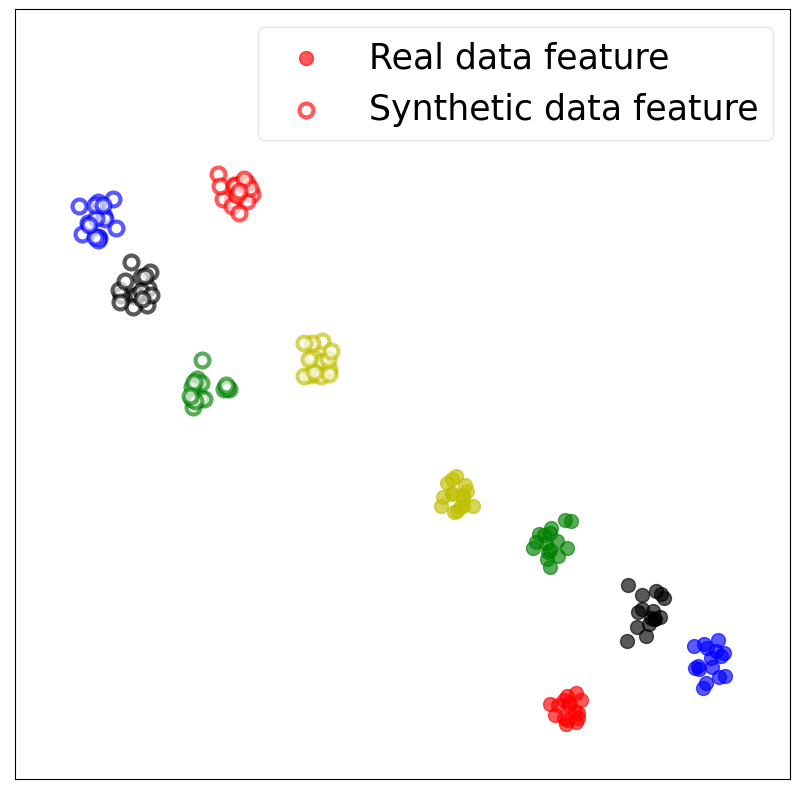} & \includegraphics[width=0.20\linewidth]{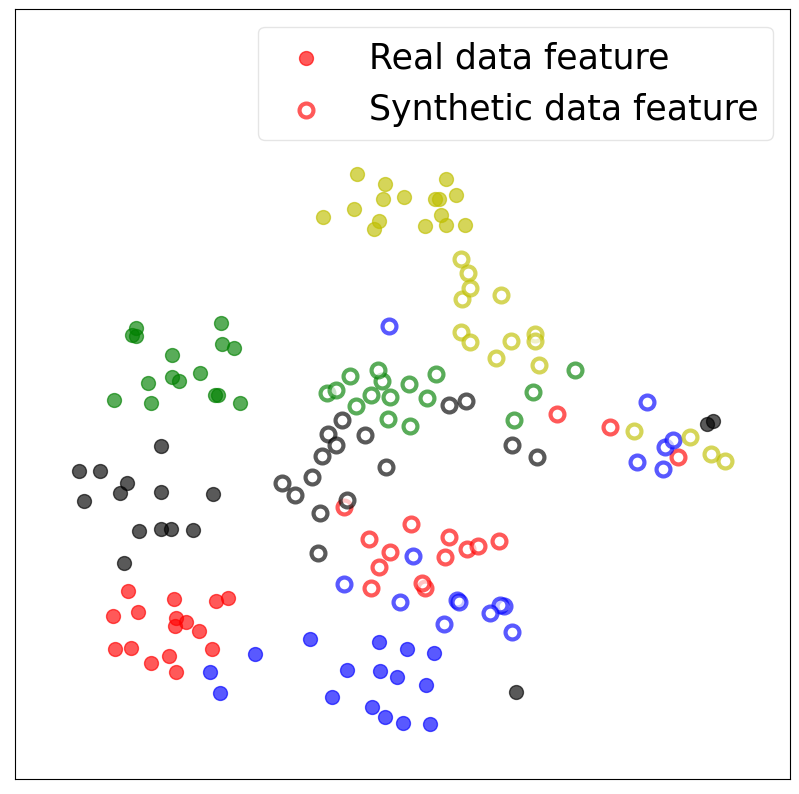} & \includegraphics[width=0.20\linewidth]{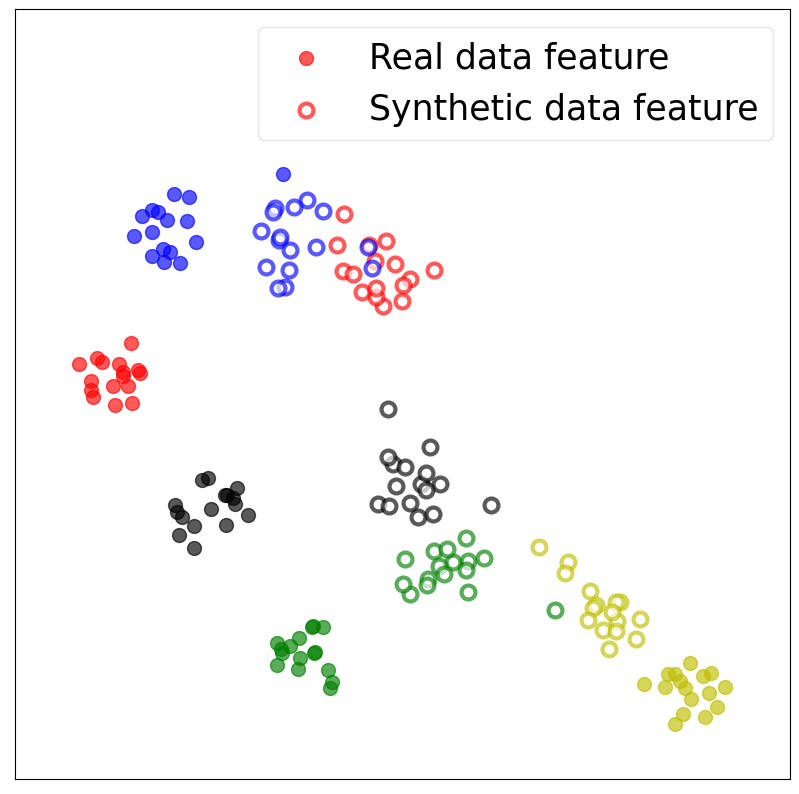}  \\
         \footnotesize Before Training & \footnotesize After Training & \footnotesize Before Training & \footnotesize After Training \\         \multicolumn{2}{c}{(c) Flowers102} & \multicolumn{2}{c}{(d) FGVCAircraft} \\

         \includegraphics[width=0.20\linewidth]{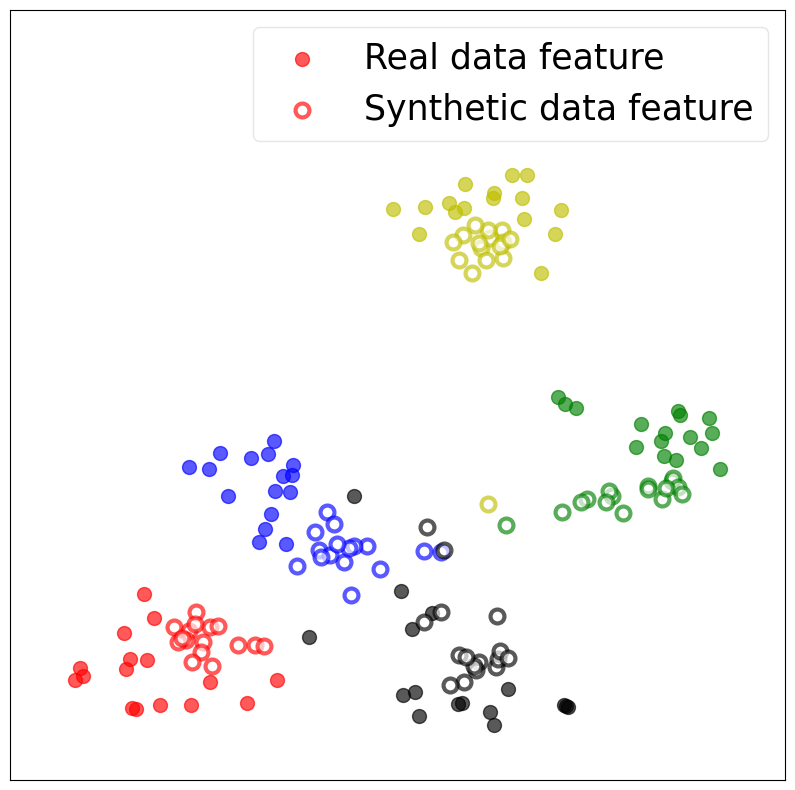} & \includegraphics[width=0.20\linewidth]{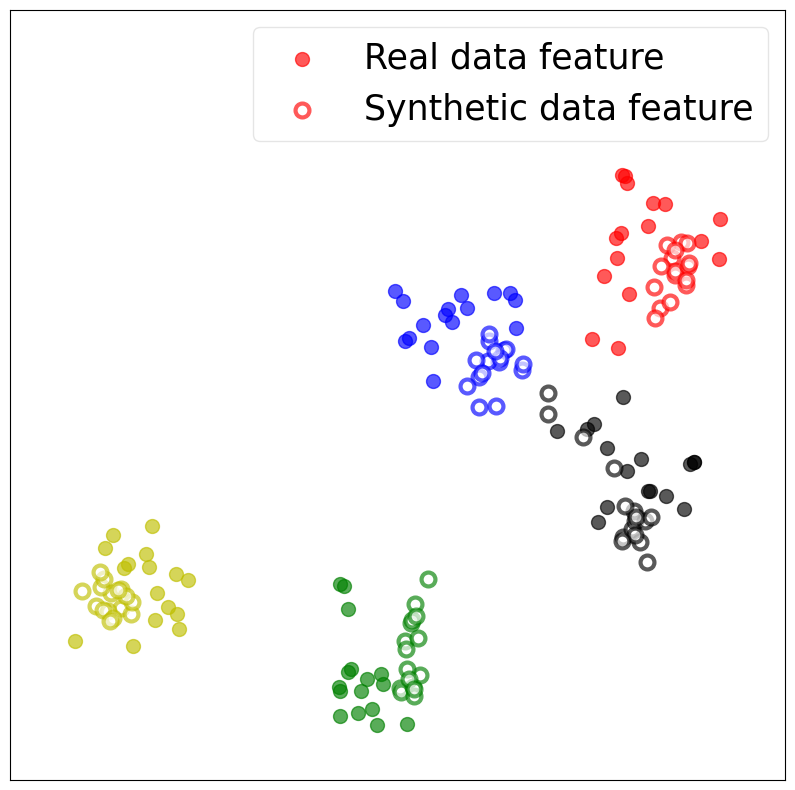} & \includegraphics[width=0.20\linewidth]{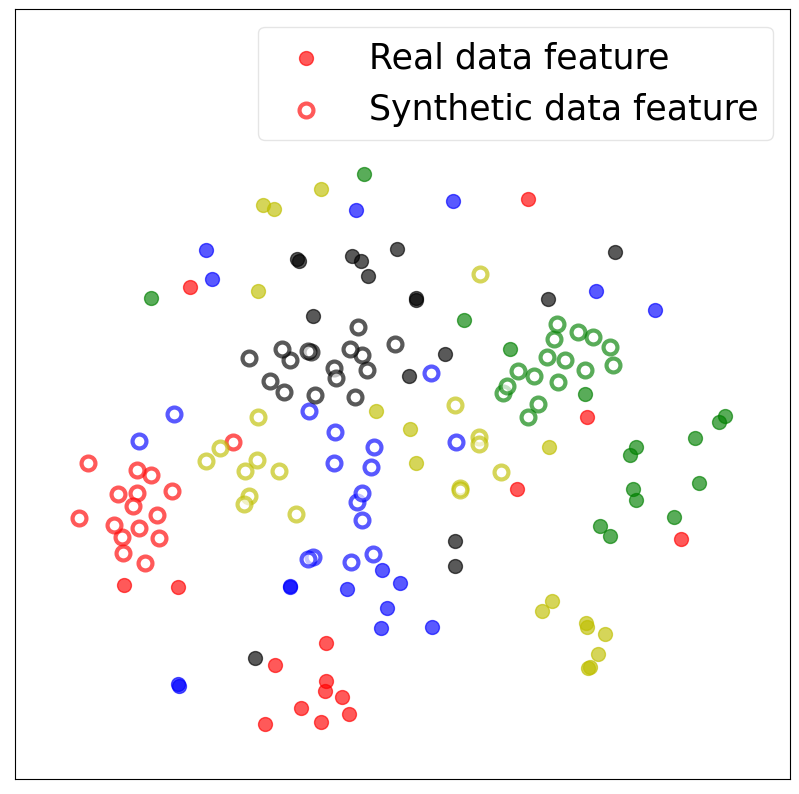} & \includegraphics[width=0.20\linewidth]{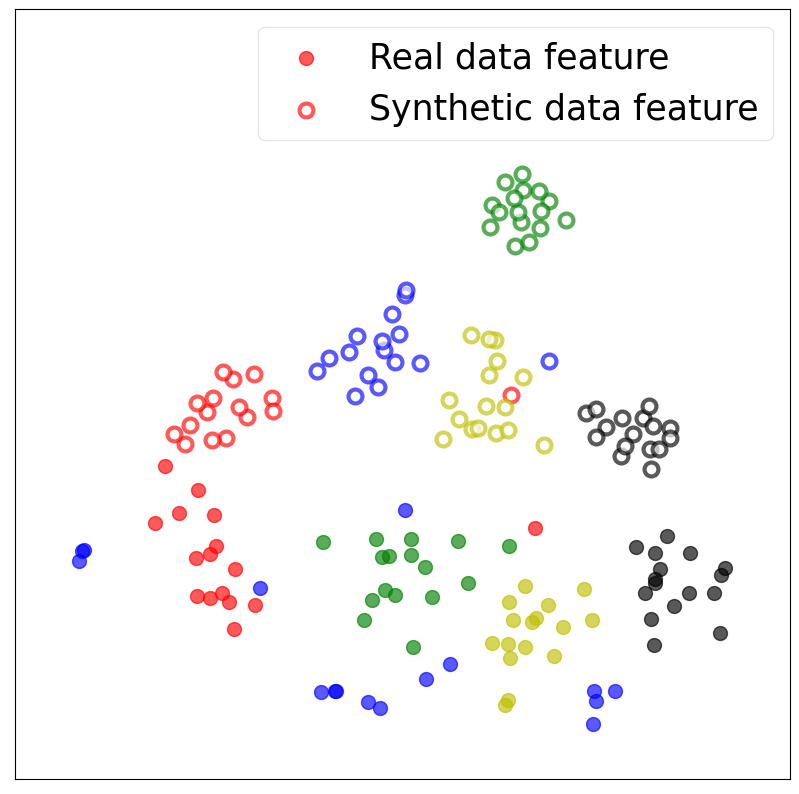}  \\
         \footnotesize Before Training & \footnotesize After Training & \footnotesize Before Training & \footnotesize After Training \\         \multicolumn{2}{c}{(e) SUN397} & \multicolumn{2}{c}{(f) DTD} \\

         \includegraphics[width=0.20\linewidth]{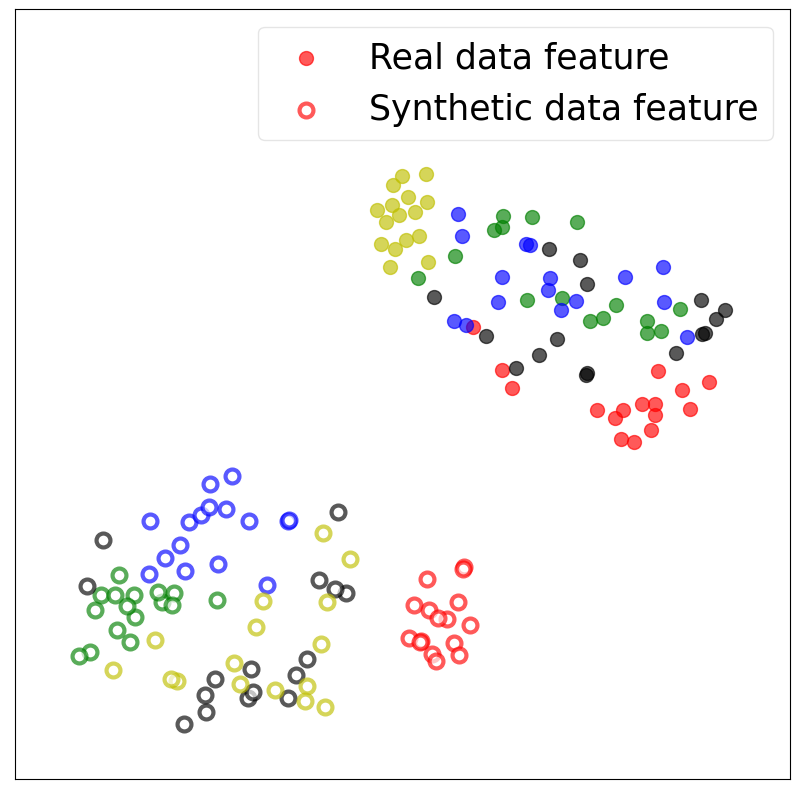} & \includegraphics[width=0.20\linewidth]{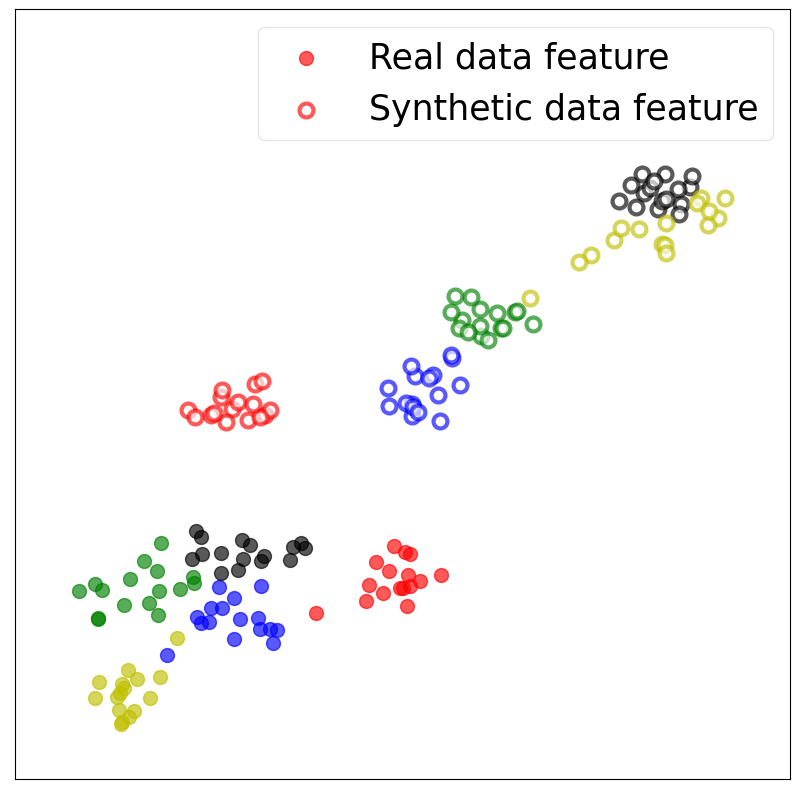} & \includegraphics[width=0.20\linewidth]{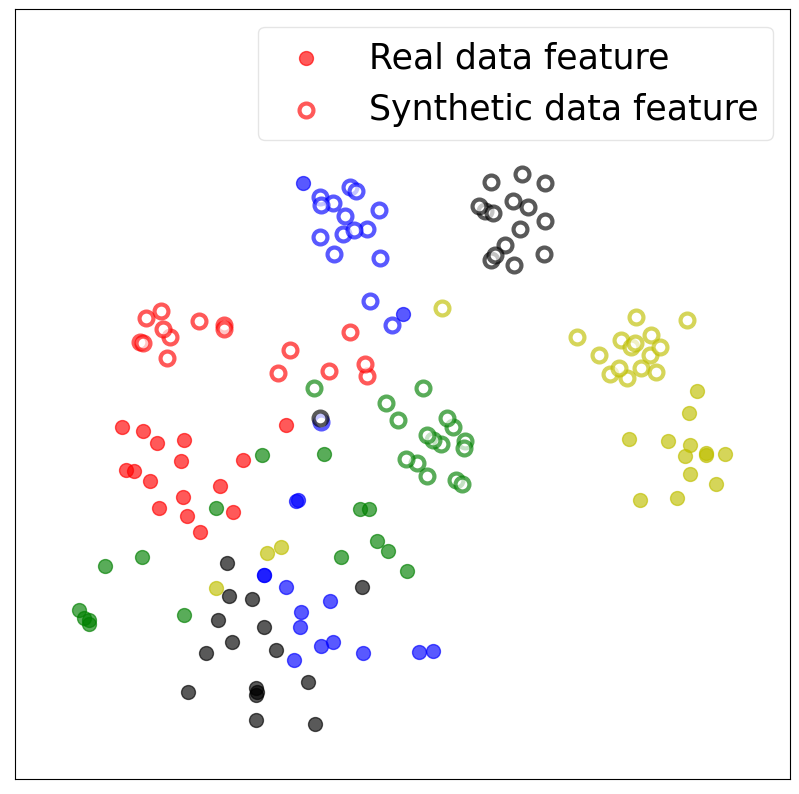} & \includegraphics[width=0.20\linewidth]{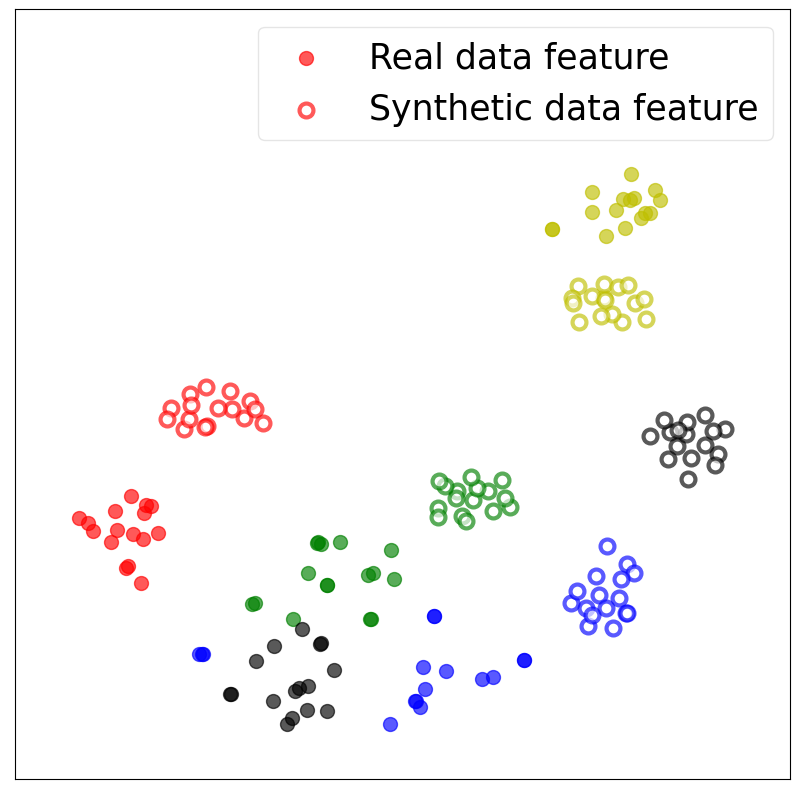}  \\
         \footnotesize Before Training & \footnotesize After Training & \footnotesize Before Training & \footnotesize After Training \\         \multicolumn{2}{c}{(g) EuroSAT} & \multicolumn{2}{c}{(h) UCF101} \\
         
         \includegraphics[width=0.20\linewidth]{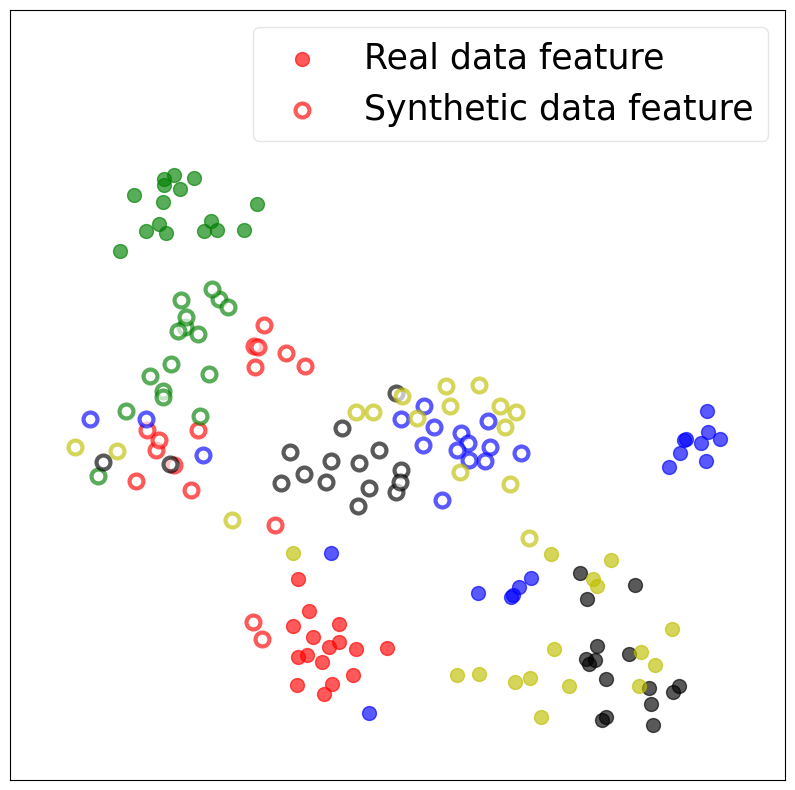} & \includegraphics[width=0.20\linewidth]{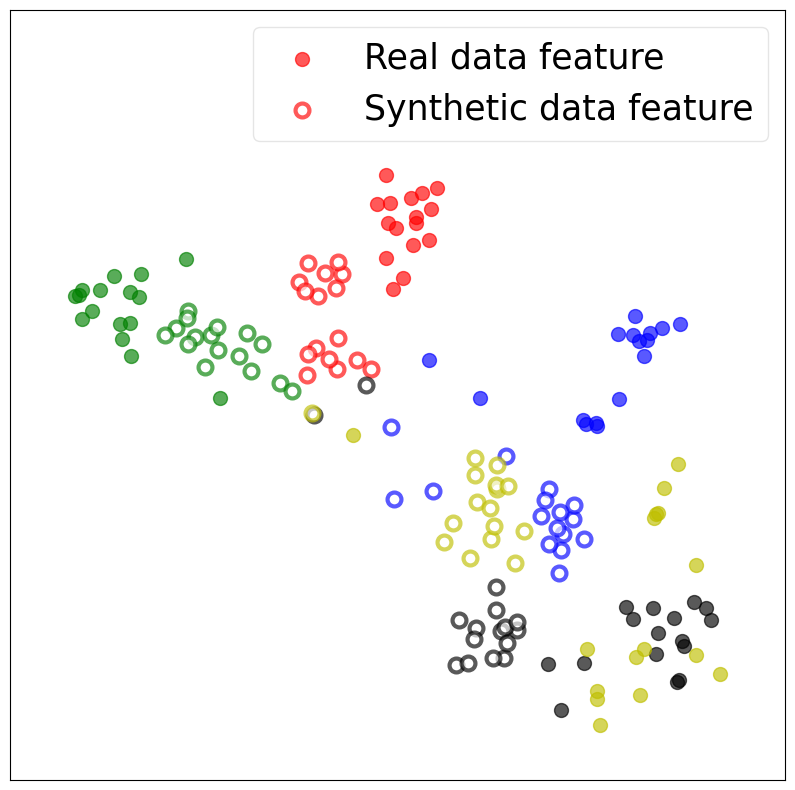} & 
         & \\
         \footnotesize Before Training & \footnotesize After Training & & \\
         \multicolumn{2}{c}{(i) ImageNet} & \\
    \end{tabular}
    \centering
    \caption{The t-SNE visualization results on other 9 datasets. The same color represents samples from the same category. All of these samples are from the novel class.}
    \label{fig:tsne-datasets}
\end{table*}

\begin{table*}[!htb]
    \centering
        \begin{tabular}{lccccc}
        \toprule
        Dataset & Classes & Train & Val & Test & Hand-crafted Prompt \\
        \midrule
        Caltech101 & 100 & 4,128 & 1,649 & 2,465 & a photo of a [CLS]. \\
        OxfordPets & 37 & 2,944 & 736 & 3,669 & a photo of a [CLS], a type of pet. \\ 
        StanfordCars & 196 & 6,509 & 1,635 & 8,041 & a photo of a [CLS]. \\
        Flowers102 &  102 & 4,093 & 1,633 & 2,463&  a photo of a [CLS], a type of flower. \\
        Food101 &  101 & 50,500 & 20,200 & 30,300 & a photo of [CLS], a type of food. \\
        FGVCAircraft &  100 & 3,334 & 3,333 & 3,333 & a photo of a [CLS], a type of aircraft. \\
        SUN397 &  397 & 15,880 & 3,970 & 19,850 & a photo of a [CLS]. \\
        DTD &  47 & 2,820 & 1,128 & 1,692 & [CLS] texture.\\
        EuroSAT &  10 & 13,500 & 5,400 & 8,100 & a centered satellite photo of [CLS]. \\
        UCF101 &  101 & 7,639 & 1,898 & 3,783 & a photo of a person doing [CLS].\\
        ImageNet &  1,000 & 1.28M & N/A & 50,000 & a photo of a [CLS] \\
        \midrule
        ImageNetV2	& 1,000	& N/A	& N/A	& 10,000 &  a photo of a [CLS] \\
        ImageNet-Sketch	& 1,000	& N/A	& N/A	& 50,889 &  a photo of a [CLS] \\
        ImageNet-A	& 200	& N/A	& N/A	& 7,500 &  a photo of a [CLS] \\
        ImageNet-R	& 200	& N/A	& N/A	& 30,000  & a photo of a [CLS] \\
        \bottomrule
        \end{tabular}    
    \caption{Detailed statistics of the datasets.}
    \label{tab:detail-datasets}
\end{table*}

\textbf{Hyperparameter Settings.} All images are randomly resized and cropped to 224 × 224, only random resize and random crop data augments are applied. We utilize the grid search to find the best hyper-parameters for all datasets. The $\alpha$ is set to 0.2 for ImageNet and Flowers102, and set to 0.1 for other datasets. The $\beta$ is set to 2.0 for EuroSAT and FGVCAircraft, and set to 0.5 for other datasets. For each result of SYNC-CLIP, we report the average result with three random seeds.

\textbf{t-SNE visualizations.} \cref{fig:tsne-datasets} illustrates the t-SNE visualization outcomes for nine additional datasets featuring novel classes. For each class, we randomly select 16 samples from both real and synthetic data. In datasets such as Caltech101 and SUN397, a commendable alignment is evident between synthetic and real data. However, in instances of failure, as observed in Flowers102 and DTD, a lack of alignment is notable, possibly due to substantial differences between synthetic and real data, potentially influenced by variations in the background of the real data. Notably, despite these disparities, certain similarities persist in the inter-class relationships within both synthetic and real data.

\subsection{Additional Synthetic Data Analysis}
\textbf{The synthetic data from different text-to-image models.} In this paper, the synthetic data are synthetic via the text-to-image models, \textit{i.e.}, DALL-E \cite{dall-e}, Stable Diffusion \cite{saharia2022photorealistic}. The synthetic data of the DALL-E model is from the public source\footnote{\href{https://github.com/OpenGVLab/CaFo}{https://github.com/OpenGVLab/CaFo}}. For the Stable Diffusion model, we utilize the public model\footnote{\href{https://github.com/Stability-AI/stablediffusion}{https://github.com/Stability-AI/stablediffusion}} to synthesize data. The ``a photo of a [category]" is used as the text input prompt for each category in the dataset. We show a part of synthetic data of Stable Diffusion \cite{saharia2022photorealistic} and the DALL-E model \cite{dall-e} in \cref{fig:vis_synthetic}. 

\textbf{The FID of synthetic data.} \cref{tab:fid} demonstrates the FID between the different synthetic data and the real data. We find that different models exhibit varying performances in terms of FID on different datasets. For instance, on fine-grained datasets such as StanfordCar and Flowers, Stable Diffusion outperforms the DALL-E model. Conversely, on the Caltech-101 dataset, DALL-E surpasses Stable Diffusion. Overall, although the synthetic data are high fidelity, they are different from the real data.

\begin{table*}
    \centering
    \resizebox{\linewidth}{!}{
        \begin{tabular}{cccccccccccc}
        \toprule
        Model                                &  Caltech101 &  OxfordPets & StanfordCars & Flowers102 & Food101 & FGVCAircraft & SUN397 & DTD & EuroSAT & UCF101 & ImageNet \\
        \midrule
        SD \cite{saharia2022photorealistic}  & 0.485 & 0.398 & 0.318 & 0.254 & 0.381 & 0.340 & 0.566 & 0.397 & 0.564 & 0.614 & 0.394\\
        DALL-E \cite{dall-e}                 & 0.337 & 0.327 & 0.460 & 0.332 & 0.516 & 0.498 & 0.507 & 0.440 & 0.550 & 0.514 & 0.442\\
        \bottomrule
        \end{tabular}    
    }
    \caption{The FID metrics of the synthetic data. Lower is better.}    
    \label{tab:fid}
\end{table*}

\begin{figure*}[!ht]
     \setlength{\tabcolsep}{0pt}
     \def\mywidth{.30}
     \begin{tabular}{l@{\hskip 10pt}c@{\hskip 5pt}c@{\hskip 5pt}c@{\hskip 5pt}}
     & \textbf{Real}  & \textbf{DALL-E} \cite{dall-e} & \textbf{SD\cite{saharia2022photorealistic}}\\
     \begin{turn}{90}
        \textbf{FGVC}
     \end{turn}
     & \includegraphics[width=\mywidth\linewidth]{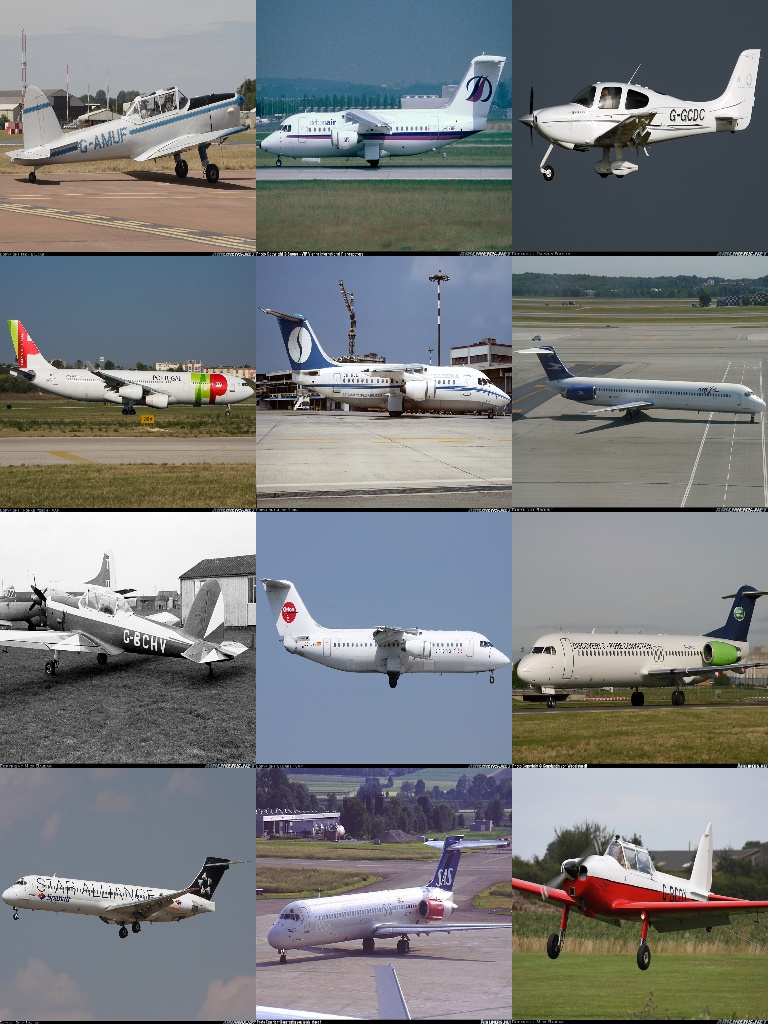} &
     \includegraphics[width=\mywidth\linewidth]{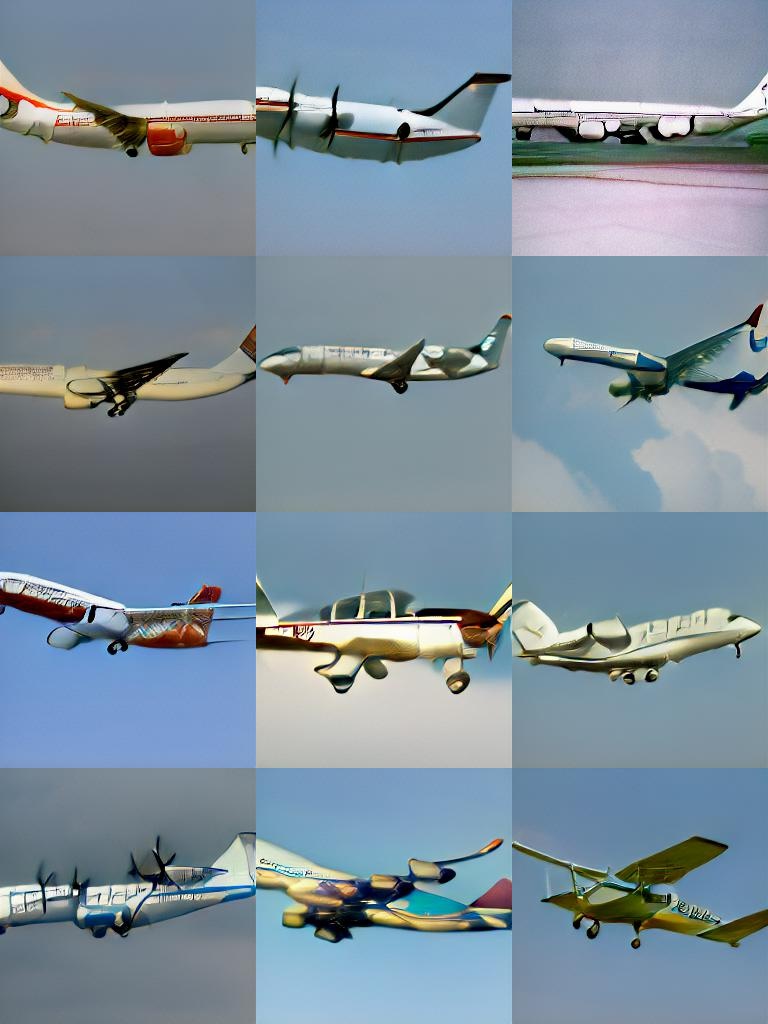} &
      \includegraphics[width=\mywidth\linewidth]{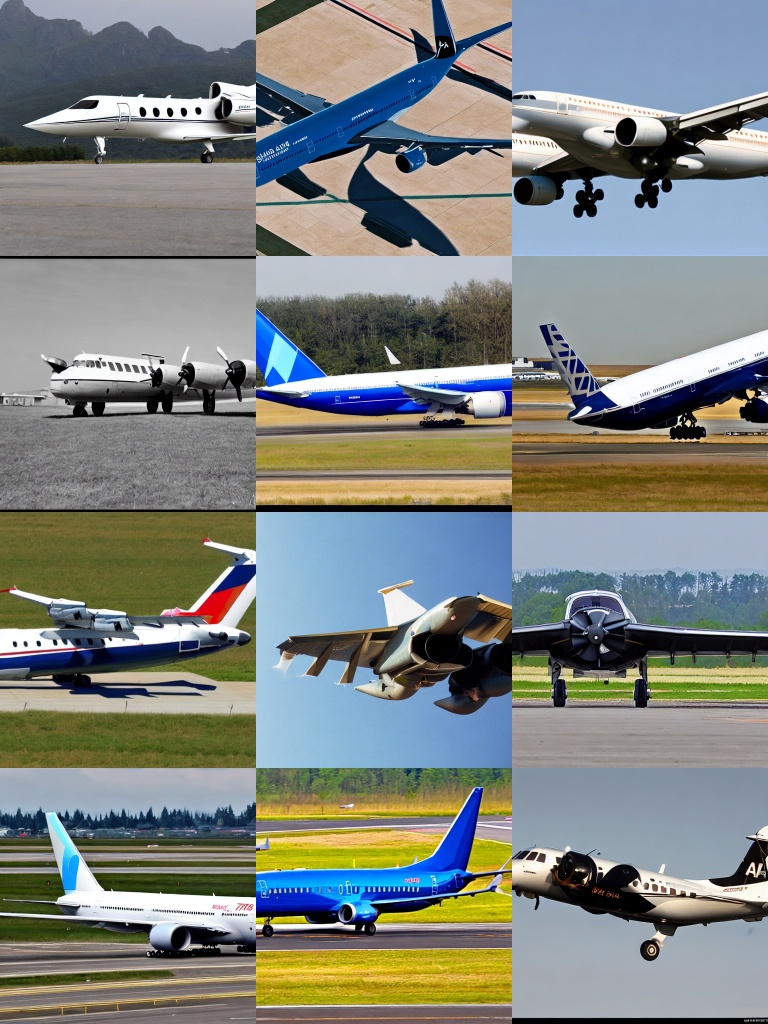} \\      
     \vspace{0.2cm}
     \begin{turn}{90}
        \textbf{Caltech101}
     \end{turn}
     & \includegraphics[width=\mywidth\linewidth]{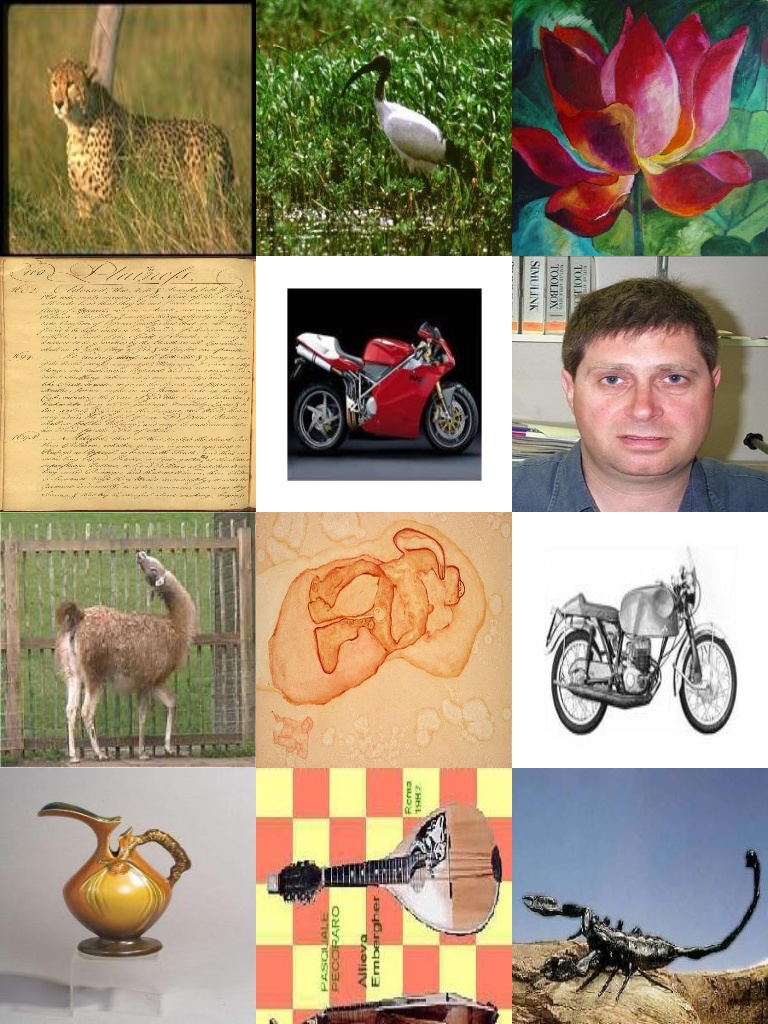} &
     \includegraphics[width=\mywidth\linewidth]{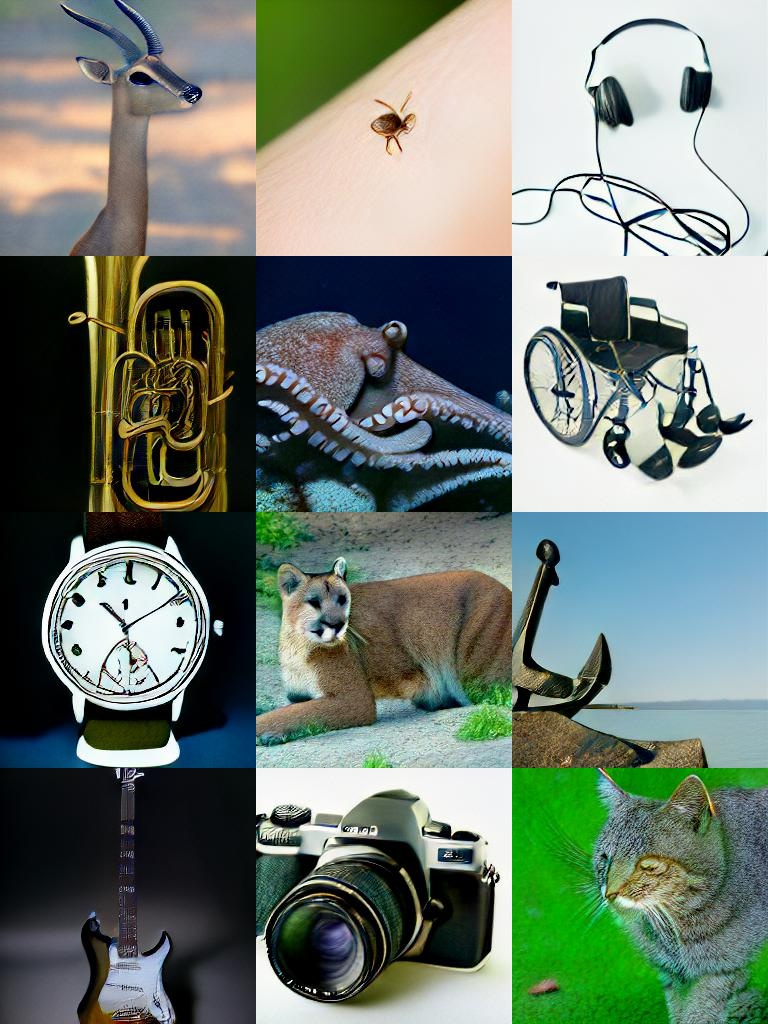} &
      \includegraphics[width=\mywidth\linewidth]{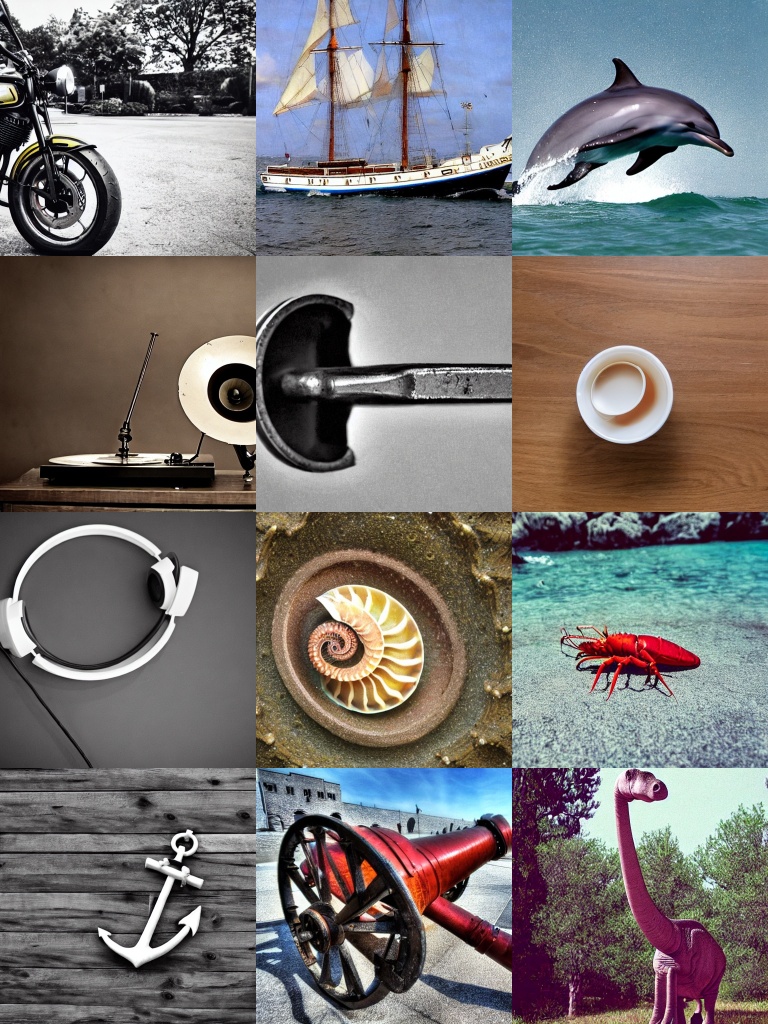} \\      
     \vspace{0.2cm}
     \begin{turn}{90}
        \textbf{OxfordPets}
     \end{turn}
     & \includegraphics[width=\mywidth\linewidth]{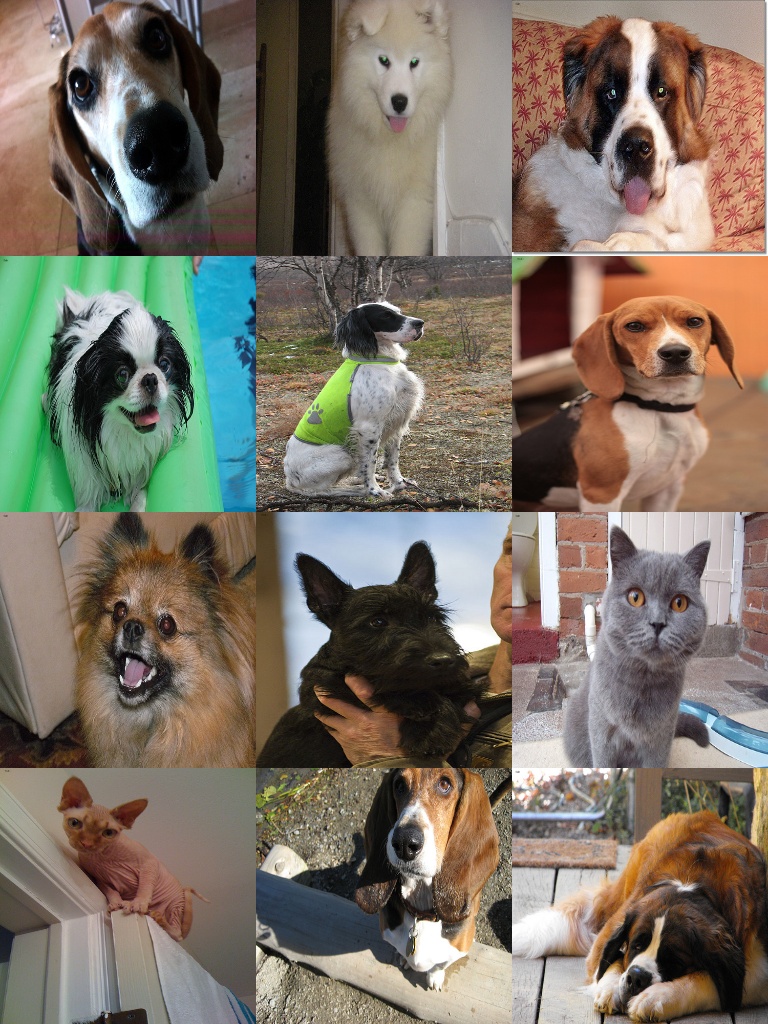} &
     \includegraphics[width=\mywidth\linewidth]{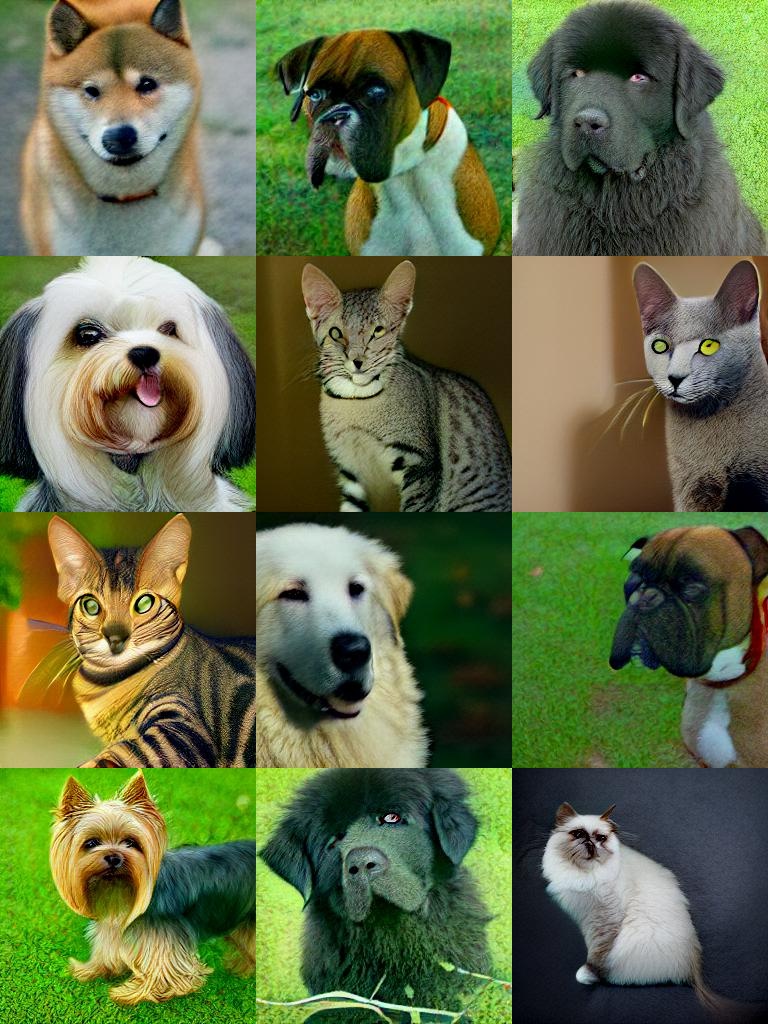} &
      \includegraphics[width=\mywidth\linewidth]{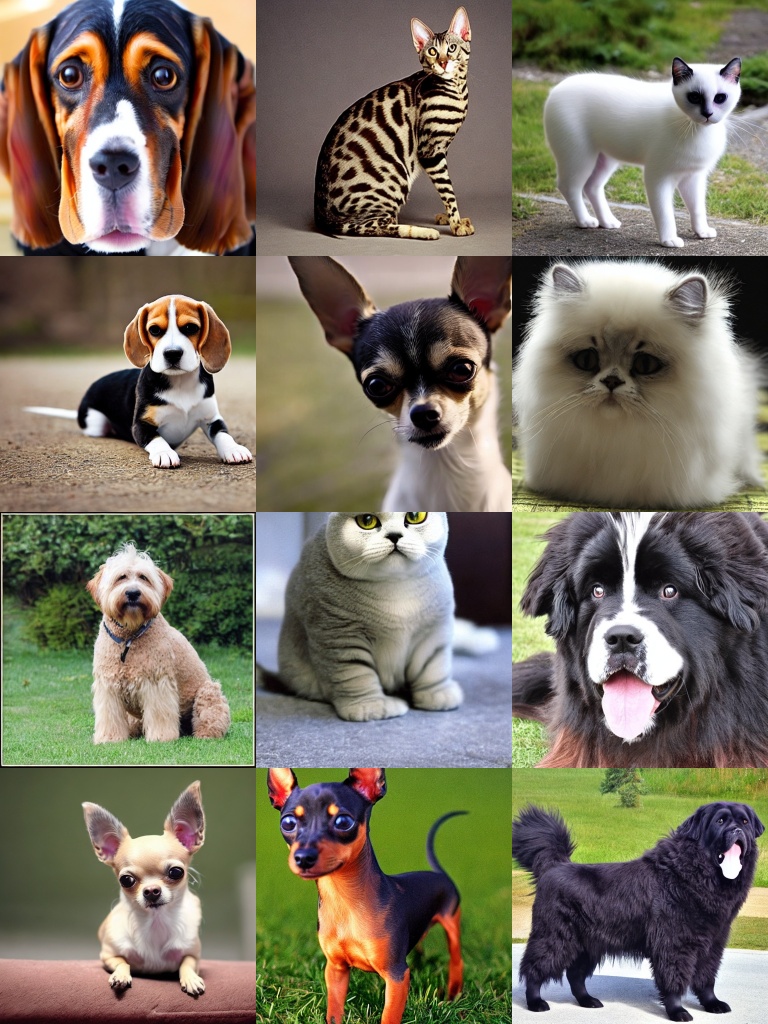} \\   
     \end{tabular}
     \vspace{-0.2cm}
    \caption{Comparison with the synthetic data and the real data.}
   \label{fig:vis_synthetic}
\end{figure*}

\begin{figure*}[!ht]
    \ContinuedFloat

     \setlength{\tabcolsep}{0pt}
     \def\mywidth{.30}
     \begin{tabular}{l@{\hskip 10pt}c@{\hskip 5pt}c@{\hskip 5pt}c@{\hskip 5pt}}
     & \textbf{Real} & \textbf{DALL-E} \cite{dall-e} & \textbf{SD\cite{saharia2022photorealistic}}\\
    \vspace{0.2cm}
     \begin{turn}{90}
        \textbf{StanfordCars}
     \end{turn}
     & \includegraphics[width=\mywidth\linewidth]{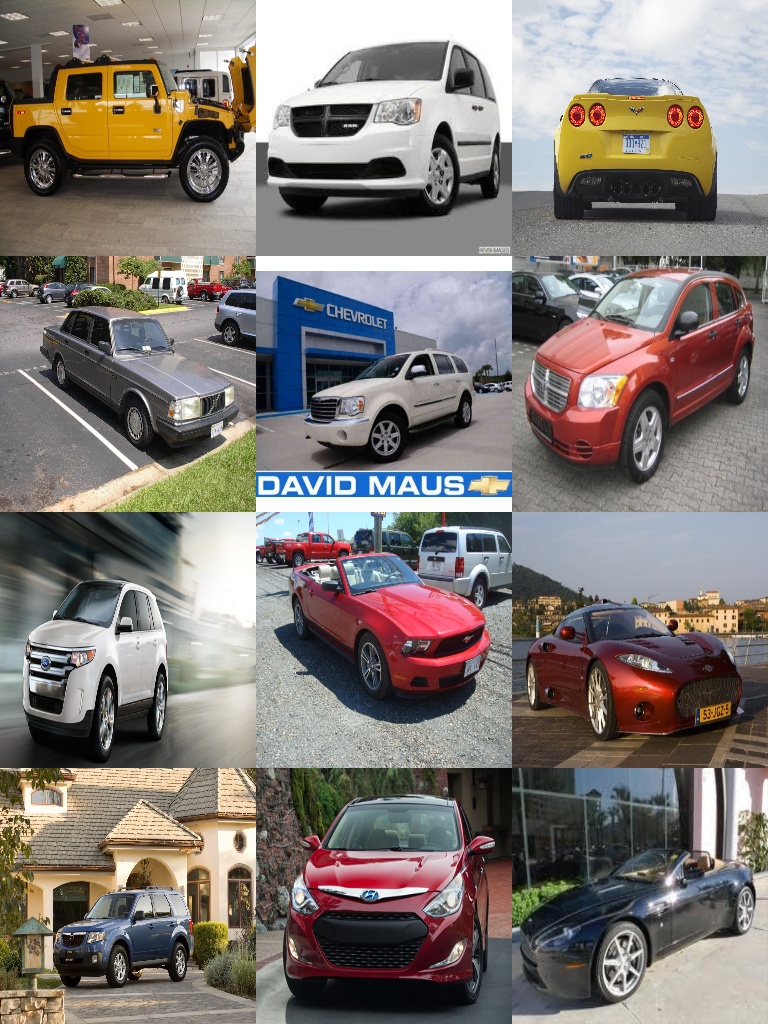} &
     \includegraphics[width=\mywidth\linewidth]{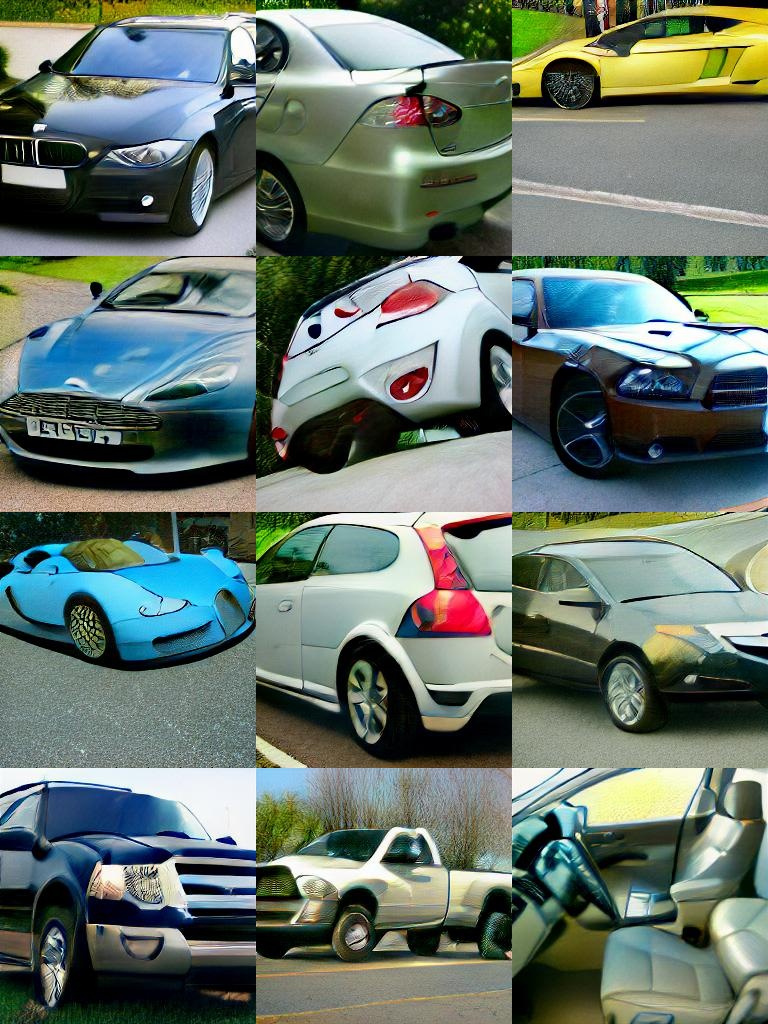} &
      \includegraphics[width=\mywidth\linewidth]{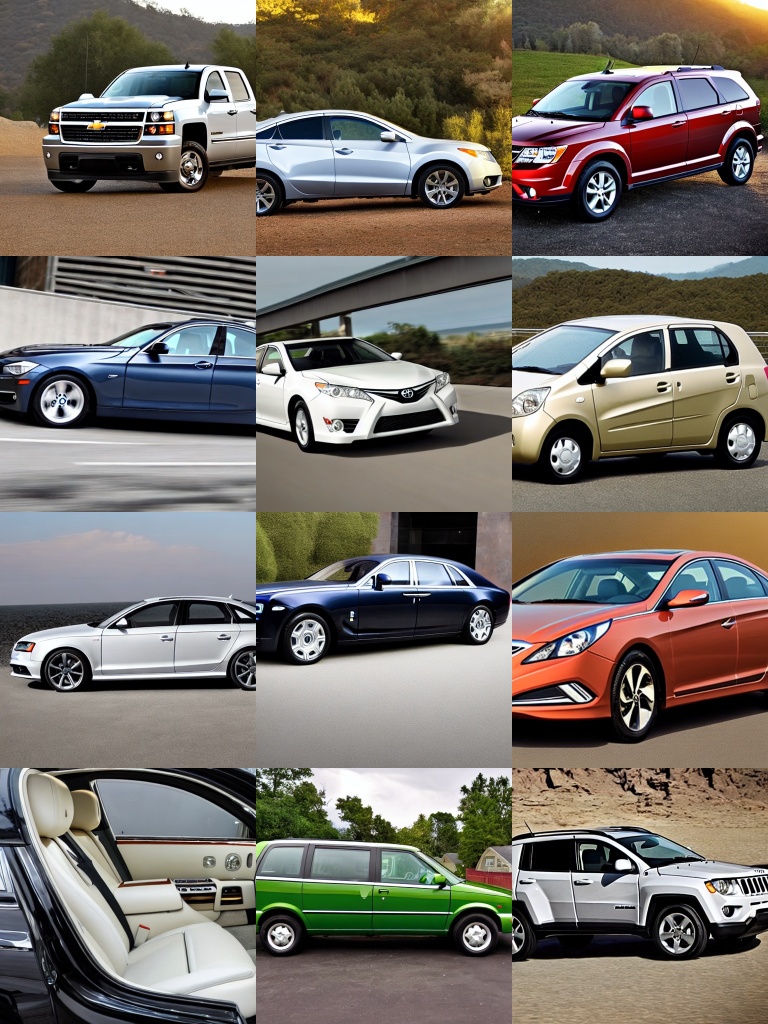} \\   

      \vspace{0.2cm}
     \begin{turn}{90}
        \textbf{Flowers102}
     \end{turn}
     & \includegraphics[width=\mywidth\linewidth]{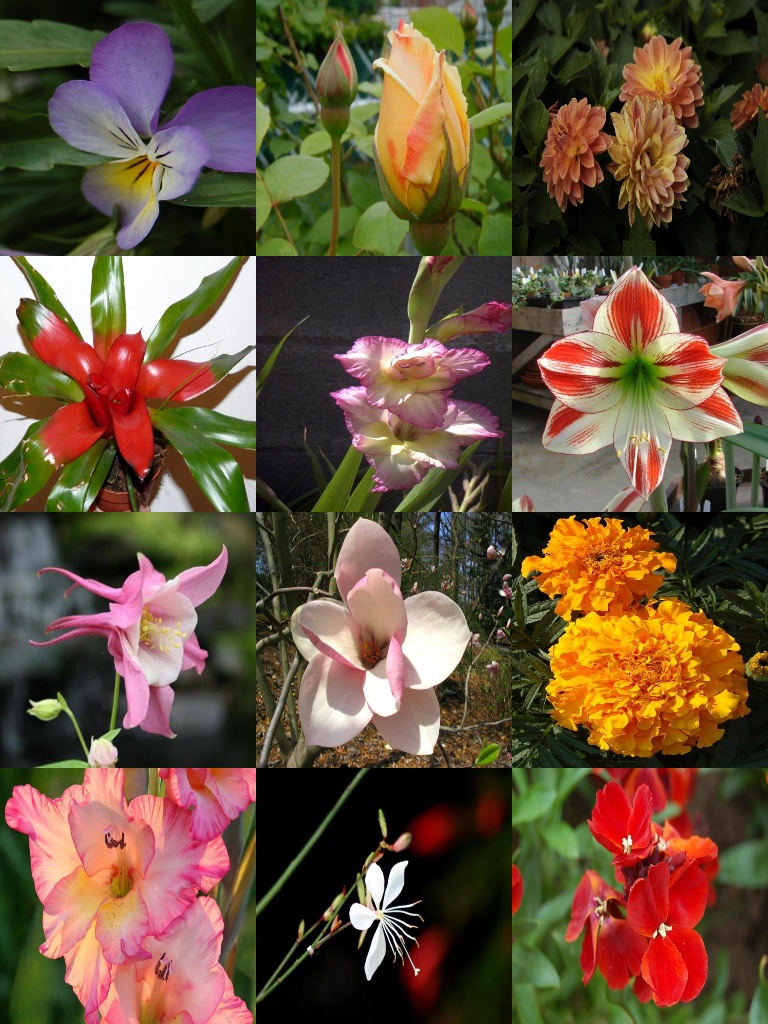} &
     \includegraphics[width=\mywidth\linewidth]{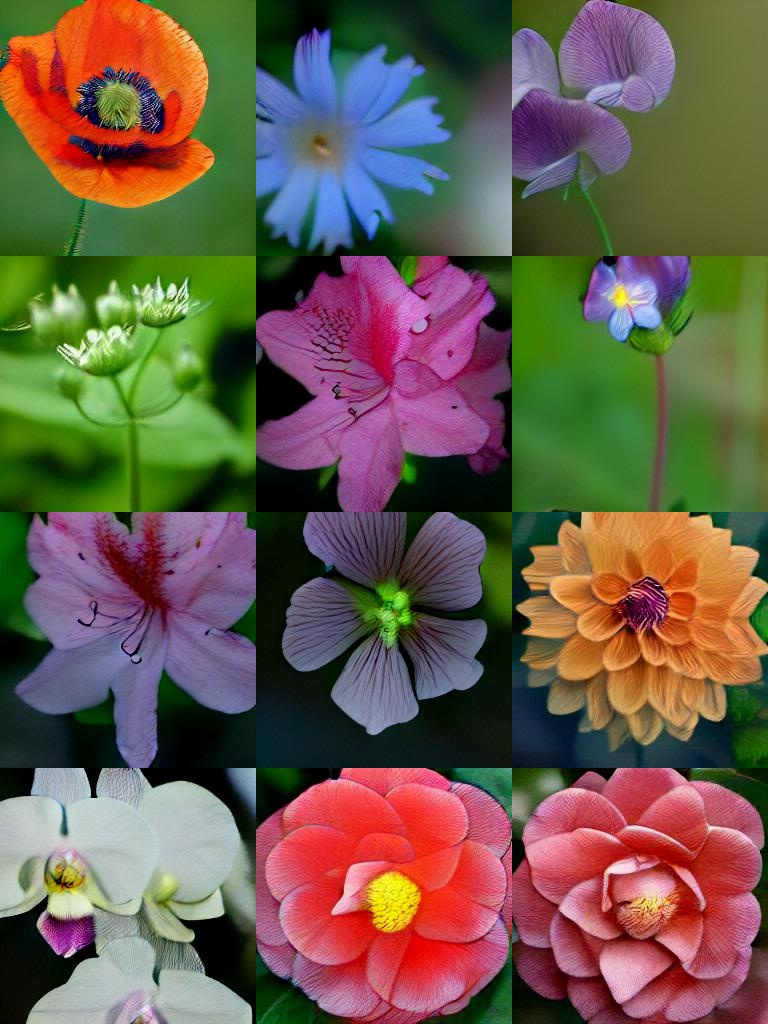} &
      \includegraphics[width=\mywidth\linewidth]{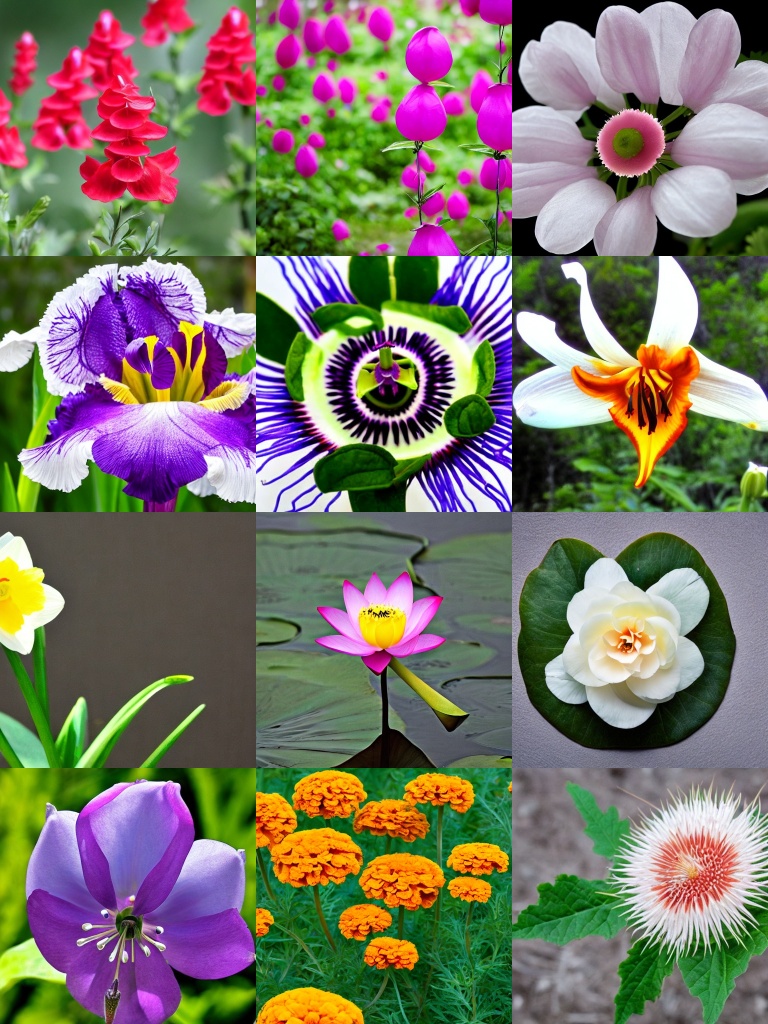} \\   

    \vspace{0.2cm}
     \begin{turn}{90}
        \textbf{Food101}
     \end{turn}
     & \includegraphics[width=\mywidth\linewidth]{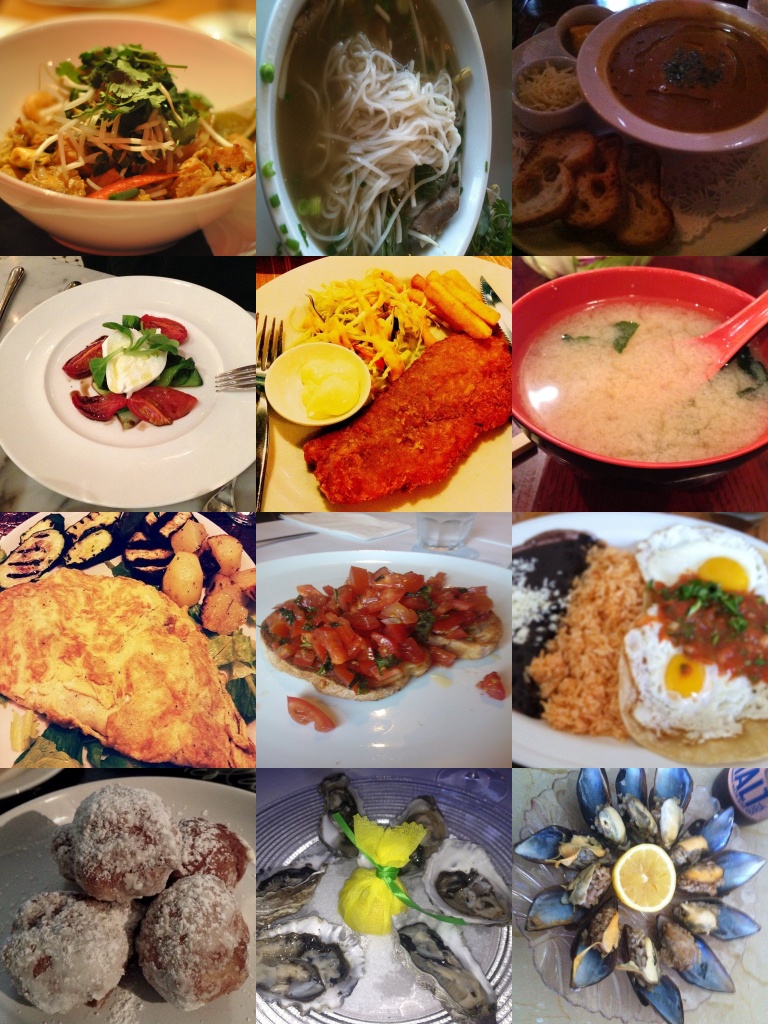} &
     \includegraphics[width=\mywidth\linewidth]{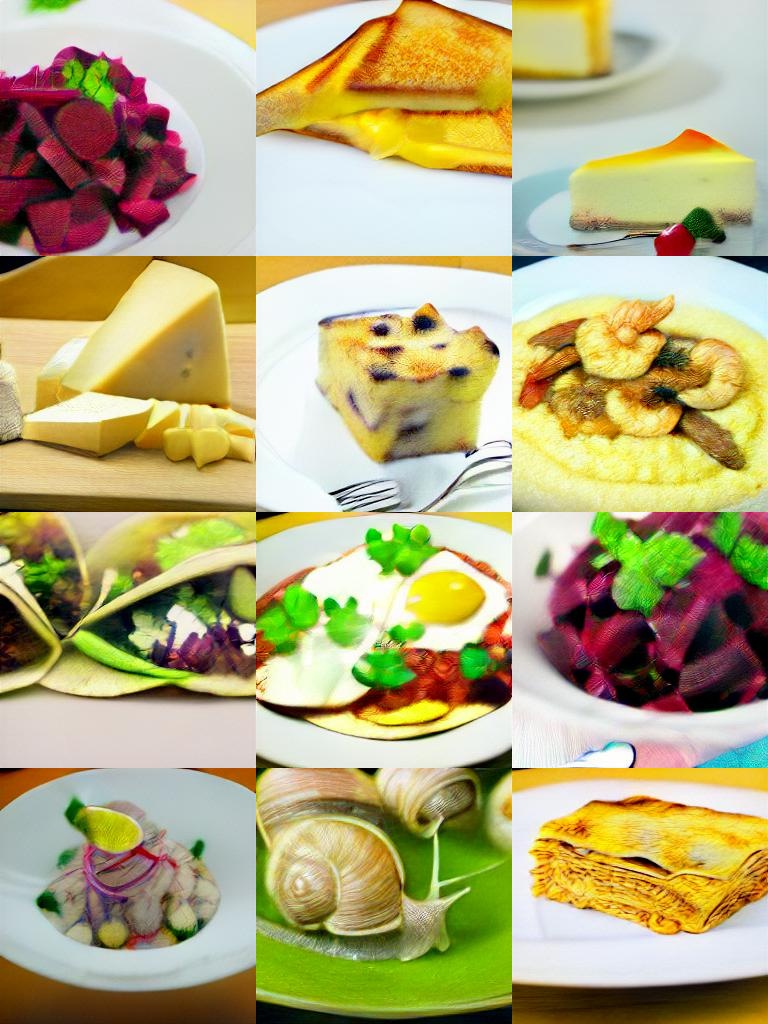} &
      \includegraphics[width=\mywidth\linewidth]{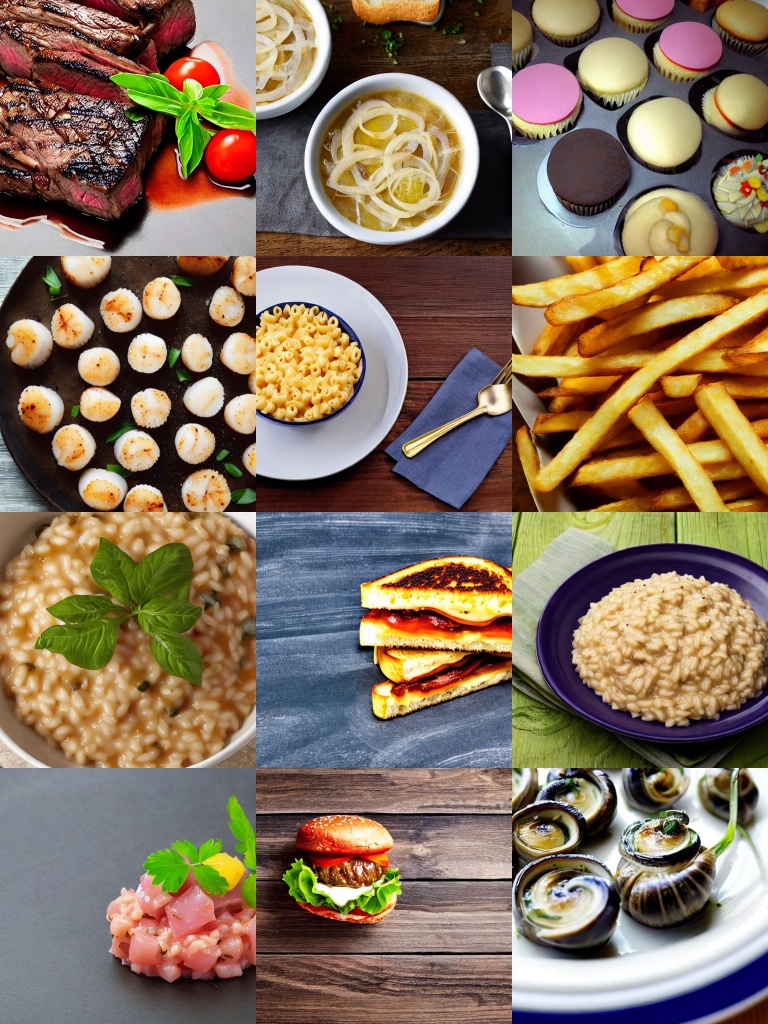} \\   

     \end{tabular}
     \vspace{-0.2cm}
    \caption{Comparison with the synthetic data and the real data.}
   \label{fig:vis_synthetic}
\end{figure*}

\begin{figure*}[!ht]
    \ContinuedFloat

     \setlength{\tabcolsep}{0pt}
     \def\mywidth{.30}
     \begin{tabular}{l@{\hskip 10pt}c@{\hskip 5pt}c@{\hskip 5pt}c@{\hskip 5pt}}
     & \textbf{Real} & \textbf{DALL-E} \cite{dall-e} & \textbf{SD\cite{saharia2022photorealistic}} \\
    \vspace{0.2cm}
     \begin{turn}{90}
        \textbf{SUN397}
     \end{turn}
     & \includegraphics[width=\mywidth\linewidth]{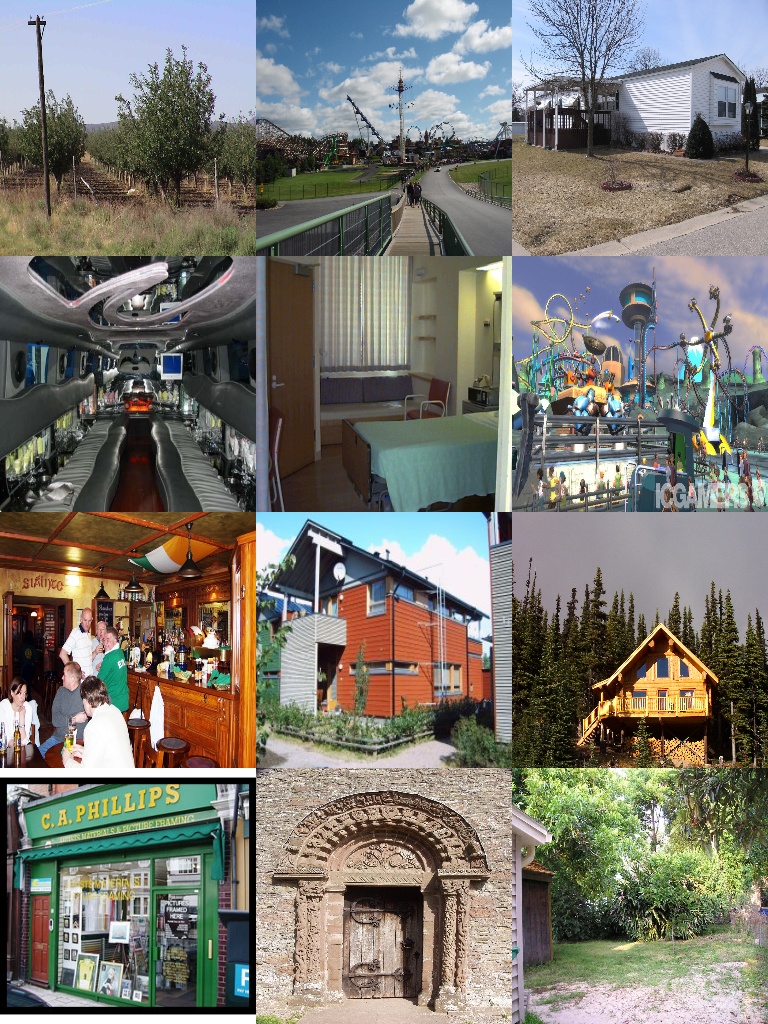} &
     \includegraphics[width=\mywidth\linewidth]{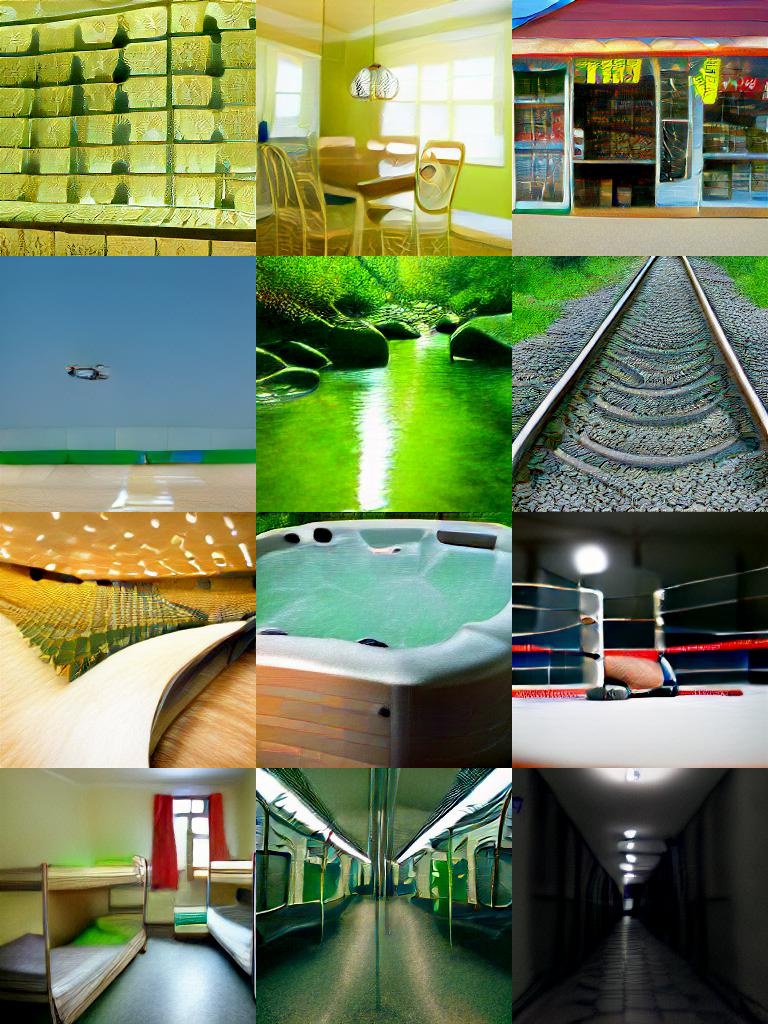} &
      \includegraphics[width=\mywidth\linewidth]{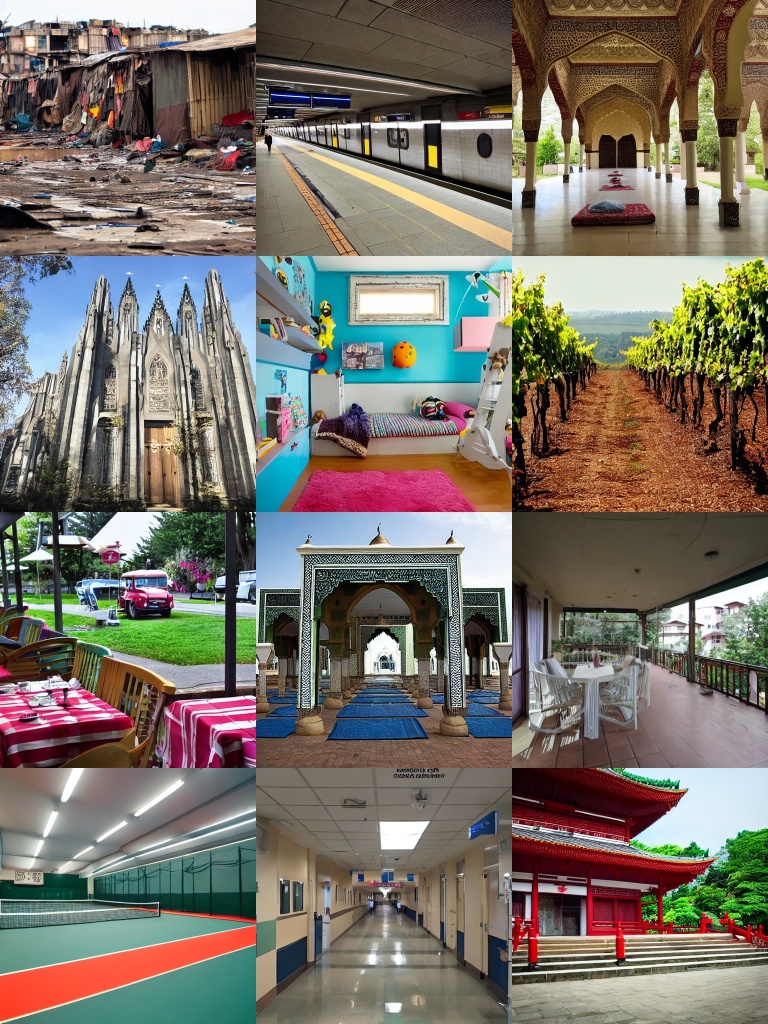} \\   

      \vspace{0.2cm}
     \begin{turn}{90}
        \textbf{DTD}
     \end{turn}
     & \includegraphics[width=\mywidth\linewidth]{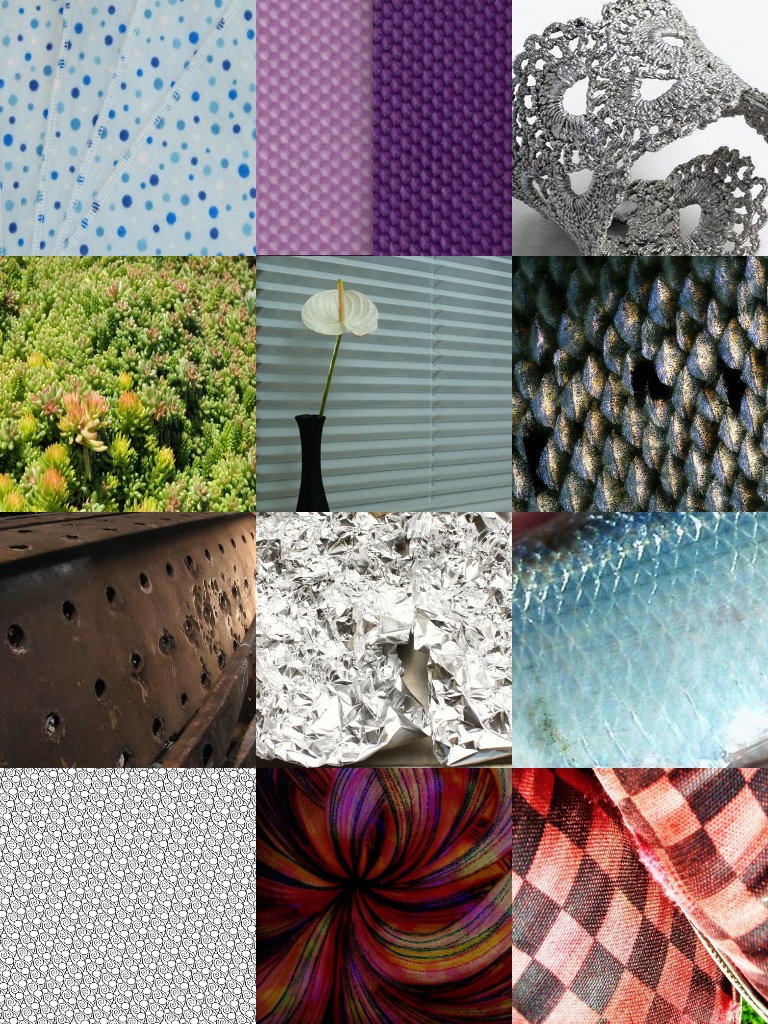} &
     \includegraphics[width=\mywidth\linewidth]{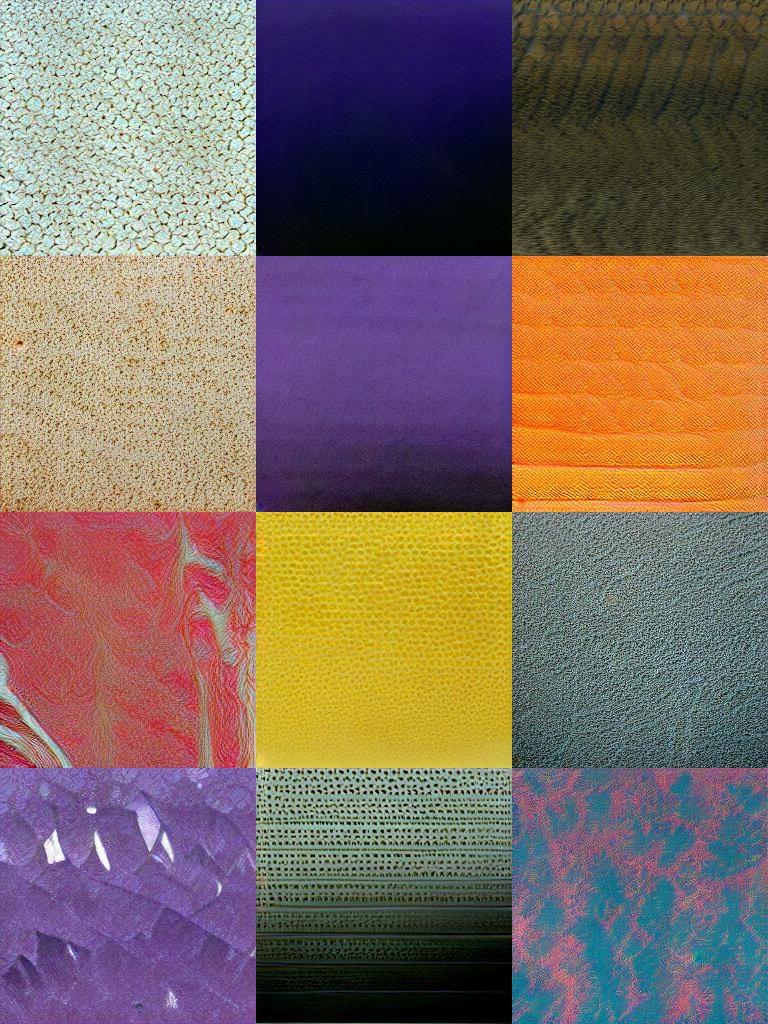} &
      \includegraphics[width=\mywidth\linewidth]{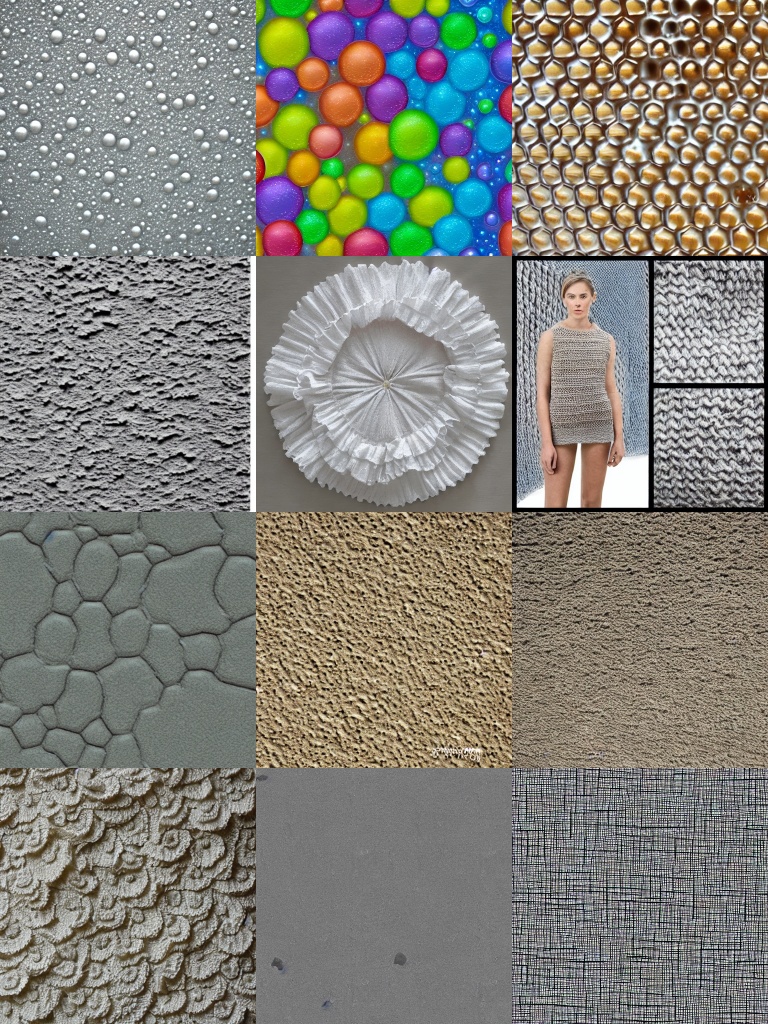} \\

      \vspace{0.2cm}
     \begin{turn}{90}
        \textbf{EuroSAT}
     \end{turn}
     & \includegraphics[width=\mywidth\linewidth]{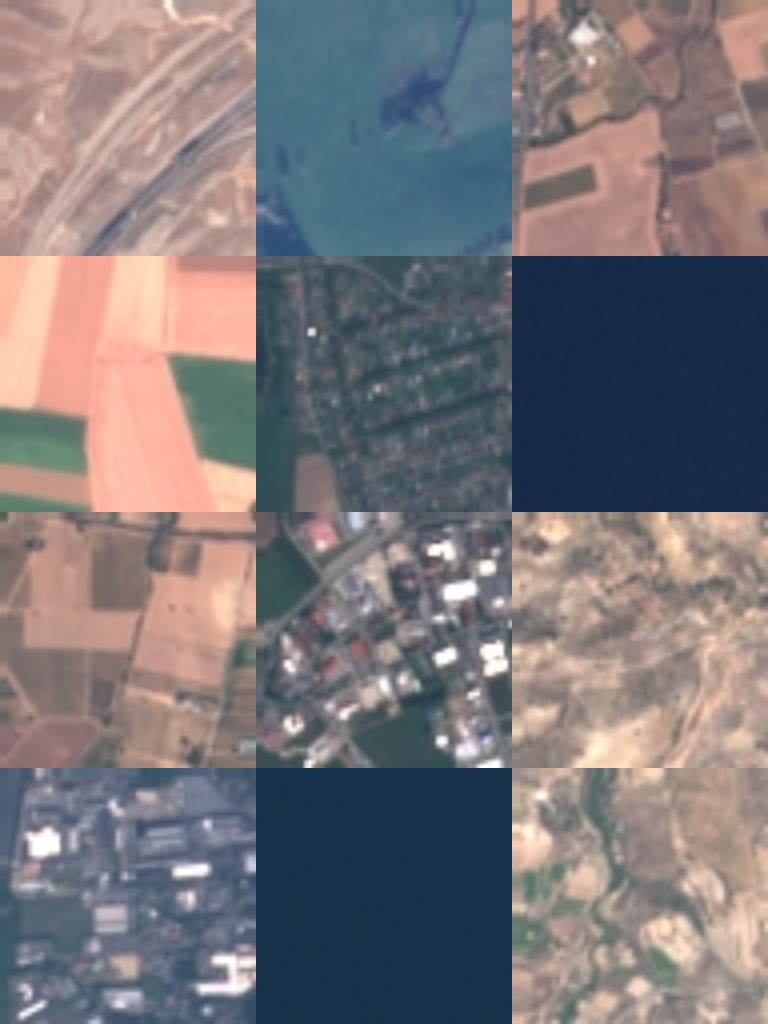} &
     \includegraphics[width=\mywidth\linewidth]{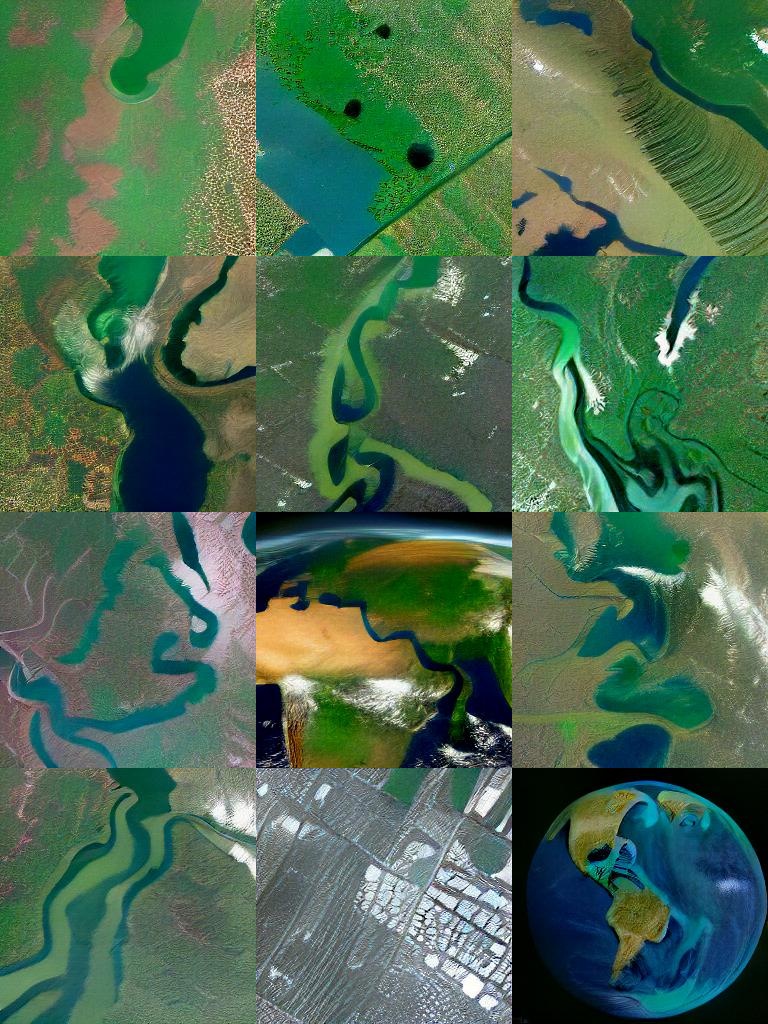} &
      \includegraphics[width=\mywidth\linewidth]{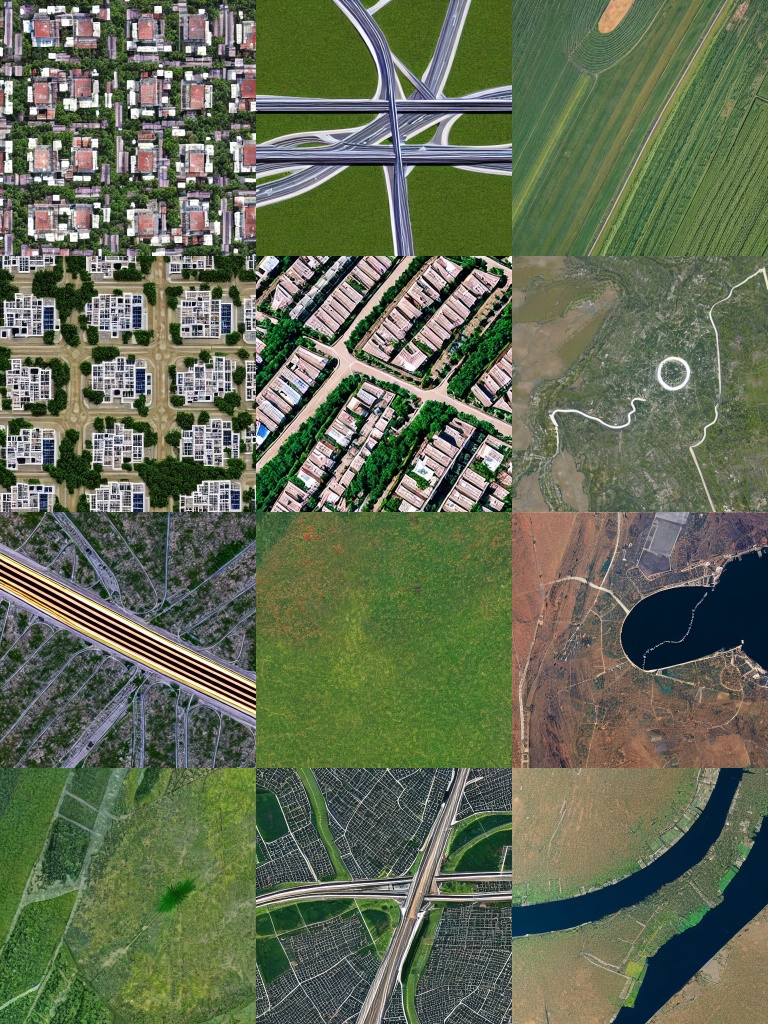} \\   
     \end{tabular}
     \vspace{-0.2cm}
    \caption{Comparison with the synthetic data and the real data.}
   \label{fig:vis_synthetic}
\end{figure*}

\begin{figure*}[!ht]
    \ContinuedFloat

     \setlength{\tabcolsep}{0pt}
     \def\mywidth{.30}
     \begin{tabular}{l@{\hskip 10pt}c@{\hskip 5pt}c@{\hskip 5pt}c@{\hskip 5pt}}
     & \textbf{Real} & \textbf{DALL-E} \cite{dall-e} & \textbf{SD\cite{saharia2022photorealistic}} \\
      \vspace{0.2cm}
     \begin{turn}{90}
        \textbf{UCF101}
     \end{turn}
     & \includegraphics[width=\mywidth\linewidth]{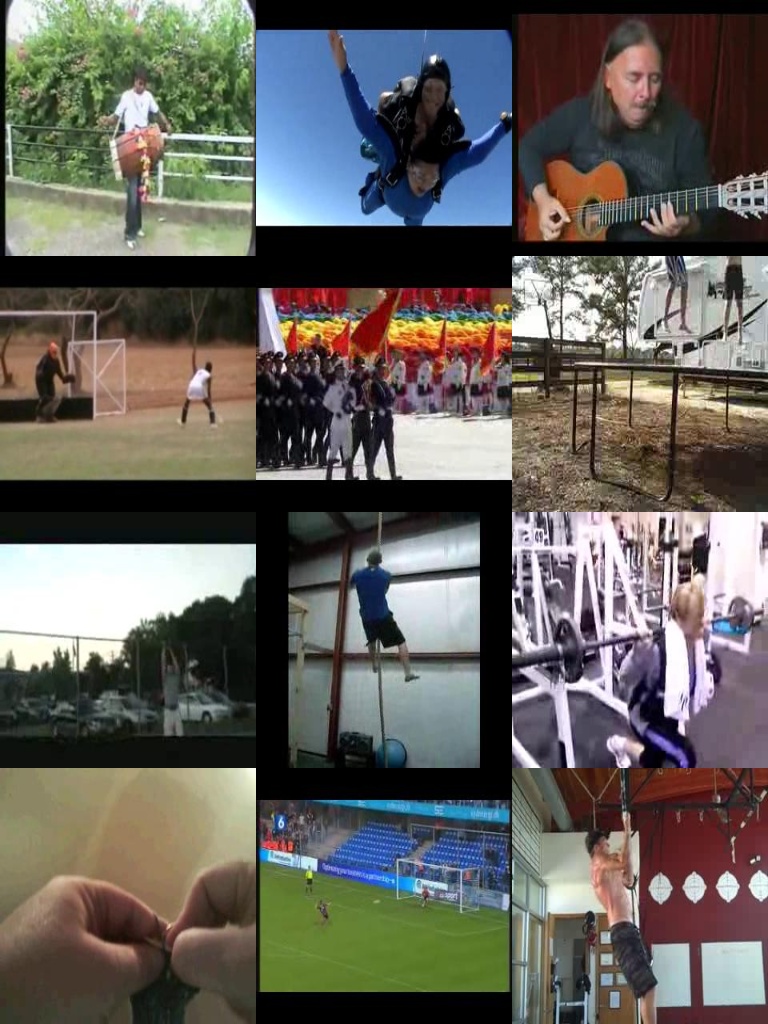} &
     \includegraphics[width=\mywidth\linewidth]{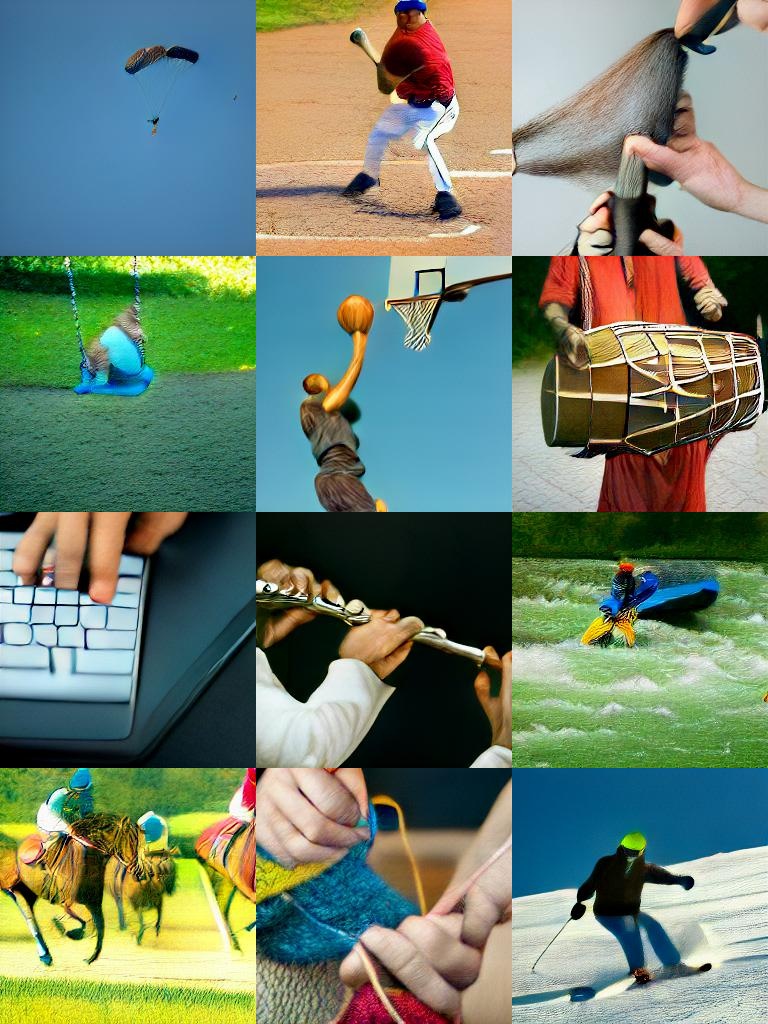} &
      \includegraphics[width=\mywidth\linewidth]{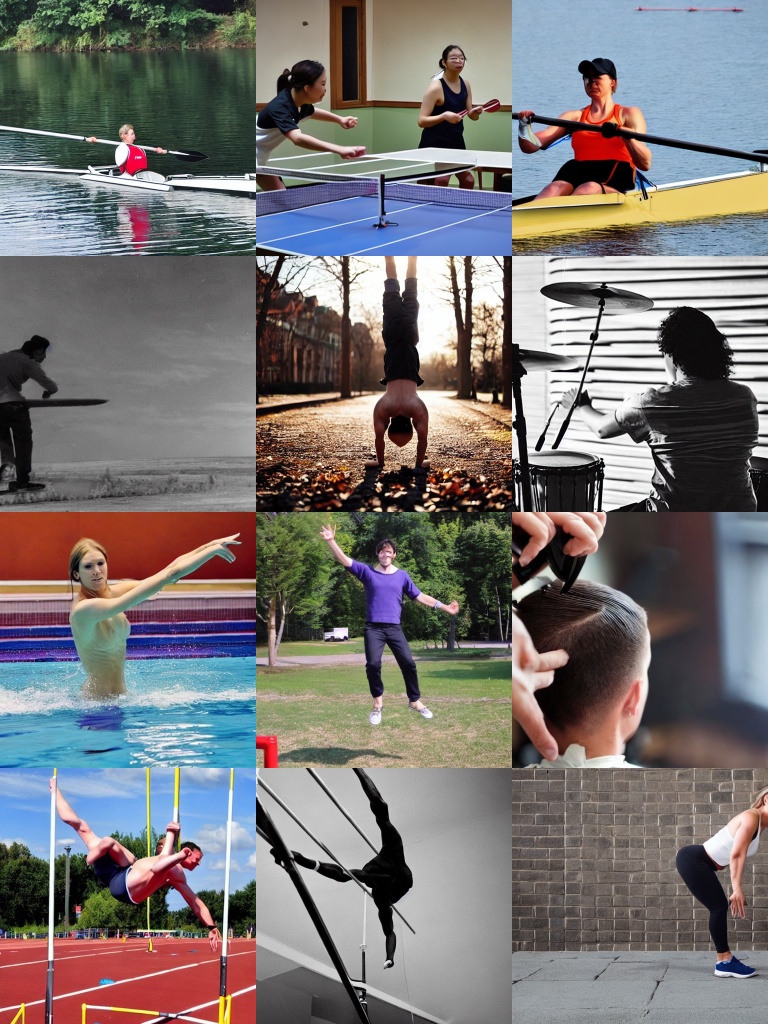} \\   

      \vspace{0.2cm}
     \begin{turn}{90}
        \textbf{ImageNet}
     \end{turn}
     & \includegraphics[width=\mywidth\linewidth]{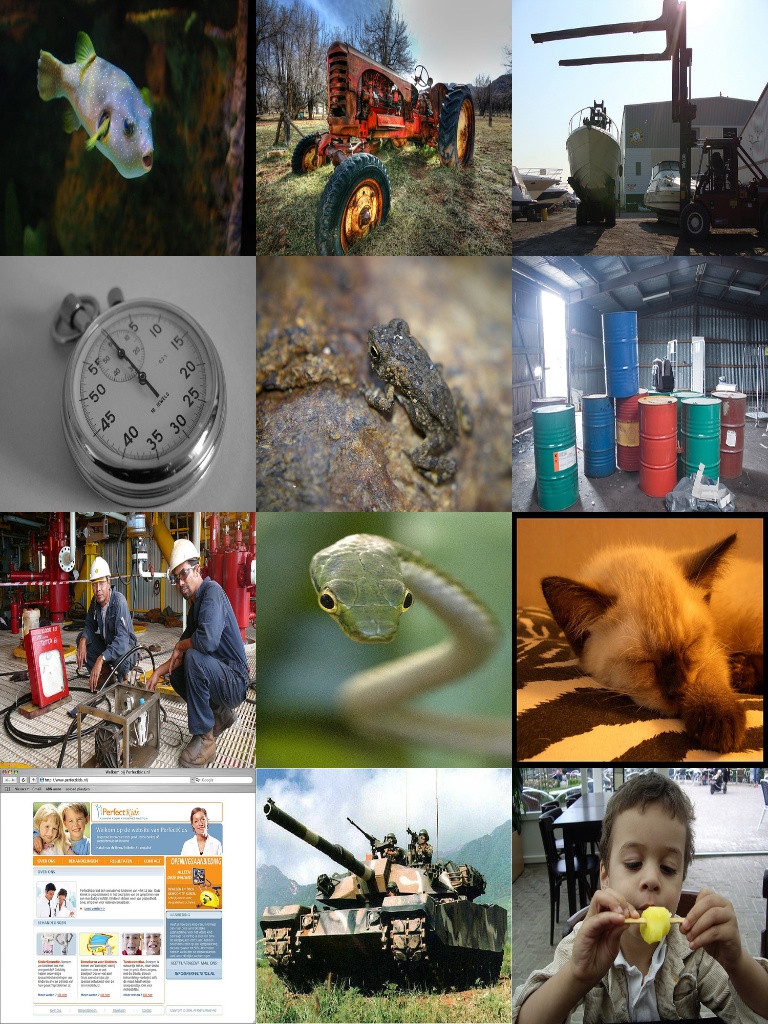} &
     \includegraphics[width=\mywidth\linewidth]{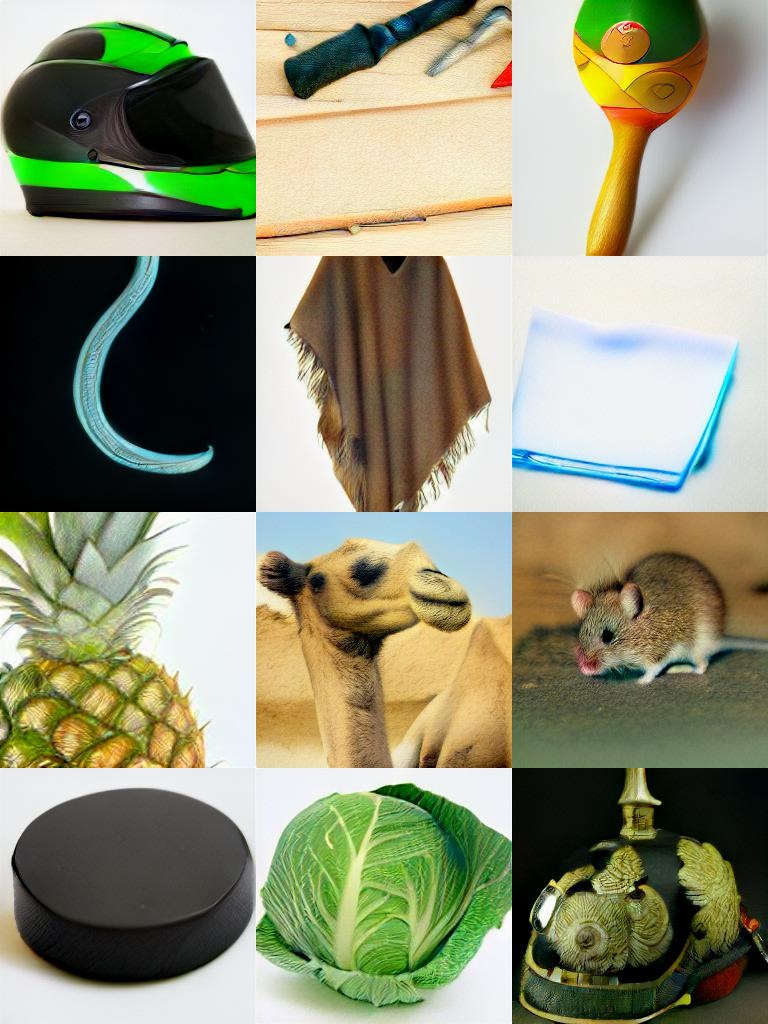} &
      \includegraphics[width=\mywidth\linewidth]{figures/images/imagenet_combined_image.jpg} \\   
      
     \end{tabular}
     \vspace{-0.2cm}
    \caption{Comparison with the synthetic data and the real data.}
   \label{fig:vis_synthetic}
\end{figure*}

\end{document}